\documentclass[11pt]{article}
\usepackage{enumerate}
\usepackage[OT1]{fontenc}
\usepackage[usenames]{color}
\usepackage{smile}
\usepackage[colorlinks,
linkcolor=red,
anchorcolor=blue,
citecolor=blue
]{hyperref}            
\usepackage{mathrsfs,fullpage}
\usepackage[protrusion=true, expansion=true]{microtype}
\usepackage{float,subfigure,caption,adjustbox,cancel,multirow,framed,xcolor}
\usepackage{amsfonts,amsmath,amssymb,amsthm,url,xspace,mathtools}
\usepackage{tikz}
\usepackage{verbatim,comment}
\usetikzlibrary{arrows,shapes}
\usepackage{authblk}
\usepackage[bottom]{footmisc}
\usepackage{lscape,diagbox}
\usepackage{algorithm}
\usepackage{algorithmicx,algpseudocode}
\usepackage[title]{appendix}
\ifx\counterwithout\undefined\usepackage{chngcntr}\fi
\counterwithout{equation}{section}
\definecolor{shadecolor}{RGB}{240,240,240}

\usepackage{math}
\allowdisplaybreaks[1]
\mathtoolsset{showonlyrefs=true}
\allowdisplaybreaks

\usepackage{alphalph}
\algrenewcommand{\algorithmiccomment}[1]{\hfill$\blacktriangleright$ #1}
\algnewcommand{\LongComment}[1]{\hfill$\triangleright$ #1}

\begin{document}

\title{Online Statistical Inference for Derivative-Free Constrained Stochastic Optimization}

\author[1]{Jianliang Ye}
\author[2]{Sen Na}
\affil[1]{School of Mathematics, Georgia Institute of Technology}
\affil[2]{School of Industrial and Systems Engineering, Georgia Institute of Technology}

\date{}

\maketitle

\begin{abstract}
We study statistical inference for nonlinear equality-constrained stochastic optimization problems in settings where derivative information is unavailable or expensive to obtain. While recent advances in stochastic sequential quadratic programming (SSQP) have enabled efficient online optimization, existing methods either rely on gradient information or suffer from high computational costs. To address these limitations, we develop a sketched derivative-free SSQP (SDF-SSQP) method that simultaneously achieves computational efficiency and valid statistical inference. By leveraging randomized sketching, the proposed method reduces the per-iteration computational cost to $O((d+m)^2)$, where $d$ is the dimension of the decision variable and $m$ is the number of constraints, making it well suited for large-scale problems. 
To overcome the challenges that both sketching and zeroth-order approximations introduce for statistical inference, we adopt a random-scaling approach that eliminates the need for explicit estimation of the asymptotic covariance matrix. We establish the asymptotic normality of the \textit{averaged} primal--dual iterates and construct a self-normalized statistic that is asymptotically pivotal.
The resulting inference procedure is fully online, requires neither covariance estimation nor matrix inversion, and incurs an additional per-iteration cost of the same order as that of the optimization update. Numerical experiments demonstrate the effectiveness of the proposed method in terms of both optimization performance and statistical inference accuracy.

\end{abstract}

\section{Introduction}

We consider solving nonlinear equality-constrained stochastic optimization problems of the form:
\begin{equation}\label{eq:opt_problem}
\min_{\bx \in \RR^d} f(\bx) = \EE_{\cP}\sbr{F(\bx; \xi)}, \quad \text{s.t.} \quad c(\bx) = \0,  
\end{equation}
where $f: \mR^d \rightarrow \RR$ denotes the stochastic objective function, $F(\cdot; \xi): \RR^d \rightarrow \RR$ denotes its realization with sample $\xi\sim\cP$, and $c:\RR^d\rightarrow \RR^m$ denotes deterministic equality constraints. Problem \eqref{eq:opt_problem} arises in a broad range of applications spanning machine learning, statistics, and scientific computing. 
In statistical learning, it naturally appears as a constrained parameter estimation problem \citep{Geyer1991Constrained, Hansen1982Large}, where $\xi$ represents an observed data sample and $f(\bx)$ denotes the expected population loss or negative log-likelihood. 
Similar constrained stochastic optimization problems also arise in~portfolio allocation optimization \citep{Du2022High, Fan2012Vast}, where the objective measures the expected performance of weighted random asset returns, while the portfolio weights must satisfy structural constraints such as budget, risk, or exposure conditions. Related formulations have further emerged in scientific machine learning, including physics-informed neural networks \citep{Raissi2019Physics} and stochastic PDE-constrained learning problems \citep{Kouri2014Inexact}, where the constraints encode physical laws or governing equations. These diverse applications motivate the development of efficient stochastic optimization algorithms for solving \eqref{eq:opt_problem} and conducting valid statistical inference on the resulting primal--dual solutions.

A classical approach for solving constrained stochastic optimization problems of the form \eqref{eq:opt_problem} is the sample average approximation (SAA) method. Given $t$ i.i.d. samples $\xi_1,\ldots, \xi_t\sim \cP$, SAA replaces the population objective by the empirical average and solves
\begin{equation*}
\min_{\bx \in \RR^d}\; \hat{f}_t(\bx):= \frac{1}{t}\sum_{i=1}^tF(\bx; \xi_i), \quad \text{s.t.}\quad c(\bx) = \0.
\end{equation*}
Let $(\hat{\bx}_t, \hat{\blambda}_t)$ denote a local primal--dual solution of this problem, also called the SAA estimator or constrained $M$-estimator. Under suitable regularity conditions, the SAA estimator exhibits $\sqrt{t}$-consistency and asymptotic normality \citep[Chapter 5]{Shapiro2021Lectures}:
\begin{equation}\label{equ:SAA}
\sqrt{t}\cdot(\hat{\bx}_t - \bx^\star, \hat{\blambda}_t- \blambda^\star) \overset{d}{\longrightarrow} \cN(\zero, \bSigma^\star),
\end{equation}
where $(\bx^\star,\blambda^\star)$ is a local solution of Problem \eqref{eq:opt_problem}, the Lagrangian function of \eqref{eq:opt_problem} is defined as
\begin{equation*}
\cL(\bx,\blambda) = f(\bx) + \blambda^\top c(\bx),
\end{equation*}
and the limiting covariance $\bSigma^\star$ has a sandwich form
\begin{equation}\label{equ:cov}
\bSigma^\star = (\nabla^2\cL(\bx^\star,\blambda^\star))^{-1}\text{\diag}\big(\Cov(\nabla F(\bx^\star;\xi)), \0\big)(\nabla^2\cL(\bx^\star,\blambda^\star))^{-1}.
\end{equation}
These classical results provide a statistical foundation for inference on the optimal solution $(\bx^\star,\blambda^\star)$. However, offline approaches such as SAA require repeatedly solving increasingly large empirical optimization problems, resulting in substantial computational and memory costs. This has motivated the development of online methods that process data sequentially and update solutions in a streaming~fashion. Projection-based stochastic gradient descent is a representative example: it enforces feasibility by projecting each iterate onto the constraint set. Almost-sure convergence, asymptotic normality, and online inference results for such methods have been established in \cite{Davis2019Stochastic,Duchi2021Asymptotic,Davis2024Asymptotic, Jiang2025Online}. Despite these advances, projection-based methods can be impractical when the feasible set is defined by nonlinear equality constraints $c(\bx)=\0$, since each projection may itself require solving a nontrivial constrained optimization problem.~In~deterministic nonlinear constrained optimization, this difficulty has long motivated projection-free algorithmic frameworks, including penalty methods, barrier methods, augmented Lagrangian methods, and sequential quadratic programming (SQP) methods \citep{Bertsekas1999Nonlinear,Nocedal2006Numerical}.

Building on these deterministic constrained optimization methods, a growing body of recent literature has extended projection-free ideas to stochastic online settings. Stochastic SQP (SSQP) methods extend classical SQP by incorporating stochastic first- and second-order information and updating the primal--dual variables by solving a sequence of quadratic subproblems \citep{Berahas2021Sequential, Berahas2023Stochastic, Na2022adaptive, Na2023Inequality, Fang2024Fully, Curtis2024Stochastic}. A representative development is the Adaptive Inexact SSQP (AI-SSQP) framework \citep{Na2025Statistical}, which is designed for online inference with an adaptive random stepsize rule and a randomized sketching solver for the quadratic~subproblems. This work establishes the asymptotic normality of the last AI-SSQP iterate and shows that, although the sketching solver reduces the per-iteration cost of solving the SQP~subproblem, it also introduces a computational--statistical tradeoff by enlarging the limiting covariance matrix relative to an exact subproblem solver, whose covariance would match \eqref{equ:cov}. For practical inference, that work estimates the limiting covariance through a plug-in approach. However, this additional covariance-estimation step negatively offsets the computational benefits of using sketching, since the sketched optimization update requires only $O((d+m)^2)$ operations per iteration, while plug-in covariance estimation incurs the dominant $O((d+m)^3)$ computational cost. Subsequent work \cite{Du2025Online} complements this result by establishing asymptotic normality for the averaged AI-SSQP iterates and developing a random-scaling inference method \citep{Lee2022Fast} that bypasses covariance estimation. The resulting self-normalized statistic has a pivotal limiting distribution and retains an $O((d+m)^2)$ computational cost for both optimization and inference with the sketched AI-SSQP update.

Although the aforementioned literature provides effective inference frameworks for \eqref{eq:opt_problem}, these methods are inherently derivative-based, requiring access to gradient, and often Hessian, information of the objective and constraints. Such requirements can be restrictive in applications where derivatives are unavailable, unreliable, or expensive to compute. For example, in hyperparameter optimization, the objective is often evaluated through complex training procedures, making higher-order derivative information difficult to obtain. Similarly, in PDE-constrained optimization, gradient evaluations may require adjoint solves, which can substantially increase computational cost. To address these limitations, \cite{Na2025Derivative} proposed a Derivative-Free SSQP (DF-SSQP) method, which replaces exact derivative evaluations with stochastic zeroth-order approximations. By leveraging randomized function evaluations, DF-SSQP retains the favorable structure of SSQP while broadening its applicability to black-box and simulation-based settings. For statistical inference, \cite{Na2025Derivative} established asymptotic normality of the last DF-SSQP iterate, with a limiting covariance that reflects the additional~noise~introduced by derivative-free estimation.

In this paper, we extend the DF-SSQP in \cite{Na2025Derivative} by replacing its exact subproblem solver with a randomized sketching solver, leading to SDF-SSQP. This modification reduces the cost of solving each SQP subproblem from $O((d+m)^3)$ to $O((d+m)^2)$, making the method more scalable for large-scale, derivative-free constrained stochastic optimization. To make inference as efficient as optimization, we also move beyond the last-iterate normal approximation and plug-in limiting covariance estimation established in \cite{Na2025Derivative}. Instead, we base inference on averaged primal--dual iterates and construct self-normalized statistics that avoid explicit covariance estimation. In particular, let $\btheta_k\coloneqq (\bx_k,\blambda_k)$ be the SDF-SSQP iterate, $\btheta^\star \coloneqq (\bx^\star,\blambda^\star)$, and $\bar\btheta_t \coloneqq \frac{1}{t}\sum_{k=0}^{t-1}\btheta_k$. We prove that
\begin{equation*}
\sqrt{t}\cdot(\bar\btheta_t-\btheta^\star)
\overset{d}{\longrightarrow}
\cN(\zero,\bSigma^\star_{\rm ave}) \quad\text{ with } \quad
\bSigma^\star_{\rm ave} =
(\nabla^2\cL(\bx^\star,\blambda^\star))^{-1}\bar\bOmega^\star(\nabla^2\cL(\bx^\star,\blambda^\star))^{-1},
\end{equation*}
where $\bar\bOmega^\star$ is the covariance determined by the leading derivative-free martingale noise; see Theorem~\ref{theorem:local_convergence} for its formula. Furthermore, to facilitate practical inference, we strengthen the above CLT to a functional CLT and construct an online random-scaling matrix $V_t$. We show that, for suitable directions $\bw\in\RR^{d+m}$ specified in Theorem~\ref{theorem:random_scaling}, we have
\begin{equation*}
\frac{\sqrt{t}\,\bw^\top(\bar\btheta_t-\btheta^\star)}{\sqrt{\bw^\top V_t\bw}}
\overset{d}{\longrightarrow}
\frac{B_1(1)}{\sqrt{\int_0^1(B_1(u)-uB_1(1))^2\,du}}
\quad \text{with}\quad
V_t \coloneqq
\frac{1}{t^2}\sum_{i=1}^{t}i^2
(\bar\btheta_i-\bar\btheta_t)
(\bar\btheta_i-\bar\btheta_t)^{\top},
\end{equation*}
where $B_1(\cdot)$ denotes a standard one-dimensional Brownian motion. This limiting distribution is pivotal, involving no unknown quantities, and its quantiles are known. Thus, confidence intervals for linear functionals of $(\bx^\star,\blambda^\star)$ can be formed without estimating or inverting $\bSigma^\star_{\rm ave}$. We mention that the matrix $V_t$ can be updated online with $O((d+m)^2)$ storage and operations, matching the cost scale of the sketched optimization step.

\vskip4pt
\textbf{Technical Contributions.} The proposed SDF-SSQP framework makes the following technical contributions.
\begin{enumerate}[label=(\alph*),topsep=0pt]
\setlength\itemsep{0.0em}
\item \textbf{Derivative-free extension of sketched online inference.} 
We extend the sketched random-scaling framework of \cite{Du2025Online} from the derivative-based AI-SSQP to the derivative-free setting. This extension is highly nontrivial since, in this setting, the gradient, constraint Jacobian, and Lagrangian Hessian are no longer available and must be approximated from random zeroth-order function evaluations. These derivative-free estimators are all \textit{biased}, and their errors propagate through both the optimization recursion and the inference procedure. 
To address this issue, we apply a momentum (moving-average) step to progressively reduce their bias along the algorithmic trajectory. Under our derivative-free debiased recursion, we establish global almost-sure convergence of the proposed SDF-SSQP method and prove local asymptotic normality for the averaged primal--dual iterates. In contrast, the analysis in \cite{Du2025Online} without a momentum step relies critically on unbiased gradient and Hessian estimators, as well as exact deterministic constraint Jacobians.

\item \textbf{Functional CLT for algorithms with momentum debiasing.} 
We prove a functional central limit theorem for the averaged primal--dual process, which is the theoretical foundation for constructing random-scaling statistics. This process-level result is stronger than asymptotic normality for averaged iterates, because random scaling depends on the uniform convergence of the entire trajectory of the averaging process rather than only on the convergence of the terminal end-point average. The result extends the functional CLT argument of \cite{Li2023Online} to online algorithms with a momentum step. The extension is nontrivial because the momentum step breaks the basic martingale-difference structure that crucial arguments of existing analyses rely on (cf. Lemma \ref{lemma:online_FCLT}). To our knowledge, such a functional CLT is unknown even for unconstrained derivative-based methods. Nevertheless, we show that this additional recursive averaging does not change the process-level Brownian-motion limit. Our extension is not specific to the proposed SDF-SSQP algorithm and can be useful for other online inference problems where biased (or heavy-tailed) derivative estimators are debiased (or robustified) through momentum~moving~average.

\item \textbf{Computational efficiency without sacrificing asymptotic statistical efficiency.} We show that sketching reduces the per-iteration computational cost from $O((d+m)^3)$ to $O((d+m)^2)$ without yielding a computational--statistical tradeoff asymptotically. This is in contrast to the sketched AI-SSQP setting of \cite{Na2025Statistical, Du2025Online}, where the randomized sketching solver reduces the computational cost of solving the SQP subproblem but also leads to lower statistical efficiency by enlarging the limiting covariance compared to that of the exact solver (the latter~being minimax optimal). Notably, in our method, the momentum step removes the effect of the sketching error, so the limiting covariance in both the averaged asymptotic normality and the functional CLT does not depend on the sketching distribution or the number of sketching steps. This covariance matches the limiting covariance of the last-iterate error of exact DF-SSQP with optimal $1/t$ stepsize \citep{Na2025Derivative}.

\end{enumerate}

\textbf{Organization and notation.}
We begin by introducing the SDF-SSQP method in Section \ref{sec:sketched_df_ssqp}. Section \ref{sec:global_convergence} introduces the assumptions and establishes global almost-sure convergence of the method. Section \ref{sec:local_convergence} analyzes the local statistical behavior and proves the averaged asymptotic normality of the primal--dual iterates. Section \ref{sec:online_inference} develops the functional central limit theorem, the random-scaling statistic, and the resulting online confidence intervals. Numerical experiments and conclusions are provided in Sections \ref{sec:numerical_experiments} and \ref{sec:conclusion}, respectively. All proofs are deferred to the appendices.

We use $\|\cdot\|$ to denote the $\ell_2$-norm for vectors, and the spectral norm for matrices. For a third-order tensor $A$, $\|A\|$ denotes the induced Euclidean operator norm $\|A\| := \sup_{\|u\|=\|v\|=\|w\|=1} |A[u,v,w]|$.
For a matrix $A$, $A^\dagger$ denotes its Moore--Penrose pseudoinverse, $\ker(A)$ denotes its null space, $\operatorname{span}(A)$ denotes the linear span of its columns, $\operatorname{Tr}(A)$ denotes its trace (if $A$ is square), and $\sigma_{\min}(A)$ denotes its smallest singular value. We let $I$ denote the identity matrix and $\zero$ denote the zero vector or matrix, with dimensions clear from the context. For two symmetric matrices $A$ and $B$, $A\succeq B$ means that $A-B$ is positive semidefinite. We use $\mathbf{1}_{\{\cdot\}}$ for the indicator function and $\lfloor\cdot\rfloor$ for the floor function. For a closed convex set $Z$, $\Pi_Z(\cdot)$ denotes the Euclidean projection onto $Z$. For a sequence of compatible matrices $\{A_k\}$,  we use the convention $\prod_{k=i}^{j}A_k=A_jA_{j-1}\cdots A_i $ for $j\geq i$ and the identity for $j<i$. For two sequences $\{a_k, b_k\}$, $a_k=O(b_k)$ ($a_k\lesssim b_k$) means that $|a_k|/|b_k|$ is bounded for all sufficiently large $k$, and $a_k=o(b_k)$ means that $|a_k|/|b_k|\to 0$. For a sequence of random variables $\{X_k\}$ and a positive deterministic sequence $\{a_k\}$, we write $X_k=O_p(a_k)$ if $X_k/a_k$ is bounded in probability, and $X_k=o_p(a_k)$ if $X_k/a_k\to 0$ in probability. 
Throughout, constants denoted by $\kappa$, $\gamma$, or $\Upsilon$, possibly with subscripts, are finite positive deterministic constants independent of the iteration index~unless~otherwise~specified.

\section{Sketched Derivative-Free SSQP}\label{sec:sketched_df_ssqp}

In this section, we introduce the sketched derivative-free stochastic sequential quadratic programming method for solving Problem \eqref{eq:opt_problem}. The method simply bridges the derivative-free estimation framework of DF-SSQP in \cite{Na2025Derivative} with the sketching-based inexact subproblem solver developed for AI-SSQP in \cite{Na2025Statistical}.

\subsection{Debiased derivatives via momentum}

Given the $k$-th iterate $\bx_k$, we draw a sample $\xi_k\sim \cP$ and two independent random perturbation directions $\bDelta_k, \tilde{\bDelta}_k \in \RR^d$. Let $\cP_{\bDelta}$ denote the distribution of the random directions. Throughout the paper, we assume that $\bDelta\sim \cP_{\bDelta}$ has mutually independent components, each symmetrically distributed about zero with absolute values bounded above and below away from zero.

\vskip4pt
\noindent$\bullet$ \textbf{Gradient and Jacobian Estimates.} Let $\{b_k, \beta_k\}$ be predefined positive sequences. We approximate the objective gradient $\nabla F(\bx_k; \xi_k)\in \RR^d$ and the constraint Jacobian $G_k := \nabla c(\bx_k)\in \RR^{m\times d}$ by $\hat{\nabla}F(\bx_k; \xi_k)$ and $\hat{\nabla}c(\bx_k)$, respectively, defined as follows:
\begin{equation}\label{eq:random_approx}
\begin{aligned}
\hat{\nabla} F(\bx_k;\xi_k) &= \frac{F(\bx_k + b_k\bDelta_k; \xi_k) - F(\bx_k - b_k\bDelta_k; \xi_k)}{2b_k}\bDelta_k^{-1} \in \RR^d,\\
\hat{\nabla} c(\bx_k) &= \frac{c(\bx_k + b_k\bDelta_k) - c(\bx_k - b_k\bDelta_k)}{2b_k}\bDelta_k^{-\top} \in \RR^{m\times d},
\end{aligned}
\end{equation}
where $\bDelta_k^{-1} := (\frac{1}{\bDelta_k^1},\cdots, \frac{1}{\bDelta_k^d}) \in \RR^d$ denotes the entrywise reciprocal of $\bDelta_k = (\bDelta_k^1,\cdots, \bDelta_k^d)$. The random perturbation directions and finite differences used in \eqref{eq:random_approx} introduce bias, which creates technical difficulties in the analysis and can slow convergence. To mitigate this bias, we apply an online momentum moving-average step to the past estimates:
\begin{equation}\label{eq:gradient_approx}
\bar{\bg}_k = (1-\beta_k)\bar{\bg}_{k-1} + \beta_k\hat{\nabla}F(\bx_k;\xi_k) \quad\quad \text{and} \quad\quad \bar{G}_k = (1-\beta_k) \bar{G}_{k-1} + \beta_k\hat{\nabla} c(\bx_k).
\end{equation}

\vskip4pt
\noindent$\bullet$ \textbf{Hessian Estimates.}
To estimate the objective and constraint Hessians, let $\{\tilde{b}_k\}$ be another predefined positive sequence. We first compute the following auxiliary gradient estimates:
\begin{equation*}
\begin{aligned}
\tilde{\nabla} F(\bx_k \pm b_k\bDelta_k; \xi_k) &= \frac{F(\bx_k \pm b_k\bDelta_k + \tilde{b}_k\tilde{\bDelta}_k; \xi_k) - F(\bx_k \pm b_k\bDelta_k ; \xi_k)}{\tilde{b}_k}\tilde{\bDelta}_k^{-1}\in \RR^d,\\
\tilde{\nabla}c(\bx_k\pm b_k\bDelta_k) &= \frac{c(\bx_k\pm b_k\bDelta_k + \tilde{b}_k\tilde{\bDelta}_k) - c(\bx_k\pm b_k\bDelta_k)}{\tilde{b}_k}\tilde{\bDelta}_k^{-\top}\in \RR^{m\times d}.
\end{aligned}
\end{equation*}
We use $\tilde{\nabla}$ to distinguish these one-sided finite-difference estimates from the centered estimates denoted by $\hat{\nabla}$ in \eqref{eq:random_approx}. This construction reduces the number of function evaluations, since $F(\bx_k\pm b_k\bDelta_k;\xi_k)$ and $c(\bx_k\pm b_k\bDelta_k)$ have already been computed for the gradient and Jacobian estimates. We then estimate the Hessians as
\begin{equation}\label{eq:Hessian_approx}
\begin{aligned}
\hat{\nabla}^2F(\bx_k;\xi_k) &= \frac{1}{2}\sbr{\frac{\delta\tilde{\nabla}F(\bx_k\pm b_k\bDelta_k;\xi_k)}{2b_k}\bDelta_k^{-\top} + \bDelta_k^{-1}\frac{\{\delta\tilde{\nabla}F(\bx_k\pm b_k\bDelta_k;\xi_k)\}^{\top}}{2b_k}},\\
\hat{\nabla}^2 c^j(\bx_k) &= \frac{1}{2}\sbr{\frac{\delta\tilde{\nabla}c^j(\bx_k\pm b_k\bDelta_k)}{2b_k}\bDelta_k^{-\top} + \bDelta_k^{-1}\frac{\{\delta\tilde{\nabla}c^j(\bx_k\pm b_k\bDelta_k)\}^\top}{2b_k}}, \quad \text{for } 1\le j\le m,
\end{aligned}
\end{equation}
where 
\begin{equation*}
\begin{aligned}
\delta\tilde{\nabla}F(\bx_k\pm b_k\bDelta_k;\xi_k) &= \tilde{\nabla}F(\bx_k+ b_k\bDelta_k;\xi_k) - \tilde{\nabla}F(\bx_k-b_k\bDelta_k;\xi_k)\in \RR^d,\\
\delta\tilde{\nabla}c^j(\bx_k\pm b_k\bDelta_k) &= \tilde{\nabla}c^j(\bx_k + b_k\bDelta_k) - \tilde{\nabla}c^j(\bx_k - b_k\bDelta_k)\in \RR^d, 
\end{aligned}
\end{equation*}
where $\tilde{\nabla}c^j$ denotes the transpose of the $j$-th row of $\tilde{\nabla}c$. We will also perform a momentum update for the Hessian estimates, as in \eqref{eq:gradient_approx}; however, we defer its introduction to the next subsection since this~momentum update targets the Lagrangian Hessian that involves the dual variables.

\subsection{Algorithm design}

Recall that the Lagrangian function is defined as $\cL(\bx,\blambda) = f(\bx) + \blambda^\top c(\bx)$, where $\blambda\in \RR^{m}$ denotes the dual multipliers. A necessary condition for $(\bx^\star, \blambda^\star)$ to be a local solution of Problem \eqref{eq:opt_problem} is to~satisfy the Karush--Kuhn--Tucker (KKT) conditions:
\begin{equation*}
\nabla \cL(\bx^\star,\blambda^\star) = \binom{\nabla_{\bx}\cL(\bx^\star,\blambda^\star)}{\nabla_{\blambda}\cL(\bx^\star,\blambda^\star)} = \binom{\nabla f(\bx^\star) + G(\bx^\star)^\top \blambda^\star}{c(\bx^\star)} = \zero.
\end{equation*}
The SSQP method can be viewed as an application of Newton's method to the equation $\nabla \cL(\bx,\blambda) = \zero$. Each iteration consists of three steps: estimating the required derivatives, computing an inexact Newton direction via sketching, and updating the primal--dual iterates with an adaptive (potentially random) stepsize. The method requires prespecified positive sequences $\{b_k, \tilde{b}_k,\alpha_k,\beta_k\}$ and parameters $s\in\{1,2,\ldots\}$, $\psi\ge 0$, and $p\ge 1$. It is initialized at $(\bx_0, \blambda_0) \in \RR^d\times \RR^m$, $\bar{\bg}_{-1}\in \RR^d$, $\bar{G}_{-1}\in\mR^{m\times d}$, and $\bar{B}_{-1} = I\in \RR^{d\times d}$.

Given $(\bx_k,\blambda_k)$ at the $k$-th iteration, we obtain the gradient and Jacobian estimators $\bar{\bg}_k$ and $\bar{G}_k$ from \eqref{eq:gradient_approx}, and compute the Hessian estimators $\hat{\nabla}^2F(\bx_k;\xi_k)$ and $\{\hat{\nabla}^2c^j(\bx_k)\}_{j=1}^m$ from \eqref{eq:Hessian_approx}. We regularize the Jacobian estimator $\bar{G}_k$ by setting $\tilde{G}_k = \bar{G}_k + \delta_k^{G}$, where $\delta_k^G \in \RR^{m\times d}$ is a perturbation matrix chosen so that $\tilde{G}_k$ has full row rank if $\bar{G}_k$ does not. With this regularized Jacobian, we compute~the debiased estimates of the Lagrangian gradient and Hessian with respect to $\bx$:
\begin{equation}\label{eq:debiased_Lagrangian_Hessian}
\bar{\nabla}_{\bx}\cL_k = \bar{\bg}_k + \tilde{G}_k^\top \blambda_k,\;\; \hat{\nabla}_{\bx}^2\cL_k = \hat{\nabla}^2F(\bx_k;\xi_k) + \sum_{j=1}^{m}\blambda_k^j\hat{\nabla}^2c^j(\bx_k),\;\; \bar{B}_k = (1-\beta_k)\bar{B}_{k-1} + \beta_k\hat{\nabla}_{\bx}^2\cL_k. 
\end{equation}
Here, $\bar{B}_k$ denotes the debiased estimate of the Lagrangian Hessian with respect to $\bx$. If only global almost-sure convergence (cf. Section \ref{sec:global_convergence}) is of interest, one may simply set $\bar{B}_k = I$. In fact, the Hessian momentum weight is less critical than the gradient momentum weight: we use the same weight $\beta_k$, but uniform averaging with $\beta_k = 1/(k+1)$ can also be used in practice \citep{Na2022Hessian}. To ensure that the Newton system~presented below is well defined, we also regularize the Hessian estimator $\bar{B}_k$ by setting $\tilde{B}_k = \bar{B}_k + \delta_k^B$, where~$\delta_k^B\in \RR^{d\times d}$ is a perturbation matrix chosen so that $\tilde{B}_k$ is positive definite on the null space $\ker(\tilde{G}_k)$ when $\bar{B}_k$~is~not.

With the above derivative approximations, we then solve the following quadratic program to obtain the update direction:
\begin{equation}\label{eq:sqp_subproblem}
\begin{aligned}
\min_{\tilde\Delta\bx_k\in\RR^d}\quad
& \frac{1}{2}\tilde\Delta\bx_k^\top \tilde{B}_k\tilde\Delta\bx_k + \bar{\nabla}_{\bx}\cL_k^\top \tilde\Delta\bx_k \\
\text{s.t.}\quad & c_k+\tilde{G}_k\tilde\Delta\bx_k =\zero.
\end{aligned}
\end{equation}
We slightly abuse the notation to let $\tilde{\Delta}\bx_k$ be the primal solution of \eqref{eq:sqp_subproblem} and $\tilde{\Delta}\blambda_k$ be the associated Lagrange multiplier. In fact, solving \eqref{eq:sqp_subproblem} is equivalent to solving the following Newton system
\begin{equation}\label{equ:KKT}
\underbrace{\begin{pmatrix}
\tilde{B}_k &\tilde{G}_k^\top\\
\tilde{G}_k &\zero
\end{pmatrix}}_{\tilde{W}_k}
\underbrace{\begin{pmatrix}
\tilde{\Delta}\bx_k\\
\tilde{\Delta}\blambda_k
\end{pmatrix}}_{\tilde{\bz}_k}
=
-\underbrace{\begin{pmatrix}
\bar{\nabla}_{\bx}\cL_k\\
c_k
\end{pmatrix}}_{\bar{\nabla}\cL_k}.
\end{equation}
The perturbations of $(\tilde{G}_k, \tilde{B}_k)$ from $(\bar{G}_k, \bar{B}_k)$ ensure that \eqref{eq:sqp_subproblem} has a unique solution, i.e., $\tilde{W}_k$ is invertible. Instead of solving the above Newton system exactly, we solve it inexactly using an iterative sketching solver. In particular, we generate a random sketching matrix $S \in \RR^{(d+m)\times q}$, whose column dimension $q\ge 1$ may also be random. We set $\bz_{k,0} = \zero$. For each $j \ge 0$, we compute
\begin{equation}\label{eq:sketching_update}
\bz_{k,j+1} = \argmin_{\bz} \nbr{\bz - \bz_{k,j}}^2 \quad\quad \text{s.t.}\quad\quad S^\top_{k,j}\tilde{W}_k\bz = -S_{k,j}^\top \bar{\nabla}\cL_k,
\end{equation}
where the matrices $S_{k,j}\sim S$ are independent and identically distributed over $j$ and are independent of $\{\xi_{0:k},(\bDelta_{0:k}, \tilde{\bDelta}_{0:k}),\baralpha_{0:k-1}\}$ (here, $\baralpha_{0:k-1}$ is the (random) stepsize sequence introduced in \eqref{eq:stepsize_selection}). The closed-form recursion of \eqref{eq:sketching_update} is
\begin{equation}\label{eq:sketching_recursion}
\bz_{k,j+1} = \bz_{k,j} - \tilde{W}_kS_{k,j}(S_{k,j}^\top\tilde{W}^2_kS_{k,j})^\dagger S_{k,j}^\top(\tilde{W}_k\bz_{k,j} + \bar{\nabla}\cL_k), \quad\quad j= 0,\ldots,s-1,
\end{equation}
where $(\cdot)^\dagger$ denotes the Moore--Penrose pseudoinverse. 

We run the sketching solver for $s$ steps and~finally update the iterate as
\begin{equation}\label{eq:update}
(\bx_{k+1},\blambda_{k+1}) = (\bx_k, \blambda_k) + \bar{\alpha}_k(\bar{\Delta}\bx_k, \bar{\Delta}\blambda_k),
\end{equation}
where $(\bar{\Delta}\bx_k, \bar{\Delta}\blambda_k):= \bz_{k,s}$ and $\bar{\alpha}_k$ is \textit{any} adaptive stepsize satisfying
\begin{equation}\label{eq:stepsize_selection}
\mu_k \alpha_k\le \bar{\alpha}_k \le \mu_k\alpha_k + \psi\alpha_k^p.
\end{equation}
Here, $\mu_k$ is a random sequence that depends on the Lipschitz parameters of the objective and constraints and stabilizes after an almost-surely finite random index; that is, $\mu_k = \mu_{K^\star}$ for all $k\geq K^\star$ for some random $K^\star<\infty$ that varies across runs. The stepsize condition \eqref{eq:stepsize_selection} generalizes \cite{Na2025Statistical, Du2025Online} from $\mu_k=1$ to a random $\mu_k$. This extension is motivated by practical numerical designs of SSQP (see \citet[Lemma 3.6]{Berahas2021Sequential} and \citet[(26)]{Curtis2024Stochastic} for concrete examples), which perform a line search to select the stepsize and can all be summarized precisely in the form of \eqref{eq:stepsize_selection} under the same condition on $\mu_k$. For our inference study, we do not require any~specific stepsize selection schemes beyond condition \eqref{eq:stepsize_selection}.

\begin{remark}\label{rem:1}

We note that the computational cost of the sketching solver depends on the sketching dimension $q$, the sketching distribution $S$, and the number of sketching steps $s$. Under the standard coverage condition that the expected projection matrix generated by one sketching step captures every direction with strength at least $\gamma_S\in(0,1]$ (see Assumption \ref{assumption:sketching}), we can show that one sketching step contracts the conditional mean-squared linear-system error by a factor of at most $\rho=1-\gamma_S$. For many commonly used sketching distributions, such as Gaussian sketches, Rademacher sketches, and randomized Kaczmarz-type coordinate sketches, one has $\gamma_S$ proportional to the relative sketching~dimension, $\gamma_S \gtrsim q/(d+m)$, up to factors of condition number of $\tilde{W}_k$.

By the above relation, for a fixed relative mean-squared error tolerance $\delta\in(0,1)$, the condition $\rho^s=(1-\gamma_S)^s\le\delta$ is satisfied whenever $s\ge \log(1/\delta)/\log(1/(1-\gamma_S))=O(1/\gamma_S)=O((d+m)/q)$. Thus, for a fixed relative reduction in the conditional mean-squared sketching error, the number of sketching steps $s$ is inversely proportional to the relative sketching dimension: a larger sketching dimension improves the contraction rate and therefore reduces the required number of sketching steps, but it also increases the cost of each sketching step.

Using randomized Kaczmarz-type sketches as an example, where the sketch selects $q$ canonical basis directions in $\mR^{d+m}$, the main cost in one sketching step comes from computing the small $q\times q$ matrix $S^\top\widetilde W_k^2S$, which costs $O((d+m)q^2)$. Combining this per-step cost with $s=O((d+m)/q)$ gives the total cost $O((d+m)^2q)$ for meeting the fixed relative mean-squared error tolerance. This suggests that, in this Kaczmarz-type setting, a smaller sketching dimension $q$ (e.g., sketching vector with $q=1$) is computationally preferable with a total cost of $O((d+m)^2)$. 

That said, this analysis reflects only a worst-case cost estimate. In practice, the optimal sketching dimension is often case-specific and depends on the sketching distribution as well as on structural properties of $\widetilde W_k$, such as sparsity, conditioning, or eigenvalue decay. For more detailed computational tradeoffs of iterative sketching methods, see \cite{Strohmer2008Randomized, Gower2015Randomized, Derezinski2024Sharp, Na2025Statistical, Kuang2026Online} and literature~therein.
\end{remark}

\section{Global Convergence Analysis}\label{sec:global_convergence}

In this section, we establish the global almost-sure convergence of the SDF-SSQP method. We first~state the assumptions used in the analysis, then present the derivative approximation bounds, and finally~combine these bounds with a merit-function reduction to establish the convergence of the method.

\subsection{Global assumptions}

We introduce below, in sequence, the regularity conditions and the conditions on the sampling, perturbation directions, and sketching distributions. All of these conditions are quite standard in the related literature.

\begin{assumption}\label{assumption:boundedness_1}
We assume that there exists an open convex set $\cX\subseteq\RR^d$ containing the sequences $\{\bx_k, \bx_k\pm b_k\bDelta_k, \bx_k\pm b_k\bDelta_k+\tilde b_k\tilde{\bDelta}_k\}_{k\geq 0}$. On $\cX$, the functions $f$ and $\{c^j\}_{j=1}^m$ are thrice differentiable with bounded first, second, and third derivatives, $f$ is bounded below by $f_{\inf}$, and $\|c(\bx)\| \le \kappa_c$ for some $\kappa_c>0$. Moreover, for some constants $\Upsilon_{\blambda}, \kappa_{1,G},\kappa_{2,G},\kappa_{1,\tilde G},\kappa_{2,\tilde G}, \kappa_{1,\tilde B},\kappa_{2,\tilde B}>0$, we have for all~$k\ge0$,
\begin{equation*}
\nbr{\blambda_k}\le\Upsilon_{\blambda},\qquad \kappa_{1,G}I\preceq G_kG_k^\top\preceq \kappa_{2,G}I,\qquad
\kappa_{1,\tilde G}I\preceq \tilde G_k\tilde G_k^\top\preceq \kappa_{2,\tilde G}I,
\end{equation*}
and $\tilde B_k$ is symmetric with $\|\tilde B_k\|\le \kappa_{2,\tilde B}$ and $\bx^\top\tilde B_k\bx\ge \kappa_{1,\tilde B}\nbr{\bx}^2$ for any $\bx\in\ker(\tilde G_k)$. 
\end{assumption}

By the bounded derivatives condition in Assumption \ref{assumption:boundedness_1}, there exists a constant $\Upsilon_u>0$ such that, for all $\bx\in\mathcal X$ and $1\le j\le m$,
\begin{equation}\label{equ:Ups_u}
\max\{ \|\nabla f(\bx)\|,\, \|\nabla^2 f(\bx)\|,\, \|\nabla^3 f(\bx)\|,\, \|\nabla c^j(\bx)\|,\, \|\nabla^2 c^j(\bx)\|,\, \|\nabla^3 c^j(\bx)\|\} \le \Upsilon_u .	
\end{equation}
Assumption \ref{assumption:boundedness_1} states regularity conditions on the trajectory generated by the algorithm: the iterates remain in a region where the objective, constraints, and regularized SQP systems are smooth and well conditioned. The boundedness of $f$, $c$, and their derivatives, full row rank of $G_k$ and $\tilde{G}_k$, and curvature requirements on $\tilde{B}_k$ are standard in SSQP and derivative-free stochastic approximation analyses \citep{Bertsekas1999Nonlinear,Nocedal2006Numerical,Berahas2021Sequential,Curtis2024Stochastic,Fang2024Fully}. Note that these regularity conditions ensure the KKT system \eqref{equ:KKT} is well defined at each step. The regularization perturbations $(\delta_k^G, \delta_k^B)$ are finite-sample safeguards: when $(\bx_k, \blambda_k)$ converges to a local solution satisfying the regularity conditions, these perturbations become inactive (cf. Lemmas~\ref{lemma:gradient_approx_convergence} and \ref{lemma:Hessian_approx_convergence}). We emphasize that the regularity conditions are \textit{necessary} for studying inference for both~the offline $M$-estimator \eqref{equ:SAA} and the online stochastic approximation estimator \citep{Shapiro2021Lectures, Duchi2021Asymptotic, Davis2024Asymptotic, Gao2025Online}.

\begin{assumption}[Sampling conditions]\label{assumption:boundedness_2}
The samples $\{\xi_k\}_{k\ge0}$ are i.i.d. drawn from $\cP$. For any $\bx\in\mX$, we assume $\EE\sbr{F(\bx;\xi)\given \bx}=f(\bx)$. Furthermore, $F(\cdot;\xi)$ is continuously differentiable almost surely and satisfies for any $\bx\in\mX$, 
\begin{equation*}
\EE\sbr{\nabla F(\bx;\xi)\given \bx}=\nabla f(\bx)\quad\quad\text{and}\quad\quad \EE\sbr{\nbr{\nabla F(\bx;\xi)-\nabla f(\bx)}^r\given \bx}\le \Upsilon_m
\end{equation*}
for some $r, \Upsilon_m>0$.
\end{assumption}

Assumption \ref{assumption:boundedness_2} describes the stochastic zeroth-order oracle. Although the algorithm queries only function values, the gradient-noise moment condition is a standard way to control finite-difference approximation error through Taylor expansion \citep{Kiefer1952Stochastic,Blum1954Multidimensional,Spall1992Multivariate,Spall2000Adaptive,Spall2003Introduction,Bhatnagar2013Stochastic}. The bounded $r$-moment condition is weaker than assuming uniformly bounded gradient noise or sub-Gaussian tails in \cite{Na2025Derivative}.

\begin{assumption}[Random direction conditions]\label{assumption:boundedness_3}
The pairs $\{(\bDelta_k,\tilde{\bDelta}_k)\}_{k\ge0}$ are i.i.d. and independent of the sampling and sketching randomness. For any $\bDelta\sim\cP_{\bDelta}$, its entries are mutually independent, symmetrically distributed about zero, and satisfy $\kappa_{\bDelta_1}\le \abr{\bDelta^j}\le \kappa_{\bDelta_2}$ for all $1\le j\le d$, where $0<\kappa_{\bDelta_1}\le\kappa_{\bDelta_2}<\infty$.
\end{assumption}

Assumption \ref{assumption:boundedness_3} is the standard condition on the random perturbation directions imposed in the aforementioned Kiefer--Wolfowitz and derivative-free literature. The centered difference ensures the cancellation of even-order terms in the Taylor expansion, while the upper and lower bounds control the perturbation magnitudes. Note that the lower bound is important because the estimators (cf.~\eqref{eq:random_approx} and \eqref{eq:Hessian_approx}) involve division by the perturbation directions. The simple independent Rademacher directions satisfy this assumption.

To present the sketching condition, we introduce the filtration notation. In particular, we let $\mF_{k} = \sigma(\{\xi_i,(\bDelta_i, \tilde{\bDelta}_i),\{S_{i,j}\}_{j=0}^{s-1}, \baralpha_i\}_{i=0}^{k})$, $k\geq 0$, be the $\sigma$-algebra containing all the randomness up to and including the $k$th iteration, while $\mF_{-1}$ is the trivial $\sigma$-algebra. We also let 
\begin{align*}
\mF_{k-2/3} & = \sigma(\{\xi_i,(\bDelta_i, \tilde{\bDelta}_i),\{S_{i,j}\}_{j=0}^{s-1}, \baralpha_i\}_{i=0}^{k-1}\cup \{\xi_k, (\bDelta_k, \tilde{\bDelta}_k)\}),\\
\mF_{k-1/3} & = \sigma(\{\xi_i,(\bDelta_i, \tilde{\bDelta}_i),\{S_{i,j}\}_{j=0}^{s-1}, \baralpha_i\}_{i=0}^{k-1}\cup \{\xi_k, (\bDelta_k, \tilde{\bDelta}_k), \{S_{k,j}\}_{j=0}^{s-1}\})
\end{align*}
be the intermediate filtrations by adding the sample and direction randomness on top of $\bx_k$ and additionally adding the sketching randomness, respectively. Clearly, we have $\cF_{k-1}\subseteq \cF_{k-2/3}\subseteq \cF_{k-1/3}\subseteq \cF_k$.
The regularization perturbations $\delta_k^G$ and $\delta_k^B$ are $\cF_{k-2/3}$-measurable.
We note that the stepsize $\baralpha_k$ is selected at the end of the $k$th iteration, after the sketched direction $(\bar{\Delta}\bx_k, \bar{\Delta}\blambda_k)$ is computed; thus, it may depend on all the randomness in $\cF_{k-1/3}$, including the current sample $\xi_k$, the perturbation directions $(\bDelta_k, \tilde{\bDelta}_k)$, and the sketching matrices $\{S_{k,j}\}_{j=0}^{s-1}$. We only require that $\baralpha_k$ does not use the randomness of future iterations, so that $(\xi_{k+1}, \bDelta_{k+1}, \tilde{\bDelta}_{k+1})$ is independent of $\cF_k$ for any $k\geq 0$.

\begin{assumption}[Sketching conditions]\label{assumption:sketching}
For every $k\geq 0$, conditional on $\cF_{k-2/3}$, the sketching matrices $S_{k,0},\ldots,S_{k,s-1}$ are independent draws from the law of $S$. Furthermore, the sketching distribution $S\in\RR^{(d+m)\times q}$ satisfies the following two conditions:
\begin{enumerate}[label=(\alph*),topsep=0pt]
\setlength\itemsep{0.0em}
\item For any $k\geq 0$, $\EE[\tilde W_kS(S^\top \tilde W_k^2S)^\dagger S^\top \tilde W_k \mid \cF_{k-2/3}]\succeq \gamma_S I$ for some $\gamma_S>0$, where the expectation is taken only with respect to the randomness of $S$.
\item $\EE[\|S\|\|S^\dagger\|]\le \Upsilon_S$ for some $\Upsilon_S>0$.
\end{enumerate}
\end{assumption}

The condition (a) in Assumption \ref{assumption:sketching} is a coverage condition on the sketching distribution used in the randomized Newton system solver. Conditional on the current Newton system, the projection matrix $\widetilde W_kS(S^\top\widetilde W_k^2S)^\dagger S^\top\widetilde W_k$ represents the part of the Newton system captured by the sketched subspace of $\widetilde W_kS$. The lower bound on its conditional expectation requires this sketched subspace to contain information from every direction of the primal--dual space with nonnegligible probability. In other words, the sketching distribution should not systematically miss any fixed direction of the Newton system. Such a condition is standard in analyses of randomized sketching methods, where it is used to show that repeated sketching steps linearly contract the error between the sketched Newton direction and the exact Newton direction \citep{Pilanci2016Iterative,RoostaKhorasani2018Sub,Luo2016Efficient,Wang2017Sketched}. More specifically, \cite{Na2025Statistical} verified this lower bound condition for randomized Kaczmarz-type sketches, where $S\sim \text{Uniform}(\{\be_i\}_{i=1}^{d+m})$ and $\be_i$ is the $i$th canonical basis vector of $\mR^{d+m}$. The condition also holds for widely used Gaussian sketches \citep{Derezinski2024Sharp}. See Remark \ref{rem:1} for the discussion of $\gamma_S$ and its dependence~on the~sketching~dimension~$q$.

The condition (b) in Assumption \ref{assumption:sketching} is also standard in asymptotic analyses of sketched Newton methods; see, e.g., \citet[Assumption 5.3]{Na2025Statistical} and \citet[Assumption 3.3]{Du2025Online}. This assumption is weaker than \citet[Assumption 3.4]{Kuang2026Online} that imposes a bounded second moment on the condition number of $S$, $\mE[\|S\|^2\|S^\dagger\|^2]$. This assumption is trivially satisfied when $q=1$, i.e., when employing sketching vectors, and is used to analyze the difference $\|\tilde W_kS(S^\top \tilde W_k^2S)^\dagger S^\top \tilde W_k - W^\star S(S^\top (W^\star)^2S)^\dagger S^\top W^\star\|$ between the two projection matrices.

\begin{assumption}\label{assumption:stepsize_selection}
The random sequence $\{\mu_k\}_{k\ge0}$ in \eqref{eq:stepsize_selection} satisfies three conditions almost surely (i.e., in each run of the method). 
\begin{enumerate}[label=(\alph*),topsep=0pt]
\setlength\itemsep{0.0em}
\item $\{\mu_k\}_{k\ge0}\subseteq[\kappa_{1,\mu}, \kappa_{2,\mu}]\subseteq(0,\infty)$. 
\item There exists a finite (potentially random) index $K^\star<\infty$ such that $\mu_k=\mu_{K^\star}$ for all $k\ge K^\star$. 
\item There exists a deterministic finite set $\{\zeta_1,\ldots,\zeta_L\}$ such that $\mu_{K^\star}\in \{\zeta_1,\ldots,\zeta_L\}$.
\end{enumerate}
\end{assumption}

Assumption \ref{assumption:stepsize_selection} abstracts a safeguarded adaptive stepsize rule from the existing literature. In particular, existing SSQP designs (see, e.g., \cite{Berahas2021Sequential, Curtis2024Stochastic, Na2025Derivative}) specify $\mu_k=\nu_k/(\tau_k\kappa_{\nabla f}+\kappa_{\nabla c})$, where $\tau_k$ is a merit parameter, $\nu_k$ is a ratio parameter, and $(\kappa_{\nabla f}, \kappa_{\nabla c})$ are the estimated Lipschitz constants of the gradient $\nabla f$ and the Jacobian $\nabla c$. In their designs, $(\nu_k, \tau_k)$ can take only a finite number of discrete values that are also uniformly bounded, and both stabilize almost surely. Thus, Assumption \ref{assumption:stepsize_selection} naturally holds in their designs. Since our main focus is the online statistical inference analysis of existing designs rather than the design of new numerical methods, we abstract their stepsize selection conditions into Assumption \ref{assumption:stepsize_selection}, thereby allowing any stepsize~selection rule as long as Assumption \ref{assumption:stepsize_selection} holds for \eqref{eq:stepsize_selection}. We note that \cite{Na2025Statistical} also considered abstracting the stepsize condition; however, their setup of $\mu_k=1$ cannot precisely cover existing numerical designs. We also mention that conditions (b) and (c) of Assumption~\ref{assumption:stepsize_selection} are only required~in~local analysis. In global analysis, it suffices to only assume (a).

To end this subsection, throughout the paper we set the algorithmic sequences as the polynomials
\begin{equation}\label{eq:sequence}
\alpha_k=\frac{\iota_1}{(k+1)^{p_1}},\qquad \beta_k=\frac{\iota_2}{(k+1)^{p_2}},\qquad b_k=\frac{\iota_3}{(k+1)^{p_3}},\qquad \tilde b_k=\frac{\iota_4}{(k+1)^{p_4}}
\end{equation}
with positive constants $\iota_i, p_i>0$ for $i=1,2,3,4$, and we additionally require $0<\beta_k\leq 1$ for all $k\geq 0$, since $\beta_k$ is the momentum weight sequence.

\subsection{Guarantees for derivative approximations}

In this subsection, we present some results on the derivative-free estimators introduced in Section \ref{sec:sketched_df_ssqp}. The first lemma quantifies the bias of the randomized finite-difference approximations. The next two lemmas show that the momentum moving-average estimators track the exact gradient and Jacobian both almost surely and in mean square. These results provide the estimation bounds needed in the global convergence analysis.

\begin{lemma}[\cite{Na2025Derivative}, Lemma 3.5]\label{lemma:approximation_rate}
Under Assumptions~\ref{assumption:boundedness_1}--\ref{assumption:boundedness_3}, we have for $1\le j \le m$,
\begin{align*}
\EE[\hat{\nabla}F(\bx_k;\xi_k) - \nabla f_k\given \cF_{k-1}] & = O(b_k^2), &
\EE[\hat{\nabla}c_k - G_k\given \cF_{k-1}] & = O(b_k^2), \\
\EE[\hat{\nabla}^2 F(\bx_k;\xi_k) - \nabla^2 f_k\given \cF_{k-1}] & = O(b_k + \tilde{b}_k^2/b_k), &\EE[\hat{\nabla}^2c_k^j - \nabla^2c_k^j\given \cF_{k-1}] &= O(b_k + \tilde{b}_k^2/b_k).
\end{align*}	
\end{lemma}

Lemma~\ref{lemma:approximation_rate} shows that the finite-difference gradient and Jacobian estimators have a bias of order $O(b_k^2)$, while the finite-difference Hessian estimators have a larger bias of order $O(b_k+\tilde b_k^2/b_k)$.

\begin{lemma}\label{lemma:gradient_approx_convergence}
Under Assumptions \ref{assumption:boundedness_1}--\ref{assumption:boundedness_3}, \ref{assumption:stepsize_selection}, suppose the sequences $(\alpha_k,\beta_k,b_k,\tilde{b}_k)$ in \eqref{eq:sequence} are specified with parameters satisfying
\begin{equation}
p_2\in (0.5, 1], \qquad p_1 > p_2, \qquad p_3 > 0.5 - 0.5p_2, \qquad r\ge 2, \qquad r(p_1-p_2) > 1. \label{eq:parameter_conditions:lemma:gradient_approx_convergence}
\end{equation}
Then, we have $\bar{\bg}_k - \nabla f_k \rightarrow \zero$ and $\bar{G}_k - G_k \rightarrow \zero$ as $k\rightarrow\infty$ almost surely. Furthermore, if $\delta_k^G$ ensures $[\kappa_{1,G}, \kappa_{2,G}] \subseteq (\kappa_{1,\tilde{G}}, \kappa_{2,\tilde{G}})$, then there exists an almost-surely finite (potentially random) index $K^\star_G<\infty$ such that for all $k\ge K^\star_G$, $\tilde{G}_k = \bar{G}_k$, i.e., $\delta_k^G = \zero$. 
\end{lemma}

\begin{proof}
See Appendix~\ref{proof_of_lemma:gradient_approx_convergence}.
\end{proof}

Lemma~\ref{lemma:gradient_approx_convergence} shows that the momentum gradient and Jacobian estimators in \eqref{eq:gradient_approx} converge almost surely to their deterministic counterparts. This highlights a key benefit of the momentum mechanism, as such convergence does not hold in its absence. The second statement further shows that the Jacobian regularization is only a transient feature of the algorithm: after a finite, possibly random time, the method operates directly with the original, unregularized Jacobian estimator.

\begin{lemma}\label{lemma:gradient_approx_rate}
Under Assumptions \ref{assumption:boundedness_1}--\ref{assumption:stepsize_selection}, suppose the sequences $(\alpha_k,\beta_k,b_k,\tilde{b}_k)$ in \eqref{eq:sequence} are specified with parameters satisfying
\begin{equation}\label{eq:parameter_conditions:lemma:gradient_approx_rate}
p_2\in(0,1), \qquad p_1>p_2, \qquad r\ge 2. 
\end{equation}
Then, we have 
\begin{equation*}
\max\cbr{\EE[\nbr{\bar{\bg}_k - \nabla f_k}^2],\; \EE[\nbr{\bar{G}_k - G_k}^2] } = O(\beta_k + b_k^4 + \alpha_k^2/\beta_k^2).
\end{equation*}
\end{lemma}

\begin{proof}
See Appendix~\ref{proof_of_lemma:gradient_approx_rate}.
\end{proof}

Lemma~\ref{lemma:gradient_approx_rate} establishes a squared-error bound for the momentum-based estimators. The three terms in the bound correspond to the variance contribution of the martingale difference noise $\hat{\nabla}F(\bx_k;\xi_k) - \EE[\hat{\nabla}F(\bx_k;\xi_k)\given \cF_{k-1}]$, the squared finite-difference bias $\EE[\hat{\nabla}F(\bx_k;\xi_k)\given \cF_{k-1}] - \nabla f_k$, and the tracking error $\nabla f_k - \nabla f_{k-1}$ due to the evolution of the underlying iterate sequence.

\subsection{Global almost-sure convergence}

We now establish the global almost-sure convergence of SDF-SSQP. To capture both feasibility and~optimality residuals, we introduce the augmented Lagrangian merit function, defined for some $\sigma,\nu>0$~as
\begin{equation}
\cL_{\sigma,\nu}(\bx,\blambda) = \cL(\bx,\blambda) + \frac{\sigma}{2}\nbr{c(\bx)}^2 + \frac{\nu}{2}\nbr{\nabla_{\bx}\cL(\bx,\blambda)}^2,\label{eq:cal_L_sigma_nu_definition}
\end{equation}
where the first penalty term penalizes the feasibility error while the second term penalizes the optimality error. 
Our goal is to show that the inexact sketched search direction $\bz_{k,s} = (\bar{\Delta}\bx_k,\bar{\Delta}\blambda_k)$ yields a sufficient decrease in this merit function, up to higher-order error terms. By establishing that these error terms are summable, we can then show that the merit function decreases sufficiently over the iterate~trajectory. This accumulated decrease subsequently implies the convergence of the KKT residuals.

The first lemma shows that the exact Newton direction $\tilde{\bz}_k = (\widetilde\Delta \bx_k,\widetilde\Delta\blambda_k)$ (cf. \eqref{equ:KKT}) is a descent direction for $\cL_{\sigma,\nu}$, up to the mismatch between the exact and derivative-free estimators. We denote~$\cL_{\sigma,\nu}^k := \cL_{\sigma,\nu}(\bx_k,\blambda_k)$ for simplicity, and similarly, we have $\nabla \cL_{\sigma, \nu}^k = (\nabla_{\bx}\cL_{\sigma,\nu}^k, \nabla_{\blambda}\cL_{\sigma,\nu}^k)$~and~$\nabla\cL_k = (\nabla_{\bx}\cL_k, \nabla_{\blambda}\cL_k) = (\nabla_{\bx}\cL_k, c_k)$.

\begin{lemma}\label{lemma:global_convergence}
Under Assumption~\ref{assumption:boundedness_1}, there exist constants $\sigma, \nu, \gamma_1, \Upsilon_1 > 0$ such that
\begin{equation*}
(\nabla \cL_{\sigma, \nu}^k)^\top \tilde{\bz}_k \le -\gamma_1(\|\tilde{\bz}_k\|^2 + \|\nabla \cL_k\|^2) + \Upsilon_1(\|\nabla f_k - \bar{\bg}_k\| + \|G_k - \tilde{G}_k\| + \|\nabla f_k - \bar{\bg}_k\|^2+ \|G_k - \tilde{G}_k\|^2),
\end{equation*}
where we recall that $\bar{\bg}_k$ and $\bar{G}_k$ are the momentum gradient and Jacobian estimates in \eqref{eq:gradient_approx}, and $\tilde{G}_k = \bar{G}_k+\delta_k^G$ is the perturbed Jacobian.

\end{lemma}

\begin{proof}
See Appendix~\ref{proof_of_lemma:global_convergence}.
\end{proof}

Lemma~\ref{lemma:global_convergence} shows that the directional derivative of the
merit function along $\tilde{\bz}_k$ is bounded above, at least, by a negative multiple of the squared KKT residual $\|\nabla \cL_k\|^2$, up to errors caused by stochastic derivative
estimation. If the gradient and Jacobian estimates were exact, then the bound would reduce to
$(\nabla \cL_{\sigma,\nu}^k)^\top \tilde \bz_k \le -\gamma_1\|\nabla \cL_k\|^2$, which gives strict descent unless the KKT residual vanishes. Thus, $\|\nabla \cL_k\|^2$ serves as a squared residual that measures progress toward optimality. The remaining terms, involving $\|\nabla f_k-\bar \bg_k\|$ and $\|G_k-\tilde G_k\|$, arise from using momentum-based gradient and Jacobian estimates in place of their exact counterparts. Among these perturbation terms, the first-order errors are dominant, while the quadratic errors are higher-order and become negligible once the estimation errors are sufficiently small. Therefore, the negative squared-residual term provides the main descent mechanism, while the perturbation terms are treated as controlled errors in the subsequent convergence~analysis.

To establish convergence of the actual algorithm, we need to transfer the above descent property of the exact direction to the inexact sketched direction $\bz_{k,s} = (\bar{\Delta}\bx_k, \bar{\Delta}\blambda_k)$. The next lemma gives a one-step expected reduction of the merit function.

\begin{lemma}\label{lemma2:global_convergence}
Under Assumptions~\ref{assumption:boundedness_1}--\ref{assumption:stepsize_selection}, there exist deterministic positive constants $(s', K', \gamma_2, \Upsilon_2)$ such that if the number of sketching steps satisfies $s\geq s'$, then for any $k\ge K'$,
\begin{align*}
\mathbb E[\mathcal L_{\sigma,\nu}^{k+1}\mid \mathcal F_{k-2/3}] &\le \mathcal L_{\sigma,\nu}^{k} -\gamma_2\alpha_k \left( \|\widetilde \bz_k\|^2+\|\nabla\mathcal L_k\|^2 \right)\\
&\quad +\Upsilon_2\alpha_k(\|\nabla f_k - \bar{\bg}_k\| + \|G_k - \tilde{G}_k\|+ \|\nabla f_k - \bar{\bg}_k\|^2 + \|G_k - \tilde{G}_k\|^2).
\end{align*}
\end{lemma}

\begin{proof}
See Appendix \ref{proof_of_lemma2:global_convergence}.
\end{proof}

Lemma~\ref{lemma2:global_convergence} shows that the descent property of the exact direction is preserved by the sketched direction, provided that the number of sketching steps $s$ is sufficiently large. This lemma also highlights the benefit of using momentum moving-average gradient and Jacobian estimates. In particular, without the moving-average update in \eqref{eq:gradient_approx}, the estimation errors $\|\nabla f_k-\hat{\nabla}F(\bx_k;\xi_k)\|$ and $\|G_k-\hat{\nabla}c(\bx_k)\|$ have constant-level variance (cf. the proof of Lemma~\ref{lemma:gradient_approx_convergence}). Consequently, their accumulated effect, even after being multiplied by the stepsize $\alpha_k$, is not summable and can offset the accumulated reduction of the merit function. In contrast, the moving-average update makes the variances of these two errors vanish over time. After multiplication by $\alpha_k$, the resulting perturbation becomes summable, and~therefore these errors do not destroy the expected descent.

Based on the above one-step expected reduction of the merit function, the following theorem sums this reduction over the trajectory and shows that the primal--dual increments vanish and the~KKT residual converges to zero almost surely.

\begin{theorem}[Global almost-sure convergence]\label{thm:global_convergence}
Under Assumptions \ref{assumption:boundedness_1}--\ref{assumption:stepsize_selection}, suppose the sequences $(\alpha_k,\beta_k,b_k,\tilde{b}_k)$ in \eqref{eq:sequence} are specified with parameters satisfying
\begin{equation}\label{eq:parameter_conditions:thm:global_convergence}
p_1\in (0.75,1],\qquad p_2\in (0.5,2p_1-1),\qquad p_3 > 0.5 - 0.5p_2, \qquad r(p_1-p_2)>1. 
\end{equation}
If the number of sketching steps satisfies $s\geq s'$ and $\delta_k^G$ ensures $[\kappa_{1,G}, \kappa_{2,G}] \subseteq (\kappa_{1,\tilde{G}}, \kappa_{2,\tilde{G}})$, then we have $\|(\bx_{k+1},\blambda_{k+1})-(\bx_k,\blambda_k)\|\to 0$ and $\|\nabla \mL_k\| = \|(\nabla_{\bx}\mL_k, c_k)\|\to 0$ as $k\to\infty$ almost surely.
\end{theorem}

\begin{proof}
See Appendix~\ref{proof_of_thm:global_convergence}.
\end{proof}

We note that the parameter conditions in \eqref{eq:parameter_conditions:thm:global_convergence} imply those in \eqref{eq:parameter_conditions:lemma:gradient_approx_convergence} and \eqref{eq:parameter_conditions:lemma:gradient_approx_rate}. In particular, the conditions $p_1\in (0.75, 1]$ and $p_2\in (0.5,2p_1-1)$ together imply that $p_2\in (0.5,1)$, which is stronger than both $p_2\in (0.5,1]$ in \eqref{eq:parameter_conditions:lemma:gradient_approx_convergence} and $p_2\in (0,1)$ in \eqref{eq:parameter_conditions:lemma:gradient_approx_rate}. Also, since $p_1 \le 1$, we have $2p_1 - 1 \le p_1$. Then, $p_2 < 2p_1 - 1 \le p_1$, so the two conditions also imply the condition $p_1 > p_2$ in both \eqref{eq:parameter_conditions:lemma:gradient_approx_convergence} and \eqref{eq:parameter_conditions:lemma:gradient_approx_rate}. Since $p_1\in (0.75,1]$ and $p_2 \in (0.5,1)$, it is also clear that $p_1 - p_2 < 0.5$. Then the last condition $r(p_1-p_2) > 1$ in \eqref{eq:parameter_conditions:thm:global_convergence} implies $r\ge 2$ in both \eqref{eq:parameter_conditions:lemma:gradient_approx_convergence} and \eqref{eq:parameter_conditions:lemma:gradient_approx_rate}. It is trivial to see that the feasible region~for~our parameters $\{p_1, p_2, p_3, r\}$ is nonempty.

Theorem \ref{thm:global_convergence} establishes the global convergence of SDF-SSQP in the usual optimization sense: the primal--dual iterates become asymptotically stationary, and the KKT residual converges to zero almost surely. The condition on the number of sketching steps $s$ quantifies the approximation accuracy of the randomized linear system solver. In particular, $s$ must be sufficiently large to ensure that the descent property established for the exact Newton direction in Lemma \ref{lemma:global_convergence} is preserved for the inexact Newton direction.

\section{Local Asymptotic Normality}\label{sec:local_convergence}

We now complement the global convergence analysis by studying the local distributional behavior of SDF-SSQP around the limiting primal--dual solution. This section has two goals. First, we establish~local convergence rates for the primal--dual errors and the derivative-free approximation errors. Second, we use these rates to isolate the leading stochastic term in the averaged primal--dual recursion and establish asymptotic normality for the averaged SDF-SSQP iterates. This result provides the foundation for the online inference procedure developed in Section \ref{sec:online_inference}.

\subsection{Local assumptions and convergence rates}

The local analysis requires the following assumption.

\begin{assumption}\label{assumption:LICQ_SOSC}
We assume that $(\bx_k,\blambda_k)$ converges almost surely to a deterministic primal--dual KKT pair $(\bx^\star,\blambda^\star)$ with $\bx^\star\in\cX$ satisfying the following conditions.
\begin{enumerate}[label=(\alph*),topsep=0pt]
\setlength\itemsep{0.0em}
\item Linear Independence Constraint Qualification (LICQ): $G^\star=\nabla c(\bx^\star)\in\mR^{m\times d}$ has full row rank.
\item Second-Order Sufficient Condition (SOSC): We assume $\bx^\top \nabla_{\bx}^2\cL(\bx^\star,\blambda^\star)\bx \ge \kappa_{1,B}\|\bx\|^2$ for any $\bx\in \{\bx\in\RR^d:G^\star \bx=\zero\}$ and $\|\nabla_{\bx}^2\cL^\star\|\le \kappa_{2,B}$ for some constants $\kappa_{1,B},\kappa_{2,B}>0$.
\end{enumerate}
\end{assumption}

Assumption \ref{assumption:LICQ_SOSC} is the standard local regularity condition for constrained nonlinear optimization. The full-row-rank condition ensures local identifiability of the constraint manifold and uniqueness of the associated multiplier $\blambda^\star$, while the second-order curvature condition ensures that $(\bx^\star,\blambda^\star)$ is a strict local solution. These are the same conditions that underlie the local analysis of offline constrained $M$-estimators, online projection-based stochastic approximation, as well as recent online inference frameworks based on SSQP methods \citep{Duchi2021Asymptotic,Davis2024Asymptotic,Na2025Statistical,Na2025Derivative}.

\begin{assumption}\label{assumption:Hessian_bound}

We assume that there exist a small radius $\delta_{\rm loc}>0$ and a constant $\Upsilon_n>0$ such that $F(\cdot;\xi)$ is twice continuously differentiable on $\cX_{\delta_{\rm loc}}\coloneqq \cX\cap \{\bx\in\RR^d:\|\bx-\bx^\star\|\le \delta_{\rm loc}\}$ almost surely and, for every $\bx\in \cX_{\delta_{\rm loc}}$,
\begin{equation*}
\EE[\|\nabla^2 F(\bx;\xi)-\nabla^2 f(\bx)\|^2\given \bx]\le \Upsilon_n.
\end{equation*}
Moreover, we let $L(\xi):=\sup_{\bx\in \cX_{\delta_{\rm loc}}}\|\nabla^2 F(\bx;\xi)\|$ and assume $\EE[ L^2(\xi)]\le \Upsilon_n$.
\end{assumption}

For notational simplicity, we use the same neighborhood radius $\delta_{\rm loc}$ and upper-bound constant $\Upsilon_n$ in the second part of the statement. Assumption \ref{assumption:Hessian_bound} is a local moment condition on the sample Hessian. Since the global convergence analysis does not require precise Hessian information, this assumption is imposed only near the target solution. Its role is to ensure that the averaged Hessian estimator remains stable in a small neighborhood of $(\bx^\star,\blambda^\star)$, thereby enabling a valid local linearization of the primal--dual recursion and controlling the higher-order remainder terms that arise in the asymptotic analysis. Similar local Hessian moment conditions are used in online covariance estimation and second-order inference analyses \citep{Chen2020Statistical, Na2025Statistical, Du2025Online}.

The next lemma complements Lemmas \ref{lemma:gradient_approx_convergence} and \ref{lemma:gradient_approx_rate} by showing that the momentum Hessian estimator converges locally to the limiting Lagrangian Hessian.

\begin{lemma}\label{lemma:Hessian_approx_convergence}

Under Assumptions \ref{assumption:boundedness_1}--\ref{assumption:stepsize_selection}, \ref{assumption:LICQ_SOSC} and \ref{assumption:Hessian_bound}, we suppose $\delta_k^G$ ensures $[\kappa_{1,G},\kappa_{2,G}] \subseteq (\kappa_{1,\tilde G},\kappa_{2,\tilde G})$ and the sequences $(\alpha_k,\beta_k,b_k,\tilde{b}_k)$ in \eqref{eq:sequence} are specified with parameters satisfying
\begin{equation}
p_1\in(0.5,1],\qquad p_2\in(0.5,p_1),\qquad p_3>0.5-0.5p_2,\qquad p_4>0.5p_3,\qquad r(p_1-p_2)>1.\label{eq:parameter_conditions:lemma:Hessian_approx_convergence}
\end{equation}
Then, $\bar B_k  \to \nabla_{\bx}^2\cL^\star$ as $k\to\infty$ almost surely. Furthermore, if $\delta_k^B$ ensures $[\kappa_{1,B},\kappa_{2,B}] \subseteq (\kappa_{1,\tilde B},\kappa_{2,\tilde B})$, then there exists an almost surely finite (potentially random) index $K_B^\star<\infty$ such that for all $k\geq K_B^\star$, $\tilde B_k=\bar B_k$, i.e., $\delta_k^B=\0$.
\end{lemma}

\begin{proof}
See Appendix \ref{proof_of_lemma:Hessian_approx_convergence}.
\end{proof}

Lemma \ref{lemma:Hessian_approx_convergence} shows that the Hessian perturbation is also only transient for the algorithm, like the Jacobian perturbation. Once the momentum Hessian estimator becomes sufficiently accurate, the regularization term $\delta_k^B$ becomes inactive, and the local behavior of the algorithm is governed~by~the unregularized averaged Hessian estimator.

To state the local convergence rate, we recall that $\btheta_k=(\bx_k,\blambda_k)$ and $\btheta^\star=(\bx^\star,\blambda^\star)$. Let $W^\star := \nabla^2\mL(\bx^\star, \blambda^\star) = \begin{pmatrix}
\nabla_{\bx}^2\cL^\star & (G^\star)^\top\\
G^\star & 0
\end{pmatrix}$ be the Lagrangian Hessian at $(\bx^\star,\blambda^\star)$, which is also the Jacobian~of the KKT mapping at $(\bx^\star,\blambda^\star)$. Then, for any $\epsilon>0$ and integer $N\geq 0$, define the stopping time
\begin{equation*}
\begin{aligned}
\tau_N(\epsilon):= \inf\Bigg\{\, k\ge N:\ &  \|\btheta_k-\btheta^\star\|>\epsilon^2
\; \mathrm{OR}\;
\|\tilde W_k^{-1}\|>1/\epsilon
\;\mathrm{OR}\;
\|\nabla \cL_k-\tilde W_k(\btheta_k-\btheta^\star)\|> \epsilon^2\|\btheta_k-\btheta^\star\|
\\
& \mathrm{OR}\; \|\nabla \cL_k\|>\|\btheta_k-\btheta^\star\|/\epsilon \;\mathrm{OR}\;
\delta_k^G\neq \zero \;\mathrm{OR}\; \delta_k^B\neq \zero \;\mathrm{OR}\;
\|\btheta_k\|>1/\epsilon \\
&\mathrm{OR}\; \nbr{\nabla \cL_k-W^\star(\btheta_k-\btheta^\star)}>\|\btheta_k-\btheta^\star\|^2/\epsilon\Bigg\}.
\end{aligned}		
\end{equation*}
The stopping time localizes the analysis to a neighborhood in which the perturbed Newton system is well conditioned, the local linearization of the KKT mapping is accurate, and both the Jacobian and Hessian perturbations are inactive. Specifically, the stopping time is defined as the first iteration at which the iterate exits this neighborhood. In fact, under the conditions of Lemma~\ref{lemma:Hessian_approx_convergence} (including the one on $\delta_k^B$), we show in Appendix~\ref{proof_of_thm:local_convergence} that, for any sufficiently small $\epsilon>0$, we have almost surely $\tau_N(\epsilon)=\infty$ for all sufficiently large $N$; consequently, $\PP(\tau_N(\epsilon)<\infty)\to 0$ as $N\to\infty$. In other words, for every run of the method, there exists a small neighborhood such that, once the iterates enter it at a sufficiently large time index, they remain there forever.

The next lemma gives the local rate of the primal--dual error and the momentum gradient approximation error.

\begin{lemma}\label{lemma:local_convergence}

Under Assumptions \ref{assumption:boundedness_1}--\ref{assumption:stepsize_selection}, suppose the sequences $(\alpha_k,\beta_k,b_k,\tilde{b}_k)$ in \eqref{eq:sequence} are specified with parameters satisfying
\begin{equation}\label{eq:parameter_conditions:lemma:local_convergence}
p_1\in(0,1),\qquad p_2\in(0,p_1),\qquad r\ge 2.
\end{equation}
Then, for any $\epsilon$ satisfying $0 < \epsilon < \kappa_{1,\mu}(1-\rho^s)/(16(\kappa_{2,\mu} + \psi\iota_1^{p-1}))$ with $\rho = 1-\gamma_S$, there exists $N_0<\infty$ such that for every $N\ge N_0$ and every $k\ge N$,
\begin{equation*}
\EE\sbr{\|\btheta_k-\btheta^\star\|^2\1_{\tau_N(\epsilon)>k}} + \EE\sbr{\|\bar{\bg}_k - \nabla f_k\|^2\1_{\tau_N(\epsilon)>k}} + \EE\sbr{\|\bar{G}_k - G_k\|^2\1_{\tau_N(\epsilon)>k}} \le \Upsilon_3(N)(\beta_k+b_k^4),
\end{equation*}
where $\Upsilon_3(N)>0$ is a constant that may depend on $N$.
\end{lemma}

\begin{proof}
See Appendix \ref{proof_of_lemma:local_convergence}.
\end{proof}

Lemma \ref{lemma:local_convergence} shows that, on the localized event $\{\tau_N(\epsilon)>k\}$, the primal--dual error and the gradient estimation error are controlled at the same order. After localization, the global tracking term $\alpha_k^2/\beta_k^2$ in Lemma \ref{lemma:gradient_approx_rate} is absorbed, leaving only the momentum fluctuation $\beta_k$ and the squared derivative-free bias term $b_k^4$. This local mean-square convergence rate is the key basis for the final averaged normality~result. 
As a complementary result, we state the convergence rate of the Hessian estimation~error.

\begin{lemma}\label{lemma:W_convergence_rate}

Under Assumptions \ref{assumption:boundedness_1}--\ref{assumption:stepsize_selection}, \ref{assumption:LICQ_SOSC} and \ref{assumption:Hessian_bound}, suppose the sequences $(\alpha_k,\beta_k,b_k,\tilde{b}_k)$ in \eqref{eq:sequence} are specified with parameters satisfying
\begin{equation}\label{eq:parameter_conditions:lemma:W_convergence_rate}
p_1\in(0,1),\qquad p_2\in(0,p_1),\qquad r\ge 2, \qquad p_4 > 0.5p_3.
\end{equation}
Then, for any $\epsilon$ satisfying the condition in Lemma~\ref{lemma:local_convergence} and $\epsilon^2 \le \delta_{\rm loc}/2$, there exists a natural number $N_0^\prime$ such that for every $N\ge N_0^\prime$ and every $k\ge N$, we have
\begin{equation*}
\EE\sbr{\|\tilde W_k-W^\star\|^2\1_{\tau_N(\epsilon)>k}} \le \Upsilon_4(N)(\beta_k+b_k^2+\tilde b_k^4/b_k^2),
\end{equation*}
where $\Upsilon_4(N)>0$ is a constant that may depend on $N$.
\end{lemma}

\begin{proof}
See Appendix \ref{proof_of_lemma:W_convergence_rate}.
\end{proof}

The lemmas in this subsection establish the fundamental local convergence rates required for the asymptotic normality analysis in the next subsection. They provide precise control of the primal--dual iterate error, as well as the gradient and Hessian estimation errors, in a neighborhood of the solution $\btheta^\star=(\bx^\star,\blambda^\star)$. These bounds are used repeatedly throughout the analysis to show that various remainder terms are asymptotically negligible and therefore do not contribute to the limiting distribution in the next subsection.

\subsection{Asymptotic normality of averaged SDF-SSQP}

We now turn to the averaged primal--dual iterate error
\begin{equation*}
\bar\btheta_t-\btheta^\star = \frac{1}{t}\sum_{k=0}^{t-1}(\btheta_k-\btheta^\star) = \frac{1}{t}\sum_{k=0}^{t-1}(\bx_k-\bx^\star,\blambda_k-\blambda^\star). 
\end{equation*}
The next theorem states that this averaged error is asymptotically Gaussian.

\begin{theorem}[Asymptotic normality of averaged SDF-SSQP]\label{theorem:local_convergence}

Under Assumptions \ref{assumption:boundedness_1}--\ref{assumption:stepsize_selection}, \ref{assumption:LICQ_SOSC}, and \ref{assumption:Hessian_bound}, we suppose $\delta_k^G$ ensures $[\kappa_{1,G},\kappa_{2,G}] \subseteq (\kappa_{1,\tilde G},\kappa_{2,\tilde G})$, $\delta_k^B$ ensures $[\kappa_{1,B},\kappa_{2,B}] \subseteq (\kappa_{1,\tilde B},\kappa_{2,\tilde B})$, and the sequences $(\alpha_k,\beta_k,b_k,\tilde{b}_k)$ in \eqref{eq:sequence} are specified with parameters satisfying
\begin{equation}\label{equ:cond4.6}
\begin{gathered}
p_1\in(0.75,1),\qquad p_2\in(0.5,2p_1-1), \qquad p_3>0.25, \qquad p_4>0.5p_3 + 0.25(1-p_2),\\
p > 1+\frac{1-p_2}{2p_1}, \qquad r(p_1-p_2)>1.
\end{gathered}
\end{equation}
Then, we have
\begin{equation*}
\sqrt{t}\cdot(\bar\btheta_t-\btheta^\star) \overset{d}{\longrightarrow} \cN(\zero,\bSigma^\star_{\rm ave}) \quad\quad \text{with}\quad \bSigma^\star_{\rm ave} := (W^\star)^{-1} \bar\bOmega^\star (W^\star)^{-1},
\end{equation*}
where
\begin{equation}\label{equ:Omega}
\bOmega^\star := \EE_{\bDelta\sim \cP_{\bDelta}} \sbr{\bDelta^{-1}\bDelta^\top \Cov_{\xi\sim \cP}(\nabla F(\bx^\star;\xi)) \bDelta\bDelta^{-\top}} \qquad\text{and}\qquad \bar\bOmega^\star := \text{\diag}\big(\bOmega^\star, \0\big).
\end{equation}

\end{theorem}

\begin{proof}
See Appendix \ref{proof_of_thm:local_convergence}.
\end{proof}

We note that the parameter conditions \eqref{equ:cond4.6} imply those in \eqref{eq:parameter_conditions:lemma:Hessian_approx_convergence}, \eqref{eq:parameter_conditions:lemma:local_convergence} and \eqref{eq:parameter_conditions:lemma:W_convergence_rate}. In particular, the conditions $p_1\in (0.75, 1)$ and $p_2\in (0.5,2p_1-1)$ imply $p_2\in (0.5,p_1)$. Since $p_2>0.5$, the condition $p_3>0.25$ implies $p_3>0.5-0.5p_2$. Moreover, since $p_1\in (0.75,1)$ and $p_2 \in (0.5,1)$, we have $p_1 - p_2 < 0.5$. Therefore the condition $r(p_1-p_2) > 1$ implies $r\ge 2$. Finally, we note that the feasible region for~our parameters $\{p_1, p_2, p_3, p_4, p, r\}$ is nonempty.

Theorem \ref{theorem:local_convergence} establishes the asymptotic normality of the averaged SDF-SSQP method. With Polyak--Ruppert averaging, the primal--dual error converges to a Gaussian limit under $\sqrt{t}$ scaling, with asymptotic covariance $\bSigma^\star_{\rm ave} = (W^\star)^{-1}\bar\bOmega^\star(W^\star)^{-1}$. 
A key observation is that this covariance depends neither on the sketching matrices nor on any sketching distribution parameters. In fact, it coincides with the asymptotic covariance of the corresponding last-iterate exact-solver DF-SSQP with statistically optimal stepsize $1/t$, established in \cite{Na2025Derivative}. Thus, while sketching reduces the computational cost of approximately solving the Newton system, it does not inflate the asymptotic covariance of the averaged estimator. Consequently, sketching asymptotically introduces \textit{no} statistical--computational tradeoff for the averaged estimator. This stands in sharp contrast to the findings of \cite{Na2025Statistical,Du2025Online, Wang2026Inference}, where sketching was shown to reduce statistical efficiency by enlarging the limiting covariance compared to that of the corresponding exact-solver methods.

This covariance invariance stems from the averaging and momentum-debiasing mechanisms of the method. The local analysis shows that the finite-difference bias, the Jacobian and Hessian perturbations, the sketching error, and the adaptive-stepsize discrepancy are all asymptotically negligible in the averaged recursion. As a result, the only leading-order fluctuation is the martingale noise generated by the stochastic zeroth-order gradient estimator, whose covariance is characterized by $\bar\bOmega^\star$.

On the other hand, \citet[Proposition 4.9]{Na2025Derivative} shows that $\bSigma^\star_{\rm ave}\succeq \bSigma^\star$ (see \eqref{equ:cov}), with the gap to the statistically optimal covariance $\bSigma^\star$, $\|\bSigma^\star_{\rm ave}-\bSigma^\star\|$, growing linearly in the decision dimension $d$. This loss of statistical efficiency relative to the derivative-based optimum is attributable to the derivative-free approximation rather than the sketching solver. Indeed, derivative-free methods rely on randomized perturbations to estimate derivative information using only a dimension-independent number of function evaluations per iteration. Consequently, the computational savings achieved by derivative-free methods come at the expense of a dimension-dependent loss in asymptotic statistical~efficiency.

\section{Online Inference via Random Scaling}\label{sec:online_inference}

In this section, we develop an online inference procedure for the primal--dual solution $\btheta^\star = (\bx^\star,\blambda^\star)$. Our inference procedure is based on a random-scaling technique \citep{Lee2022Fast}, where we construct a proper scaling matrix to studentize the averaged iterate error, thereby bypassing the covariance estimation and deriving a pivotal statistic whose limiting distribution is free of any unknown problem-dependent quantities. To that end, we need a process-level strengthening of Theorem \ref{theorem:local_convergence} from CLT to functional~CLT.

\begin{theorem}[Functional CLT]\label{theorem:online_FCLT}
Under the assumptions of Theorem \ref{theorem:local_convergence}, we have
\begin{equation*}
\frac{1}{\sqrt{t}}\sum_{k=0}^{\lfloor ut\rfloor-1}(\btheta_k-\btheta^\star) \overset{d}{\Longrightarrow} (\bSigma^\star_{\rm ave})^{1/2} B_{d+m}(u),\quad\quad u\in[0,1],
\end{equation*}
where $B_{d+m}(\cdot)$ denotes a standard $(d+m)$-dimensional Brownian motion, and the convergence is weak convergence in $D(\sbr{0,1},\RR^{d+m})$ equipped with the Skorokhod topology.
\end{theorem}

\begin{proof}
See Appendix~\ref{proof_of_theorem:online_FCLT}.
\end{proof}

Theorem \ref{theorem:online_FCLT} strengthens Theorem \ref{theorem:local_convergence} by extending the limit theorem for the terminal average to a functional limit theorem for the entire partial-sum process. This process-level characterization is a key ingredient in random-scaling inference, allowing the fluctuations of the averaged trajectory to be used to construct a data-driven normalization factor, which in turn yields the $t$-like limiting~distribution in Theorem \ref{theorem:random_scaling}.

Before presenting the inference results, we state a technical lemma needed for the proof of Theorem~\ref{theorem:online_FCLT}. The lemma controls a martingale remainder term induced by the momentum-based moving-average recursion. Establishing this result is the main technical step in extending the functional CLT for algorithms without momentum \citep{Li2023Online} to the momentum setting considered here. All notation and parameters, including $(\alpha_k,\beta_k,\mu_k)$, are as defined throughout the paper.

\begin{lemma}\label{lemma:online_FCLT}
Let $\{\mu_k\}_{k\ge 0}$ satisfy Assumption \ref{assumption:stepsize_selection}; let $H\in\RR^{(d+m)\times (d+m)}$ be symmetric positive definite; and let $\{\bD_k\}_{k\ge0}$ be an $\RR^{d+m}$-valued martingale difference sequence with respect to some filtration $\{\mF_k\}_{k\ge0}$ satisfying for some $\varrho>2$, $\sup_{k\ge0}\EE[\nbr{\bD_k}^{\varrho}]<\infty$. Then, if $\alpha_k = \iota_1/(k+1)^{p_1}$~and $\beta_k = \iota_2/(k+1)^{p_2}$ satisfy
\begin{equation*}
0<p_2<p_1<1,
\end{equation*}
we have
\begin{align*}
\max_{0\le a\le t-1}\nbr{\frac{1}{\sqrt t} (\mu_{a+1}\alpha_{a+1}H)^{-1}
\sum_{k=0}^{a}\sum_{i=k}^{a}\prod_{j=i+1}^{a+1}(I-\mu_j\alpha_jH)\mu_i\alpha_iH \prod_{h=k+1}^{i} (1-\beta_h)\beta_k\bD_k} \to0
\end{align*}
in probability as $t\to\infty$.
\end{lemma}

\begin{proof}
See Appendix~\ref{proof_of_lemma:online_FCLT}.
\end{proof}

In the proof of Theorem \ref{theorem:online_FCLT}, we apply Lemma \ref{lemma:online_FCLT} to the martingale difference sequence $\bD_k = -(W^\star)^{-1}(\hat\nabla\cL_k - \EE[\hat\nabla\cL_k\given \cF_{k-1}])$, where $\hat\nabla\cL_k = (\hat{\nabla} F(\bx_k;\xi_k) + \hat{\nabla} c(\bx_k)^\top\blambda_k, c_k)$ is the zeroth-order estimate of the Lagrangian gradient. This sequence is the leading stochastic term determining the limiting covariance in Theorem \ref{theorem:local_convergence}, and is naturally also the leading term driving the partial-sum process to the limiting distribution $(\bSigma^\star_{\rm ave})^{1/2} B_{d+m}(u)$ in Theorem \ref{theorem:online_FCLT}.
Unlike the proof of Theorem \ref{theorem:local_convergence}, where it suffices to show that the remaining terms are negligible at the terminal time, the functional setting requires establishing their negligibility under the Skorokhod topology. Lemma \ref{lemma:online_FCLT} is crucial for this purpose, as it shows that the maximum norm of a key remainder process converges to zero in probability. This uniform control implies that the remainder is asymptotically negligible in the Skorokhod topology and therefore does not affect the limiting process.

With the functional CLT result established in Theorem \ref{theorem:online_FCLT}, we can prove the following theorem, which enables us to do online inference through random scaling.

\begin{theorem}[Online inference via random scaling]\label{theorem:random_scaling}

Under the assumptions of Theorem \ref{theorem:local_convergence}, we~suppose $\bOmega^\star\succ 0$ for $\bOmega^\star$ defined in \eqref{equ:Omega}. For any $\bw = (\bw_{\bx}, \bw_{\blambda})\in\mR^{d+m}$ such that either $\bw_{\bx}\notin \text{span}((G^\star)^\top)$ or $\bw_{\blambda}\neq \0$, we have
\begin{equation}\label{eq:random_scaling_limit}
\frac{\sqrt{t}\bw^\top(\bar\btheta_t-\btheta^\star)}{\sqrt{\bw^\top V_t\bw}}
\overset{d}{\longrightarrow}
\frac{B_1(1)}{\sqrt{\int_0^1(B_1(u)-uB_1(1))^2\,du}}\quad \text{with}\quad V_t :=\frac{1}{t^2}\sum_{i=1}^{t}i^2(\bar\btheta_i-\bar\btheta_t)(\bar\btheta_i-\bar\btheta_t)^{\top},
\end{equation}
where $B_1(\cdot)$ is a standard one-dimensional Brownian motion.
\end{theorem}

\begin{proof}
See Appendix \ref{proof_of_theorem:random_scaling}.
\end{proof}

We first mention that the condition on $\bw$ in the statement is imposed to exclude degenerate inference directions. Specifically, directions that violate the condition, namely $\bw\in\text{span}((G^\star)^\top)\times \{\0\}$, correspond to directions normal to the constraint manifold, along which the averaged primal--dual error exhibits no asymptotic fluctuation. Consequently, the asymptotic variance vanishes and the~random-scaling denominator degenerates. In other words, for the constrained model parameters $\bx^\star$ (i.e., when $\bw_{\blambda}=\0$), inference is~meaningful only along directions $\bw_{\bx}$ with a nonzero tangential component \citep{Shapiro2021Lectures}. The condition $\bOmega^\star\succ 0$ holds, in particular, whenever $\Cov_{\xi\sim \cP}(\nabla F(\bx^\star;\xi))\succ 0$ (see Appendix~\ref{proof_of_theorem:random_scaling}).

Theorem \ref{theorem:random_scaling} shows that the proposed test statistic is asymptotically pivotal, with a limiting distribution free of unknown problem-dependent quantities. In particular, the limiting distribution in \eqref{eq:random_scaling_limit} does not depend on the asymptotic covariance matrix $\bSigma^\star_{\rm ave}$. The quantiles of this limiting distribution are tabulated in \cite{Abadir1997Two}. Theorem \ref{theorem:random_scaling} immediately yields the following \mbox{corollary},~which provides asymptotically valid confidence intervals for linear functionals of $\btheta^\star= (\bx^\star,\blambda^\star)$.

\begin{corollary}\label{corollary:RS_CI}
Under the conditions of Theorem \ref{theorem:random_scaling}, for any $\phi\in(0,1)$,
\begin{equation*}
\PP\rbr{\bw^\top(\bx^\star,\blambda^\star) \in
\big[\bw^\top(\bar{\bx}_t,\bar{\blambda}_t)\pm U_{1-\phi/2} \sqrt{\bw^\top V_t\bw/t}\big]} \to 1-\phi \qquad \text{as } t\to\infty,
\end{equation*}
where $U_{1-\phi/2}$ denotes the $(1-\phi/2)\times 100\%$ quantile of the limiting distribution in \eqref{eq:random_scaling_limit}.
\end{corollary}

Equivalently, to test $H_0: \bw^\top(\bx^\star,\blambda^\star)=a$, we use the statistic
\begin{align*}
T_t(a) := \frac{\sqrt{t}(\bw^\top(\bar{\bx}_t,\bar{\blambda}_t)-a)}{\sqrt{\bw^\top V_t\bw}}\quad\quad \text{which satisfies}\quad\quad T_t(a)
\overset{d}{\longrightarrow}
\frac{B_1(1)}{\sqrt{\int_0^1(B_1(u)-uB_1(1))^2\,du}}
\end{align*}
under $H_0$. Therefore, an asymptotic level-$\phi$ two-sided test rejects $H_0$ whenever $\abr{T_t(a)}>U_{1-\phi/2}$.

We emphasize that the entire inference procedure can be implemented online. The random-scaling matrix $V_t$ is constructed solely from differences between averaged iterates, $\bar\btheta_i-\bar\btheta_t$. As a result, it can be updated recursively using the history of the averaged iterates. Moreover, the inference procedure requires no additional derivative evaluations, covariance estimation, or matrix inversions. Consequently, the additional computational cost of online inference is only $O((d+m)^2)$ per iteration, which is of the same order as the computational cost of the sketched primal--dual update.

To conclude this section, we discuss in the following remark how the proposed inference procedure can be extended to stochastic constrained problems.

\begin{remark}[Extension to stochastic constraints]

The proposed inference procedure can be extended to stochastic constrained problems. Consider the extension of Problem \eqref{eq:opt_problem} in which the constraints are also defined through an expectation:
\begin{equation*}
\min_{\bx\in\mR^d} f(\bx):=\EE[F(\bx;\xi)] \qquad \text{s.t.} \qquad c(\bx):=\EE[C(\bx;\xi)]=\0,
\end{equation*}
where $C(\cdot;\xi): \mR^d\rightarrow \RR^m$. The key distinction from the deterministic constraint setting is that $c(\bx)$ is no longer available. Thus, in addition to estimating the objective value $f(\bx)$ through the sample $F(\bx;\xi)$, we must also estimate the constraint value $c(\bx)$ through the sample $C(\bx;\xi)$. In this setting, the constraint Jacobian estimator will be changed from \eqref{eq:random_approx} to
\begin{equation*}
\hat{\nabla} C(\bx_k;\xi_k) := \frac{C(\bx_k+b_k\bDelta_k;\xi_k)-C(\bx_k-b_k\bDelta_k;\xi_k)}{2b_k}\bDelta_k^{-\top},
\end{equation*}
while the Hessian estimator \eqref{eq:Hessian_approx} will be changed in the same way by replacing $c(\cdot)$ with $C(\cdot;\xi)$.

For such problems, we extend the moving-average update \eqref{eq:gradient_approx} to the constraint value by defining $\bar{c}_k = (1-\beta_k)\bar{c}_{k-1} + \beta_kC(\bx_k;\xi_k)$. 
Under analogous smoothness, moment, LICQ, and SOSC assumptions, the analysis strategy developed in this paper carries over directly. Indeed, the stochastic finite-difference estimators for $G = \nabla c$ still exhibit the same bias and moment orders as those for $\nabla f$ established in Lemma \ref{lemma:approximation_rate}, and the moving-average recursion for $\bar{c}_k$ can be controlled by the same tracking arguments used for $\bar{\bg}_k$ and $\bar{G}_k$ in Lemma \ref{lemma:gradient_approx_rate}; the only difference is that the squared finite-difference bias term $b_k^4$ is absent, as $C(\cdot ;\xi)$ is assumed to be an unbiased estimate of $c$. Finally, we can reasonably expect to show for $\bar\btheta_t = (\bar{\bx}_t,\bar{\blambda}_t)$ that
\begin{equation*}
\sqrt{t}\cdot(\bar\btheta_t-\btheta^\star) \overset{d}{\longrightarrow} \cN(\zero,\bSigma_{{\rm ave},{\rm ec}}^\star) \quad\quad \text{with}\quad \bSigma_{{\rm ave},{\rm ec}}^\star := (W^\star)^{-1} \bar{\bOmega}_{\rm ec}^\star (W^\star)^{-1},
\end{equation*}
where
\begin{equation*}
\bar{\bOmega}_{\rm ec}^\star = \mE_{\bDelta\sim \cP_{\bDelta}} \sbr{
\begin{pmatrix}
\bDelta^{-1}\bDelta^\top & \0\\
\0 & I
\end{pmatrix}\Cov_{\xi\sim \cP}\begin{pmatrix}
\nabla F(\bx^\star;\xi) + \nabla C(\bx^\star;\xi)^\top\blambda^\star\\
C(\bx^\star;\xi)
\end{pmatrix}\begin{pmatrix}
\bDelta\bDelta^{-\top} & \0\\
\0 & I
\end{pmatrix}}.
\end{equation*}
Compared with the covariance $\bar{\bOmega}^\star$ in \eqref{equ:Omega} for deterministic constraints, $\bar{\bOmega}_{\rm ec}^\star$ has the same structural form but incorporates additional randomness from both the stochastic constraint value $C(\bx^\star;\xi)$ and the stochastic constraint Jacobian $\nabla C(\bx^\star;\xi)^\top\blambda^\star$, together with their cross-covariances with the objective gradient noise. With this asymptotic normality result, the functional CLT and the random-scaling inference procedure extend naturally to the stochastic-constraint setting. We defer a rigorous analysis of this extension to future work.
\end{remark}

\section{Numerical Experiments}\label{sec:numerical_experiments}

We apply SDF-SSQP to learn an initialization for training linear and logistic regression models. For each candidate initialization $\bx\in\RR^d$, we perform a fixed number of Riemannian SGD steps, average the updated parameters, and evaluate this average on independent validation data. The outer objective therefore measures the validation performance attained after training from $\bx$.

\subsection{Experimental setup}

The outer decision variable $\bx\in\RR^d$ is the initialization parameter of an inner regression training procedure. We consider two constraints:
\begin{equation*}
c(\bx)=\mathbf{1}_d^\top\bx-1 \quad\text{(linear constraint)},
\qquad
c(\bx)=\bx^\top\bx-1 \quad\text{(unit-sphere constraint)}.
\end{equation*}
For training and validation samples $\xi=(\xi_{\rm tr},\xi_{\rm val})$, let $F(\bx;\xi)$ denote the validation loss obtained after a fixed number of training updates from $\bx$. The objective therefore selects an initialization according to its expected validation performance after the prescribed training budget.

\vskip4pt
\noindent \textbf{Data generation and regression losses.}
We consider dimensions $d\in\{5,20,40\}$. For each dimension and constraint, we first specify the data-generating regression parameter $\bbeta^\star$ by setting, for $j=1,\ldots,d$,
\begin{equation*}
\beta_j^\star= \begin{cases}
j/\sum_{r=1}^{d}r, & \text{linear constraint},\\[2pt]
j/(\sum_{r=1}^{d}r^2)^{1/2}, & \text{unit-sphere constraint}.
\end{cases}
\end{equation*}
Thus, $\bbeta^\star$ satisfies the corresponding constraint. For each observation $(\ba,y)$, we draw the feature vector $\ba\sim\mathcal{N}(\zero,C_d)$, using one of three covariance designs:
\begin{equation*}
C_d= \begin{cases}
I_d, & \text{identity},\\
0.8I_d+0.2T_d, & \text{Toeplitz},\\
0.8I_d+0.2\mathbf{1}_d\mathbf{1}_d^\top, & \text{equi-correlation},
\end{cases}
\qquad (T_d)_{ij}=0.5^{|i-j|}.
\end{equation*}
All three designs have unit marginal variances. The Toeplitz design has correlations that decay with coordinate separation, while the equi-correlation design has correlation $0.2$ between distinct features. Conditional on $\ba$, we generate the response according to the regression model. For linear regression, we use
\begin{equation*}
y=\ba^\top\bbeta^\star+\varepsilon, 
\qquad
\ell(\bu;(\ba,y)) = \frac12(\ba^\top\bu-y)^2,
\end{equation*}
where $\varepsilon=0$ for training and $\varepsilon\sim\mathcal{N}(0,1)$ independently of $\ba$ for validation, and $\ell$ is the loss at a regression parameter $\bu\in\RR^d$. For logistic regression, we use
\begin{equation*}
y\mid\ba\sim\mathrm{Bernoulli}\bigl( (1+e^{-\ba^\top\bbeta^\star})^{-1}\bigr),
\qquad
\ell(\bu;(\ba,y)) =\log(1+e^{\ba^\top\bu})-y\ba^\top\bu.
\end{equation*}
At each outer iteration, we draw $n_{\rm tr}=20$ training observations and $n_{\rm val}=20$ validation observations independently according to their respective data distributions. We write these samples as $\xi_{\rm tr}=(\xi_{{\rm tr},j})_{j=1}^{n_{\rm tr}}$ and $\xi_{\rm val}=(\xi_{{\rm val},i})_{i=1}^{n_{\rm val}}$.

\vskip4pt
\noindent\textbf{Inner training and outer objective.} For the $k$th outer step at $\bx_k$, we initialize $\bu_0$ by projecting $\bx_k$ onto the corresponding constraint set and perform $n_{\rm tr}$ Riemannian SGD steps. In particular, the $j$th update uses the observation $\xi_{{\rm tr},j}$ and stepsize $\eta_j=0.1/j$. Let $\bu_j(\bx_k;\xi_{\rm tr})$ denote the parameter after the $j$th training update. We return the averaged SGD iterate $\mathcal{A}_{n_{\rm tr}}(\bx_k;\xi_{\rm tr}) =\frac{1}{n_{\rm tr}}\sum_{j=1}^{n_{\rm tr}} \bu_j(\bx_k;\xi_{\rm tr})$.

Then, we evaluate the training output on the independent validation sample:
\begin{equation*}
F(\bx_k;\xi)
=\frac{1}{n_{\rm val}}\sum_{i=1}^{n_{\rm val}}
\ell\bigl(\mathcal{A}_{n_{\rm tr}}(\bx_k;\xi_{\rm tr});
\xi_{{\rm val},i}\bigr).
\end{equation*}
This scalar is the sampled objective value supplied to the outer algorithm. Since differentiating $F$ with respect to $\bx$ would require differentiating through the training procedure, SDF-SSQP is particularly useful for this hyperparameter optimization problem, as it instead estimates the required derivatives by perturbing the initialization, repeating the training and validation evaluations, and using the returned objective values. All perturbed evaluations within an outer iteration use the same training and validation samples, as in \eqref{eq:random_approx} and \eqref{eq:Hessian_approx}.

\vskip4pt
\noindent\textbf{Outer optimization and parameter settings.}
We apply SDF-SSQP from Section~\ref{sec:sketched_df_ssqp} with two KKT solvers. The sketched solver performs $s=d$ coordinate projection steps per outer iteration, with coordinates sampled uniformly with replacement, while the exact solver, denoted by $s=\infty$, solves the KKT system exactly. Both solvers use the same derivative-free estimation procedure and parameter choices. For both solvers, we initialize at $\bx_0=\mathbf{1}_d/d$ for the linear constraint and $\bx_0=\mathbf{1}_d/\sqrt{d}$ for the unit-sphere constraint, with $\lambda_0=0$. For the outer stepsize, momentum, and finite-difference sequences in \eqref{eq:sequence}, we set $(p_1,p_2,p_3,p_4)=(0.751,0.501,0.251,0.251)$ and $\iota_1=\iota_3=\iota_4=1$, $\iota_2=0.01$.

For each configuration, we perform 200 independent runs, each with $10^5$ outer iterations. We discard the first 10\% of each trajectory and compute the tail average and random-scaling normalizer from the remaining iterates. Discarding a fixed fraction of the initial iterates does not alter the limiting distribution: by Theorem~\ref{theorem:online_FCLT}, the rescaled partial-sum process of the retained iterates converges weakly to the same Brownian limit, so the statistic in \eqref{eq:random_scaling_limit} computed from the retained iterates has the same pivotal limit by the argument of Theorem~\ref{theorem:random_scaling}. For both KKT solvers, we apply the inference procedure in Section~\ref{sec:online_inference}, using the critical value $6.747$ to construct nominal 95\% coordinatewise confidence intervals (the quantile table is provided in \cite{Abadir1997Two}).

We evaluate estimation accuracy and inferential performance using the data-generating parameter $\bbeta^\star$ as the reference.
In linear regression, noiseless training keeps the inner iterates at
$\bbeta^\star$ when initialized there, so $\bbeta^\star$ is a known
minimizer of the expected validation loss after training. In logistic
regression, we compare the solvers under stochastic training using
independent training and validation samples from the same logistic model.
The data-generating parameter $\bbeta^\star$ minimizes the shared
population logistic loss and serves as the reference for MAE and coverage.
We note that the optimal initialization may differ from $\bbeta^\star$ after finitely many stochastic training steps.
In the logistic experiments, both solvers yield estimated initializations close to the data-generating parameter $\bbeta^\star$, with all reported MAEs below $5\times10^{-3}$ (Table~\ref{tab:sec6-results}).
The mean absolute error (MAE) is computed as the absolute difference between the primal tail average and $\bbeta^\star$, averaged over coordinates and runs. Empirical coverage is the fraction of coordinatewise confidence intervals containing the corresponding coordinate of $\bbeta^\star$, averaged over coordinates and runs. The mean interval length is similarly averaged over coordinates and runs. The FLOP counts include outer updates and random-scaling inference. We exclude data generation and inner SGD updates from these counts to more directly compare the computational costs of the exact and sketched KKT solvers.

\subsection{Numerical results}
\begin{table}[p]
\centering
\captionsetup{textfont=it}
\begingroup
\small
\setlength{\tabcolsep}{2.8pt}
\renewcommand{\arraystretch}{1.00}
\setlength{\arrayrulewidth}{0.35pt}
\setlength{\doublerulesep}{1.2pt}
\begin{tabular}{|c|c|c|c|r|r|r|r|r|r|r|r|}
\hline
\multirow{3}{*}{$d$} & \multirow{3}{*}{\shortstack{Design\\Cov.}} & \multirow{3}{*}{Constraint} & \multirow{3}{*}{$s$} & \multicolumn{4}{c|}{Lin} & \multicolumn{4}{c|}{Logit} \\
\cline{5-12}
& & & & MAE & Cov & Len & \multirow{2}{*}{FLOPs/iter} & MAE & Cov & Len & \multirow{2}{*}{FLOPs/iter} \\
& & & & $(10^{-2})$ & $(\%)$ & $(10^{-2})$ & & $(10^{-2})$ & $(\%)$ & $(10^{-2})$ & \\
\hline
\multirow{12}{*}{5} & \multirow{4}{*}{Identity} & \multirow{2}{*}{Linear} & 5 & 0.33 & {\red 96.80} & {\red 2.12} & 899.83 & 0.19 & {\red 95.80} & {\red 1.30} & 900.50 \\
\cline{4-12}
 &  &  & $\infty$ & 0.18 & 99.60 & 2.07 & 916.83 & 0.10 & 98.80 & 0.92 & 917.59 \\
\cline{3-12}
 &  & \multirow{2}{*}{Sphere} & 5 & 0.32 & 99.80 & 4.20 & 925.83 & 0.34 & 95.20 & 1.94 & 926.50 \\
\cline{4-12}
 &  &  & $\infty$ & 0.23 & 99.80 & 2.69 & 942.83 & 0.10 & {\red 96.70} & {\red 0.92} & 943.64 \\
\cline{2-12}
 & \multirow{4}{*}{Toeplitz} & \multirow{2}{*}{Linear} & 5 & 0.33 & {\red 97.20} & {\red 2.18} & 899.83 & 0.20 & {\red 96.30} & {\red 1.33} & 900.50 \\
\cline{4-12}
 &  &  & $\infty$ & 0.18 & 99.50 & 2.16 & 916.83 & 0.09 & 98.50 & 0.88 & 917.58 \\
\cline{3-12}
 &  & \multirow{2}{*}{Sphere} & 5 & 0.33 & 98.20 & 4.13 & 925.83 & 0.32 & 97.00 & 1.96 & 926.50 \\
\cline{4-12}
 &  &  & $\infty$ & 0.23 & 99.40 & 2.67 & 942.83 & 0.21 & {\red 98.00} & {\red 1.27} & 943.83 \\
\cline{2-12}
 & \multirow{4}{*}{Equi-corr.} & \multirow{2}{*}{Linear} & 5 & 0.33 & {\red 94.30} & {\red 2.14} & 899.83 & 0.20 & {\red 96.90} & {\red 1.27} & 900.50 \\
\cline{4-12}
 &  &  & $\infty$ & 0.19 & 99.50 & 2.24 & 916.83 & 0.09 & 98.60 & 0.83 & 917.59 \\
\cline{3-12}
 &  & \multirow{2}{*}{Sphere} & 5 & 0.32 & 98.80 & 3.88 & 925.83 & 0.31 & 98.00 & 1.98 & 926.50 \\
\cline{4-12}
 &  &  & $\infty$ & 0.19 & 99.30 & 2.69 & 942.83 & 0.08 & {\red 98.00} & {\red 0.76} & 943.64 \\
\hline\hline
\multirow{12}{*}{20} & \multirow{4}{*}{Identity} & \multirow{2}{*}{Linear} & 20 & 0.35 & 91.53 & 2.68 & 9865.77 & 0.22 & 92.60 & 1.87 & 9867.02 \\
\cline{4-12}
 &  &  & $\infty$ & 0.34 & {\red 97.23} & {\red 2.76} & 14178.04 & 0.19 & 98.15 & 1.64 & 14179.33 \\
\cline{3-12}
 &  & \multirow{2}{*}{Sphere} & 20 & 0.50 & {\red 97.60} & {\red 4.26} & 9965.72 & 0.37 & 99.30 & 3.95 & 9966.97 \\
\cline{4-12}
 &  &  & $\infty$ & 0.37 & 98.63 & 4.04 & 14278.01 & 0.38 & 98.85 & 3.65 & 14680.23 \\
\cline{2-12}
 & \multirow{4}{*}{Toeplitz} & \multirow{2}{*}{Linear} & 20 & 0.35 & 91.93 & 2.69 & 9865.76 & 0.22 & {\red 93.38} & {\red 1.93} & 9867.00 \\
\cline{4-12}
 &  &  & $\infty$ & 0.34 & 97.85 & 2.82 & 14178.12 & 0.19 & 98.50 & 1.63 & 14179.22 \\
\cline{3-12}
 &  & \multirow{2}{*}{Sphere} & 20 & 0.51 & {\red 97.53} & {\red 4.01} & 9965.72 & 0.37 & 99.38 & 3.88 & 9966.97 \\
\cline{4-12}
 &  &  & $\infty$ & 0.37 & 98.33 & 3.93 & 14278.02 & 0.47 & 99.30 & 4.23 & 14649.67 \\
\cline{2-12}
 & \multirow{4}{*}{Equi-corr.} & \multirow{2}{*}{Linear} & 20 & 0.42 & 88.38 & 2.74 & 9865.75 & 0.21 & {\red 93.10} & {\red 1.75} & 9867.00 \\
\cline{4-12}
 &  &  & $\infty$ & 0.38 & {\red 97.25} & {\red 3.07} & 14177.99 & 0.17 & 98.05 & 1.53 & 14179.24 \\
\cline{3-12}
 &  & \multirow{2}{*}{Sphere} & 20 & 0.39 & {\red 97.60} & {\red 3.69} & 9965.72 & 0.34 & 99.50 & 3.92 & 9966.97 \\
\cline{4-12}
 &  &  & $\infty$ & 0.39 & 98.50 & 3.84 & 14278.30 & 0.23 & 99.28 & 3.37 & 14279.22 \\
\hline\hline
\multirow{12}{*}{40} & \multirow{4}{*}{Identity} & \multirow{2}{*}{Linear} & 40 & 0.43 & 91.88 & 3.04 & 37126.77 & 0.27 & 88.83 & 1.72 & 37129.12 \\
\cline{4-12}
 &  &  & $\infty$ & 0.47 & 97.43 & 4.05 & 75750.39 & 0.25 & {\red 96.91} & {\red 2.00} & 75752.75 \\
\cline{3-12}
 &  & \multirow{2}{*}{Sphere} & 40 & 0.53 & {\red 96.18} & {\red 4.08} & 37326.72 & 0.35 & {\red 97.86} & {\red 2.99} & 37328.97 \\
\cline{4-12}
 &  &  & $\infty$ & 0.50 & 97.96 & 4.17 & 75950.14 & 0.27 & 98.88 & 2.88 & 75952.54 \\
\cline{2-12}
 & \multirow{4}{*}{Toeplitz} & \multirow{2}{*}{Linear} & 40 & 0.44 & 91.46 & 3.06 & 37126.82 & 0.27 & 89.05 & 1.73 & 37129.12 \\
\cline{4-12}
 &  &  & $\infty$ & 0.44 & 97.53 & 3.64 & 75750.09 & 0.25 & {\red 97.25} & {\red 2.01} & 75752.95 \\
\cline{3-12}
 &  & \multirow{2}{*}{Sphere} & 40 & 0.51 & {\red 96.35} & {\red 3.99} & 37326.72 & 0.35 & {\red 98.01} & {\red 2.93} & 37328.97 \\
\cline{4-12}
 &  &  & $\infty$ & 0.49 & 97.44 & 4.09 & 75952.94 & 0.26 & 98.88 & 2.85 & 75952.14 \\
\cline{2-12}
 & \multirow{4}{*}{Equi-corr.} & \multirow{2}{*}{Linear} & 40 & 0.52 & 89.79 & 3.36 & 37126.87 & 0.25 & 88.46 & 1.56 & 37129.07 \\
\cline{4-12}
 &  &  & $\infty$ & 0.49 & {\red 97.31} & {\red 4.03} & 75750.09 & 0.22 & {\red 96.75} & {\red 1.82} & 75752.39 \\
\cline{3-12}
 &  & \multirow{2}{*}{Sphere} & 40 & 0.61 & {\red 92.50} & {\red 3.69} & 37326.72 & 0.30 & 98.64 & 2.83 & 37328.97 \\
\cline{4-12}
 &  &  & $\infty$ & 0.69 & 94.01 & 3.94 & 75956.60 & 0.23 & 99.04 & 2.72 & 75956.84 \\
\hline
\end{tabular}
\endgroup
\caption{Results for constrained initialization learning. Lin and Logit denote linear and logistic regression; $s=d$ and $s=\infty$ denote the sketched and exact KKT solvers. MAE, Cov (\%) and Len report absolute error relative to $\bbeta^\star$, empirical coverage of the coordinates of $\bbeta^\star$ by nominal 95\% confidence intervals, and full interval length, averaged over coordinates and runs. FLOPs/iter reports the combined cost of outer updates and random-scaling inference, divided by the total number of outer iterations. Within each matched configuration, red highlights favorable coverage--length combinations, prioritizing coverage close to the nominal $95\%$ level and shorter intervals among solvers with comparably good coverage. When coverage and interval length are comparable, we highlight the solver with lower FLOPs/iter.}
\label{tab:sec6-results}
\end{table}

Table~\ref{tab:sec6-results} compares SDF-SSQP with the sketched and exact KKT solvers. We consider constrained initialization learning across two regression models (linear and logistic), three dimensions ($d=5,20,40$), three covariance designs (identity, Toeplitz, and equi-correlation), and two constraints (linear and unit sphere). For each configuration, we report computational cost, MAE, empirical coverage, and interval length based on 200 runs. Based on our experimental results, we observe that the sketched solver can achieve estimation accuracy and inferential performance comparable to those of the exact solver at substantially lower computational cost, particularly in higher-dimensional settings.

\textbf{Computational Efficiency.}
The sketched solver requires fewer FLOPs/iter than the exact solver in every configuration, with a more pronounced advantage as the dimension increases. Including both outer updates and random-scaling inference, the savings relative to the exact solver are approximately $2\%$ at $d=5$, $30$--$32\%$ at $d=20$, and $51\%$ at $d=40$. In particular, at $d=40$, the sketched solver more than halves the computational cost relative to the exact solver, even after the common random-scaling inference cost is included. For example, for linear regression under the identity design and sphere constraint at $d=40$, the costs are $37{,}326.72$ FLOPs/iter for the sketched solver and $75{,}950.14$ FLOPs/iter for the exact solver. These results highlight the computational advantage of replacing the exact KKT solve with sketching, especially in higher-dimensional settings.

\begin{figure}[H]\centering\includegraphics[width=\textwidth]{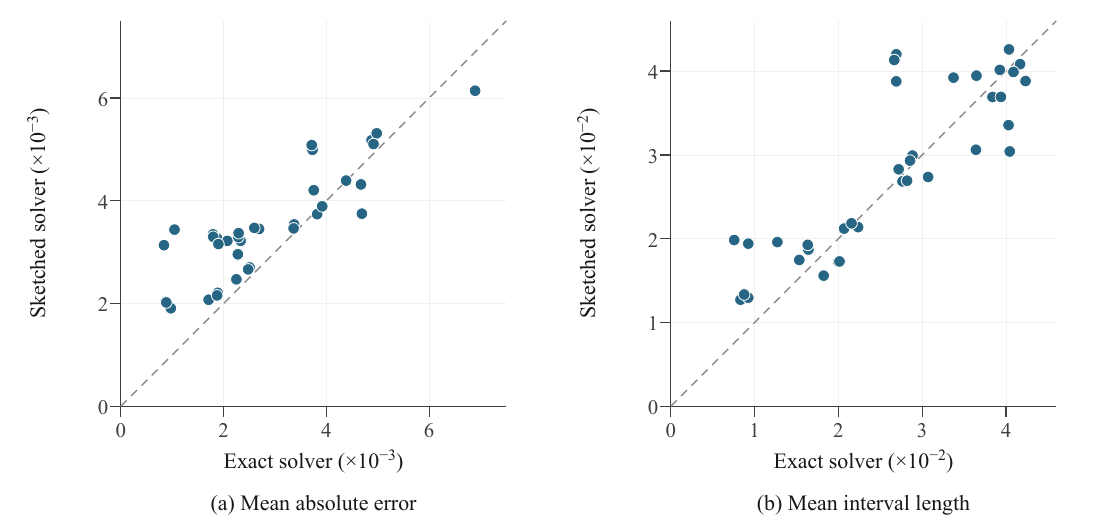}\caption{Comparison of (a) mean absolute error and (b) mean confidence-interval length for the sketched and exact KKT solvers. Each point represents one of the 36 matched configurations, with each statistic averaged over coordinates and 200 runs per solver. All combinations of regression model, dimension, covariance design, and constraint are included. The dashed line indicates equal values for the two solvers.}\label{fig:solver-comparison}\end{figure}

\textbf{Estimation Accuracy.}
The sketched solver exhibits estimation accuracy comparable to that of the exact solver, with MAEs generally on the order of $10^{-3}$ in both regression models, as illustrated in Figure~\ref{fig:solver-comparison}(a). In many configurations, the two solvers yield very similar MAEs. For example, linear regression at $d=40$ under the identity design and sphere constraint yields MAEs of $0.53\times10^{-2}$ and $0.50\times10^{-2}$ for the sketched and exact solvers, respectively. For logistic regression at the same dimension, the Toeplitz design with the linear constraint gives MAEs of $0.27\times10^{-2}$ for the sketched solver and $0.25\times10^{-2}$ for the exact solver.
Taken together, these results show that the lower computational cost of the sketched solver is accompanied by similar estimation errors.

\textbf{Inference and Statistical Efficiency.}
The sketched solver attains empirical coverage of $\bbeta^\star$ close to the nominal $95\%$ level in many configurations across both regression models. For example, its coverage is $96.35\%$ for linear regression under the Toeplitz design and sphere constraint at $d=40$, and $95.20\%$ for logistic regression under the identity design and sphere constraint at $d=5$. The corresponding exact-solver coverages are $97.44\%$ and $96.70\%$, respectively. In some configurations, the empirical coverage of $\bbeta^\star$ departs more markedly from $95\%$. For example, the sketched solver yields coverage of $88.83\%$ for logistic regression under the identity design and linear constraint at~$d=40$, while its coverage is $99.50\%$ under the equi-correlation design and sphere constraint at $d=20$.

Turning to interval length, the sketched solver produces confidence intervals with lengths generally comparable to those of the exact solver in both regression models, as illustrated in Figure~\ref{fig:solver-comparison}(b). In many configurations, the difference in interval length is small relative to the widths of the intervals themselves. For example, for linear regression under the Toeplitz design and sphere constraint at $d=40$, the sketched and exact solvers give interval lengths of $3.99\times10^{-2}$ and $4.09\times10^{-2}$, respectively. Under the same design, constraint, and dimension, the corresponding lengths for logistic regression are $2.93\times10^{-2}$ and $2.85\times10^{-2}$. The similarity in interval length complements the comparable estimation accuracy reported above. This is consistent with the sketching-independent asymptotic covariance characterized in Theorem~\ref{theorem:local_convergence}, which identifies no sketching-induced inflation of the limiting covariance.

\section{Conclusion}\label{sec:conclusion}

In this paper, we developed a sketched derivative-free stochastic SQP (SDF-SSQP) method for nonlinear constrained stochastic optimization. The method combines zeroth-order finite-difference derivative estimators, momentum-based debiasing, and an iterative sketching solver for the primal--dual Newton system. It thus applies to black-box and simulation-based problems in which derivatives are unavailable or prohibitively expensive to obtain, while reducing the per-iteration computational cost from the cubic complexity of a full Newton solve to the quadratic complexity of a sketched update.
We established both optimization and statistical guarantees. On the optimization side, we proved global almost-sure convergence of the primal--dual increments and the KKT residual under standard assumptions. On the statistical side, we proved asymptotic normality of the averaged primal--dual iterates, with a covariance that is independent of sketching components and coincides with that of the last-iterate exact-solver DF-SSQP method with statistically optimal stepsize $1/t$ in \cite{Na2025Derivative}. Consequently, sketching reduces computational cost \textit{without} sacrificing asymptotic statistical efficiency, and any covariance inflation relative to derivative-based methods stems from the derivative-free approximation rather than the sketching solver.
Finally, we developed a fully online inference procedure based on random scaling. The resulting self-normalized statistic is asymptotically pivotal and yields confidence intervals for linear functionals of the primal--dual solution without covariance estimation or matrix inversion, and can be updated recursively at a cost matching the sketched optimization step. Its key technical ingredient is a functional CLT for the averaged primal--dual process, established through a maximal negligibility argument for the moving-average debiasing error. This technique is of independent interest for analyzing other online stochastic~approximation methods with recursive momentum schemes or averaged derivative estimators.

Several directions remain for future work. First, it would be useful to relax the full-row-rank assumption on the constraint Jacobian and extend the theory to rank-deficient settings. Second, extending the theory to nonlinear inequality constraints would require tracking active sets, complementarity, and possible changes in the active constraint manifold under stochastic zeroth-order observations, building on ideas from active-set and inequality-constrained SSQP methods \citep{Na2023Inequality}. Third, the current framework focuses on deterministic constraints and could be extended to settings in which the constraints themselves are subject to uncertainty or depend on stochastic observations. Finally, sharper nonasymptotic bounds and large-scale numerical studies for black-box, simulation-based, and PDE-constrained problems would further clarify the computational~and statistical advantages of the proposed method.

\section*{Disclosure of AI usage}

The methodology, theoretical results, and proofs were developed entirely by the authors before the recent substantial advances in AI mathematical reasoning. All writing was done by the authors. OpenAI's GPT 6.1 Sol and some earlier models were used to proofread the paper and assist with writing code for the simulations.

\bibliographystyle{my-plainnat}
\bibliography{ref}

\appendix
\pagebreak
\numberwithin{equation}{section}
\numberwithin{theorem}{section}

\makeatletter
\renewcommand{\@seccntformat}[1]{%
\ifnum\pdfstrcmp{#1}{section}=0
Appendix \csname the#1\endcsname.\quad
\else
\csname the#1\endcsname\quad
\fi}
\makeatother

\section{Preliminary Lemmas}

This appendix collects several auxiliary results that will be used repeatedly in the proofs. The preliminary lemmas are stated in a self-contained form. Unless explicitly stated otherwise, the notation appearing in a given lemma is local to that lemma and should not be interpreted as referring to the notation introduced in the main paper.

\begin{lemma}[\cite{Ruszczynski1980Feasible}, Lemma~1]\label{pre_lemma:as_convergence}
Let $(\Omega, \cF, \PP)$ be a probability space and let $\{\cF_k\}_{k\geq 0}$ be an increasing sequence of $\sigma$-algebras contained in $\cF$. Let $\{\bm{\eta}_k, \bz_k\}_{k\geq 0}$ be sequences of $\cF_k$-measurable $\RR^d$-valued random variables satisfying the relations
\begin{align*}
&\bz_{k+1} = \Pi_Z((1-\rho_k)\bz_k + \rho_k\bxi_k), \quad \bz_0\in Z,\\
&\EE[\bxi_k\given \cF_k] = \bm{\eta}_k + \bb_k,
\end{align*}
where $\{\rho_k\}_{k\ge 0}$ is a deterministic sequence of positive weights, the set $Z\subseteq \RR^d$ is convex and closed, and $\Pi_Z(\cdot)$ is the projection onto the set $Z$. Suppose the following conditions hold:
\begin{enumerate}[label=(\alph*),topsep=0pt]
\setlength\itemsep{0.0em}
\item All accumulation points of the sequence $\bm{\eta}_k$ belong to $Z$ almost surely;
\item There exists a constant $C$ such that $\EE[\|\bxi_k\|^2\given \cF_k] \le C$ for all $k\ge 0$;
\item $\sum_{k=0}^\infty \EE[\rho_k^2 + \rho_k\nbr{\bb_k}] < \infty$ and $\sum_{k=0}^\infty \rho_k = \infty$ almost surely;
\item $\nbr{\bm{\eta}_{k+1}-\bm{\eta}_k}/\rho_k\to 0$ almost surely.
\end{enumerate}
Then $\bz_k - \bm{\eta}_k \to \zero$ almost surely.     
\end{lemma}

\begin{lemma}[Adapted from Lemma 1 in \cite{Polyak1992Acceleration}]\label{pre_lemma:sum_of_product_1}
Let $M\in\mR^{d\times d}$ be symmetric positive definite, and let $\{a_k\}_{k\ge 0}$ be a deterministic sequence with $a_k = \iota (k+1)^{-p}$ for $\iota > 0$ and~$p\in (0,1)$. Define
\begin{equation*}
B_i^t = a_iM\sum_{k=i}^t \prod_{j=i+1}^k (I-a_jM) \qquad \text{and} \qquad A_i^t = B_i^t - I.
\end{equation*}
Then there exists a constant $\Upsilon > 0$ such that $\max\{\nbr{B_i^t}, \nbr{A_i^t}\} \le \Upsilon$ for all $i,t \ge 0$. Furthermore, $\lim_{t\to \infty}\frac{1}{t}\sum_{i=0}^{t}\nbr{A_i^t} = 0$. 
\end{lemma}

\begin{lemma}[Adapted from Lemma B.3 in \cite{Na2025Statistical}]\label{pre_lemma:sum_of_product_2}

Let $\alpha_k = \iota_1(k+1)^{-p_1}$ and $\beta_k = \iota_2(k+1)^{-p_2}$ be two sequences with $\iota_1,\iota_2,p_1,p_2 > 0$. The following results hold. 
\begin{enumerate}[label=(\alph*),topsep=0pt]
\setlength\itemsep{0.0em}
\item Suppose $0 < p_2 \le 1$. Let $l\ge 1$ be an integer and $a_1,\ldots,a_l$ be constants. Let $\chi = 0$ if $0 < p_2 < 1$ and $\chi = -p_1/\iota_2$ if $p_2 = 1$. Then, as long as $\sum_{t=1}^l a_t + \chi > 0$, we have
\begin{align*}
&\lim_{k\to\infty}\frac{1}{\alpha_k}\sum_{i=0}^{k}\prod_{j=i+1}^k\prod_{t=1}^l (1-a_t\beta_j)\beta_i\alpha_i = \frac{1}{\sum_{t=1}^l a_t + \chi}, \\
&\lim_{k\to\infty}\frac{1}{\alpha_k}\cbr{\sum_{i=0}^k\prod_{j=i+1}^k\prod_{t=1}^l(1-a_t\beta_j)\beta_i\alpha_ie_i + b\prod_{j=0}^k\prod_{t=1}^l(1-a_t\beta_j)} = 0,
\end{align*}
where the second result holds for any constant $b$ and sequence $\{e_i\}$ such that $e_i\to 0$.    
\item
If $0 < p_2 < p_1 \le 1$, then 
\begin{equation*}
\lim_{k\to \infty} \frac{1}{\alpha_k}\sum_{i=0}^k\prod_{j=i+1}^k(1-\alpha_j)(1-\beta_j)\alpha_i\beta_i = 1.
\end{equation*}    
\end{enumerate}
\end{lemma}

In Lemmas~\ref{pre_lemma:sketching_1} and~\ref{pre_lemma:sketching_2}, we use the notation introduced in the main paper. The additional notation defined in these two lemmas will also be used later in the proofs and analysis.

\begin{lemma}[Adapted from Lemma~4.4 in \cite{Na2025Statistical}]\label{pre_lemma:sketching_1}
Under Assumption \ref{assumption:sketching}, for any $k\geq 0$, let us define two sketching matrices 
\begin{equation*}
\tilde{C}_{k,j} := I-(\tilde{W}_kS_{k,j}(S_{k,j}^\top\tilde{W}_{k}^2S_{k,j})^\dagger S_{k,j}^\top \tilde{W}_k) \quad\quad \text{and} \quad\quad \tilde{C}_k = \prod_{j=0}^{s-1} \tilde{C}_{k,j}.
\end{equation*}
We also let $C_k = \EE[\tilde{C}_k\given \cF_{k-2/3}]$. Then, the following results hold (only (b)-(d) rely on Assumption~\ref{assumption:sketching}).
\begin{enumerate}[label=(\alph*),topsep=0pt]
\setlength\itemsep{0.0em}	
\item We have $\bz_{k,s} = (I-\tilde{C}_k)\tilde{\bz}_{k}$ and $\|\tilde{C}_k\|\leq 1$.
\item Let $\rho := 1-\gamma_S.$ We have $0\le \rho < 1$.
\item $\EE[\bz_{k,s} - \tilde{\bz}_{k}\given \cF_{k-2/3}] = - C_k\tilde{\bz}_{k}$, and $\| C_k\|\le \rho^s$.
\item $\EE[\|\bz_{k,s}-\tilde{\bz}_k\|^2\given \cF_{k-2/3}] \le \rho^s\|\tilde{\bz}_{k}\|^2$.
\end{enumerate}
\end{lemma}

\begin{lemma}[Adapted from Corollary~5.4 in \cite{Na2025Statistical}]\label{pre_lemma:sketching_2}

Recall that $\sigma_{\rm min}(\cdot)$ denotes the smallest singular value and $W^\star := \nabla^2\mL(\bx^\star, \blambda^\star)$. Let 
\begin{equation*}
\tilde{C}^\star:= \prod_{j=1}^s(I-W^\star S_j(S_j^\top(W^\star)^2S_j)^\dagger S_j^\top W^\star),\quad C^\star:= \EE[\tilde{C}^\star] = (I - \EE[W^\star S(S^\top (W^\star)^2S)^\dagger S^\top W^\star])^s,
\end{equation*}
where $S_1,\ldots, S_s \stackrel{\mathrm{i.i.d.}}{\sim} S$.
Then, under Assumption \ref{assumption:sketching}, 
\begin{equation*}
\nbr{C_k - C^\star}\le 2s\Upsilon_S \|\tilde{W}_k - W^\star\|/\sigma_{\rm min}(W^\star).
\end{equation*}
\end{lemma}

\begin{lemma}[\cite{Orabona2020Almost}]\label{pre_lemma:convergence_to_zero}
Let $\{b_t, \eta_t\}_{t\ge 0}$ be two non-negative sequences and $\{\ba_t\}_{t\ge 0}$ a sequence of vectors. Let $p\ge 1$ and assume $\sum_{t=0}^\infty \eta_tb_t^p < \infty$ and $\sum_{t=0}^{\infty}\eta_t = \infty$. Assume also that there exists $L\ge 0$ such that $\abr{b_{t+\tau} - b_t} \le L(\sum_{i=t}^{t+\tau -1}\eta_ib_i + \|\sum_{i=t}^{t+\tau-1}\eta_i\ba_i\|)$, where $\ba_t$ is such that $\|\sum_{t=0}^\infty \eta_t\ba_t\| < \infty$. Then, $b_t$ converges to $0$. 
\end{lemma}

\section{Proofs of Section \ref{sec:global_convergence}}

In this section, we provide proofs for Section \ref{sec:global_convergence}. For proofs in this and later sections, we use $\cC_1, \cC_2, \ldots$ to denote generic deterministic constants, which can be different for each appearance.

\subsection{Proof of Lemma \ref{lemma:gradient_approx_convergence}}\label{proof_of_lemma:gradient_approx_convergence}

The almost-sure convergence proof is an application of Lemma \ref{pre_lemma:as_convergence}. By \eqref{eq:gradient_approx}, we have $\bar{\bg}_k = (1-\beta_k)\bar{\bg}_{k-1} + \beta_k\hat{\nabla}F(\bx_k;\xi_k)$. Since $\nabla f_k \in \cF_{k-1}$, we also have  $\EE[\hat{\nabla}F(\bx_k;\xi_k)\given \cF_{k-1}] = \nabla f_k + \bb_k$, where 
\begin{equation}\label{eq:3_proof_of_lemma:gradient_approx_convergence}
\bb_k := \EE\sbr{\hat{\nabla}F(\bx_k;\xi_k) - \nabla f_k\given \cF_{k-1}} = O(b_k^2)
\end{equation}
by Lemma \ref{lemma:approximation_rate}. We now verify the conditions in Lemma \ref{pre_lemma:as_convergence}. Condition $(a)$ is trivially satisfied because we do not use a projection step, or equivalently because the projection is onto $\RR^d$. For condition $(b)$, we note that
\begin{equation*}
\hat{\nabla} F(\bx_k; \xi_k) = \frac{1}{2b_k}\rbr{\int_{-b_k}^{b_k}\inner{\nabla F(\bx_k+ s\bDelta_k;\xi_k)}{\bDelta_k}ds}\bDelta^{-1}_k.
\end{equation*}
Then, by Jensen's inequality,
\begin{align}\label{eq:1_proof_of_lemma:gradient_approx_convergence}
\|\hat{\nabla}F(\bx_k;\xi_k)\|^2 & = \abr{\frac{1}{2b_k}\int_{-b_k}^{b_k}\inner{\nabla F(\bx_k+ s\bDelta_k;\xi_k)}{\bDelta_k}ds}^2\|\bDelta_k^{-1}\|^2\notag\\
& \le \frac{1}{2b_k}\int_{-b_k}^{b_k}\abr{\inner{\nabla F(\bx_k + s\bDelta_k; \xi_k)}{\bDelta_k}}^2ds\nbr{\bDelta_k^{-1}}^2\notag\\
&\le d^2 \kappa_{\bDelta_1}^{-2}\kappa_{\bDelta_2}^2 \frac{1}{2b_k}\int_{-b_k}^{b_k}\nbr{\nabla F(\bx_k + s\bDelta_k;\xi_k)}^2 ds, 
\end{align}
where the first inequality is from Jensen's inequality and the second is by Assumption \ref{assumption:boundedness_3}. Then, we further have 
\begin{align} \label{eq:2_proof_of_lemma:gradient_approx_convergence}
& \EE[\|\hat{\nabla}F(\bx_k;\xi_k)\|^2\given \cF_{k-1}] = \EE\sbr{\EE[\|\hat{\nabla}F(\bx_k;\xi_k)\|^2\given \cF_{k-1}, \bDelta_k]\given \cF_{k-1}}\notag\\
&\stackrel{\mathclap{\eqref{eq:1_proof_of_lemma:gradient_approx_convergence}}}{\le}\; \EE\sbr{\EE\sbr{d^2 \kappa_{\bDelta_1}^{-2}\kappa_{\bDelta_2}^2 \frac{1}{2b_k}\int_{-b_k}^{b_k}\nbr{\nabla F(\bx_k + s\bDelta_k;\xi_k)}^2 ds\given \cF_{k-1}, \bDelta_k}\given \cF_{k-1}}\notag\\
& = d^2 \kappa_{\bDelta_1}^{-2}\kappa_{\bDelta_2}^2 \frac{1}{2b_k}\EE\sbr{\int_{-b_k}^{b_k}\EE[\nbr{\nabla F(\bx_k+s\bDelta_k;\xi_k)}^2\given \cF_{k-1}, \bDelta_k]ds\given \cF_{k-1}}\notag\\
&= d^2 \kappa_{\bDelta_1}^{-2}\kappa_{\bDelta_2}^2 \frac{1}{2b_k}\EE\sbr{\int_{-b_k}^{b_k}\EE[\nbr{\nabla F(\bx_k+s\bDelta_k;\xi_k)}^2\given \bx_k+s\bDelta_k]ds\given \cF_{k-1}},
\end{align}
where the penultimate equality follows from Tonelli's theorem, and the last equality uses the fact that $\xi_k$ is independent of $\sigma(\cF_{k-1}, \bDelta_k)$. Next, we reduce the conditional second moment of the estimator to a uniform moment bound on the stochastic gradient evaluated along the perturbation path. In particular, by Assumption \ref{assumption:boundedness_2} and Jensen's inequality,
\begin{equation*}
\EE[\nbr{\nabla F(\bx;\xi) - \nabla f(\bx)}^2\given \bx] \le (\EE[\nbr{\nabla F(\bx;\xi)-\nabla f(\bx)}^r\given \bx])^{2/r} \le \Upsilon_m^{2/r}.
\end{equation*}
By Assumption \ref{assumption:boundedness_1}, we further have
\begin{equation*}
\EE[\|\nabla F(\bx;\xi)\|^2\given \bx] \le 2\EE[\nbr{\nabla F(\bx;\xi) - \nabla f(\bx)}^2\given \bx] + 2\nbr{\nabla f(\bx)}^2 \stackrel{\eqref{equ:Ups_u}}{\leq} 2\Upsilon_m^{2/r} + 2\Upsilon_u^2.
\end{equation*}
Then \eqref{eq:2_proof_of_lemma:gradient_approx_convergence} gives
\begin{equation}
\EE[\|\hat{\nabla}F(\bx_k;\xi_k)\|^2\given \cF_{k-1}] \le d^2\kappa_{\bDelta_1}^{-2}\kappa_{\bDelta_2}^2(2\Upsilon_m^{2/r} + 2\Upsilon_u^2), \label{eq:hat_nabla_F_second_moment_bound}
\end{equation}
so the condition $(b)$ in Lemma \ref{pre_lemma:as_convergence} is satisfied. By \eqref{eq:3_proof_of_lemma:gradient_approx_convergence} and our assumption that $p_2\in (0.5, 1]$ and $p_3 > 0.5 - 0.5p_2$, the condition $(c)$ is automatically satisfied. Finally, we check the condition $(d)$.~That~is, $\|\nabla f_{k+1} - \nabla f_k\|/\beta_{k} \to 0$ almost surely. By \eqref{eq:stepsize_selection} and Assumption~\ref{assumption:stepsize_selection}, under the condition $p\geq 1$, we have
\begin{equation}\label{eq:bar_alpha_bound}
\bar{\alpha}_k \le \mu_k\alpha_k + \psi \alpha_k^p\le (\kappa_{2,\mu} + \psi\alpha_k^{p-1})\alpha_k \le (\kappa_{2,\mu} + \psi\iota_1^{p-1})\alpha_k \eqqcolon \kappa_{3,\mu}\alpha_k.
\end{equation}
By Lipschitz continuity property, we have
\begin{align}
\|\nabla f_{k+1} - \nabla f_k\| &\stackrel{\eqref{equ:Ups_u}}{\leq} \Upsilon_u\bar{\alpha}_k\|\bar{\Delta}\bx_k\|\le 2\Upsilon_u\bar{\alpha}_k\|\tilde{\bz}_k\| \stackrel{\eqref{equ:KKT}}{\le} 2\kappa_{3,\mu}\Upsilon_u \alpha_k \|\tilde{W}_k^{-1}\| (\|\bar{\nabla}_{\bx}\cL_k\| + \|c_k\|)\\
&\stackrel{\eqref{eq:debiased_Lagrangian_Hessian}}{\le} 2\kappa_{3,\mu}\Upsilon_u \alpha_k \|\tilde{W}_k^{-1}\|(\|\bar{\bg}_k\|+ \|\blambda_k\|\|\tilde{G}_k\| + \|c_k\|)\\
&\le 2\kappa_{3,\mu}\Upsilon_u\alpha_k \|\tilde{W}_k^{-1}\|(\|\bar{\bg}_k\| + \Upsilon_{\blambda}\sqrt{\kappa_{2,\tilde{G}}} + \kappa_c), \label{eq:nabla_f_step_difference_bound}
\end{align}
where the second inequality is by Lemma \ref{pre_lemma:sketching_1} (a) and the last inequality is by Assumption \ref{assumption:boundedness_1}. By Assumption \ref{assumption:boundedness_1} and \citep[Lemma~1]{Na2022adaptive}, we have $\|\tilde{W}_k^{-1}\| \le \Upsilon_W$ for some constant $\Upsilon_W$. Therefore, we have
\begin{align}\label{eq:4_proof_of_lemma:gradient_approx_convergence}
\EE[\nbr{\nabla f_{k+1} - \nabla f_{k}}^{r}] &\le \EE[\|2\kappa_{3,\mu}\Upsilon_u \Upsilon_W\alpha_k(\|\bar{\bg}_k\| + \Upsilon_{\blambda}\sqrt{\kappa_{2,\tilde{G}}} + \kappa_c)\|^r] \nonumber\\
&\le 2^{r-1}(2\kappa_{3,\mu}\Upsilon_u \Upsilon_W\alpha_k)^r(\EE[\|\bar{\bg}_k\|^r] + (\Upsilon_{\blambda}\sqrt{\kappa_{2,\tilde{G}}} + \kappa_c)^r). 
\end{align}
The last inequality uses the convexity inequality $(a+b)^r \le 2^{r-1}(a^r + b^r)$ for any $a,b\ge 0$ and $r\ge 1$. We now derive a uniform bound for $\EE\sbr{\nbr{\bar{\bg}_k}^r}$ for \eqref{eq:4_proof_of_lemma:gradient_approx_convergence}. By the same analysis as in \eqref{eq:1_proof_of_lemma:gradient_approx_convergence} and \eqref{eq:2_proof_of_lemma:gradient_approx_convergence},
\begin{align*}
\EE[\|\hat{\nabla}F(\bx_k;\xi_k)\|^r\given \cF_{k-1}] & \le \EE\sbr{d^r\kappa_{\bDelta_1}^{-r}\kappa_{\bDelta_2}^r\frac{1}{2b_k}\int_{-b_k}^{b_k}\|\nabla F(\bx_k + s\bDelta_k;\xi_k)\|^rds \given \cF_{k-1}}\\
&= d^r\kappa_{\bDelta_1}^{-r}\kappa_{\bDelta_2}^r\frac{1}{2b_k}\EE\sbr{\int_{-b_k}^{b_k}\EE[\|\nabla F(\bx_k + s\bDelta_k;\xi_k)\|^r\given \bx_k + s\bDelta_k]ds \given \cF_{k-1}}.
\end{align*}
By the convexity inequality again, we have
\begin{align*}
\EE[\|\nabla F(\bx;\xi)\|^r\given \bx] \le 2^{r-1}(\EE[\|\nabla F(\bx;\xi) - \nabla f(\bx)\|^r\given \bx] + \|\nabla f(\bx)\|^r) \le 2^{r-1}(\Upsilon_m + \Upsilon_u^r).
\end{align*}
Therefore, $\EE[\|\hat{\nabla}F(\bx_k;\xi_k)\|^r\given \cF_{k-1}] \le 2^{r-1}d^r\kappa_{\bDelta_1}^{-r}\kappa_{\bDelta_2}^r(\Upsilon_m + \Upsilon_u^r)$.
Let $\Upsilon_{\bar{\bg}} := \max\{2^{r-1}d^r\kappa_{\bDelta_1}^{-r}\kappa_{\bDelta_2}^r(\Upsilon_m + \Upsilon_u^r), \nbr{\bar{\bg}_{-1}}^r\}$.
We show that $\EE[\|\bar{\bg}_k\|^r] \le \Upsilon_{\bar{\bg}}$ by induction. For $k = -1$, the statement holds automatically by our definition of $\Upsilon_{\bar{\bg}}$. We assume that $\EE[\|\bar{\bg}_k\|^r] \le \Upsilon_{\bar{\bg}}$ for the purpose of induction. Then 
\begin{align}\label{eq:bar_g_r_moment_bound}
\EE[\|\bar{\bg}_{k+1}\|^r] &\stackrel{\eqref{eq:gradient_approx}}{=} \EE[\|(1-\beta_{k+1})\bar{\bg}_k + \beta_{k+1}\hat{\nabla} F(\bx_{k+1};\xi_{k+1})\|^r] \nonumber\\
&\le (1-\beta_{k+1})\EE[\|\bar{\bg}_k\|^r] + \beta_{k+1}\EE[\|\hat{\nabla} F(\bx_{k+1}; \xi_{k+1})\|^r] \nonumber\\
&\le (1-\beta_{k+1})\Upsilon_{\bar{\bg}} + \beta_{k+1}\Upsilon_{\bar{\bg}} = \Upsilon_{\bar{\bg}}, 
\end{align}
where the first inequality is given by Jensen's inequality. Therefore, we have proved our statement through induction, and it follows from \eqref{eq:4_proof_of_lemma:gradient_approx_convergence} that 
\begin{equation}
\EE\sbr{\nbr{\nabla f_{k+1} - \nabla f_{k}}^{r}} \le 2^{r-1}(2\kappa_{3,\mu}\Upsilon_u \Upsilon_W\alpha_k)^r(\Upsilon_{\bar{\bg}} + (\Upsilon_{\blambda}\sqrt{\kappa_{2,\tilde{G}}} + \kappa_c)^r). \label{eq:nabla_f_step_difference_r_moment_bound}
\end{equation}
Then, we have
\begin{align*}
\EE\sbr{\sum_{k=1}^\infty\frac{\|\nabla f_{k+1} - \nabla f_{k}\|^r}{\beta_k^r}} &= \sum_{k=1}^{\infty}\frac{\EE[\|\nabla f_{k+1} - \nabla f_{k}\|^r]}{\beta_k^r}\\
&\stackrel{\eqref{eq:nabla_f_step_difference_r_moment_bound}}{\le} \sum_{k=1}^{\infty} 2^{r-1}(2\kappa_{3,\mu}\Upsilon_u \Upsilon_W)^r(\Upsilon_{\bar{\bg}} + (\Upsilon_{\blambda}\sqrt{\kappa_{2,\tilde{G}}} + \kappa_c)^r)(\alpha_k/\beta_k)^r < \infty,
\end{align*}
where the last inequality is from our assumption that $r(p_1-p_2) > 1$. Then, we get $\|\nabla f_{k+1}-\nabla f_{k}\|^r/\beta_k^r\to 0$ almost surely, so equivalently $\|\nabla f_{k+1} - \nabla f_{k}\|/\beta_k \to 0$ almost surely. Thus, by Lemma \ref{pre_lemma:as_convergence}, $\bar{\bg}_k - \nabla f_k \to \zero$ almost surely. The same argument applies to $\bar{G}_k$ and gives $\bar{G}_k - G_k \to\zero$ almost surely. From this convergence, there exists a random $K_{G}^\star$ such that, for $k \ge K_{G}^\star$, $\|\bar{G}_k \bar{G}_k^\top - G_kG_k^\top\| \le \min\{\kappa_{1,G} - \kappa_{1,\tilde{G}}, \kappa_{2,\tilde{G}} - \kappa_{2,G}\}$. Then, by Weyl's inequality, $\kappa_{1,\tilde{G}}\cdot I\preceq \bar{G}_k\bar{G}_k^\top \preceq\kappa_{2,\tilde{G}}\cdot I$. Since $\delta^G_k$ is introduced only to enforce this condition, we have $\delta^G_k = \zero$ for $k \ge K_G^\star$.

\subsection{Proof of Lemma \ref{lemma:gradient_approx_rate}}\label{proof_of_lemma:gradient_approx_rate}

We let $\nabla f_{-1} := \bar{\bg}_{-1}$ and we begin by unrolling the momentum recursion for the gradient estimator. We have
\begin{align*}
&\bar{\bg}_k - \nabla f_k \stackrel{\eqref{eq:gradient_approx}}{=} (1-\beta_k)\bar{\bg}_{k-1} + \beta_k\hat{\nabla}F(\bx_k;\xi_k) - \nabla f_k\\
&= (1-\beta_k)(\bar{\bg}_{k-1} - \nabla f_{k-1}) + (1-\beta_k)(\nabla f_{k-1} - \nabla f_{k}) + \beta_k(\EE[\hat{\nabla} F(\bx_k; \xi_k)\given \cF_{k-1}] - \nabla f_k)\\
&\quad + \beta_k(\hat{\nabla} F(\bx_{k}; \xi_k) - \EE[\hat{\nabla} F(\bx_{k}; \xi_k)\given \cF_{k-1}])\\
&= \sum_{i=0}^k\prod_{j=i}^k(1-\beta_j)(\nabla f_{i-1} -\nabla f_i) + \sum_{i=0}^k\prod_{j=i+1}^k (1-\beta_j)\beta_i(\EE[\hat{\nabla} F(\bx_i;\xi_i)\given \cF_{i-1}] - \nabla f_i)\\
&\quad + \sum_{i=0}^k\prod_{j=i+1}^k(1-\beta_j)\beta_i(\hat{\nabla}F(\bx_i;\xi_i) - \EE[\hat{\nabla} F(\bx_i;\xi_i)\given \cF_{i-1}]). 
\end{align*}
We let 
\begin{align*}
&\cE_k^1 := \EE\sbr{\nbr{\sum_{i=0}^k\prod_{j=i+1}^k(1-\beta_j)\beta_i(\hat{\nabla}F(\bx_i;\xi_i) - \EE[\hat{\nabla} F(\bx_i;\xi_i)\given \cF_{i-1}])}^2},\\
&\cE_k^2 := \EE\sbr{\nbr{\sum_{i=0}^k\prod_{j=i+1}^k (1-\beta_j)\beta_i(\EE[\hat{\nabla} F(\bx_i;\xi_i)\given \cF_{i-1}] - \nabla f_i)}^2},\\
&\cE_k^3 := \EE\sbr{\nbr{\sum_{i=0}^k\prod_{j=i}^k(1-\beta_j)(\nabla f_{i-1} -\nabla f_i)}^2}.
\end{align*}
The three quantities above will be treated separately: $\cE_k^1$ is controlled by martingale orthogonality, $\cE_k^2$ by the bias of the zeroth-order estimator, and $\cE_k^3$ by the movement of the iterates. Since for $i< j$,
\begin{align*}
&\EE[(\hat{\nabla}F(\bx_i;\xi_i) - \EE[\hat{\nabla}F(\bx_i;\xi_i)\given \cF_{i-1}])\cdot(\hat{\nabla}F(\bx_j;\xi_j) - \EE[\hat{\nabla}F(\bx_j;\xi_j)\given \cF_{j-1}])]\\
&= \EE\sbr{(\hat{\nabla}F(\bx_i;\xi_i) - \EE[\hat{\nabla} F(\bx_i;\xi_i)\given \cF_{i-1}])\cdot\EE\sbr{\hat{\nabla} F(\bx_j;\xi_j) - \EE[\hat{\nabla} F(\bx_j;\xi_j)\given \cF_{j-1}] \given \cF_{i}}}\\
&= \EE[(\hat{\nabla}F(\bx_i;\xi_i) - \EE[\hat{\nabla} F(\bx_i;\xi_i)\given \cF_{i-1}])\cdot(\EE[\hat{\nabla}F(\bx_j;\xi_j)\given \cF_i] - \EE[\hat{\nabla}F(\bx_j;\xi_j)\given \cF_i])] = 0,
\end{align*}
we have 
\begin{equation*}
\cE_k^1 = \sum_{i=0}^{k}\prod_{j=i+1}^k(1-\beta_j)^2\beta_i^2\EE[\|\hat{\nabla}F(\bx_i;\xi_i) - \EE[\hat{\nabla}F(\bx_i;\xi_i) \given \cF_{i-1}]\|^2].
\end{equation*}
Under Assumption~\ref{assumption:boundedness_1}, by \eqref{eq:hat_nabla_F_second_moment_bound}, we have $\EE[\|\hat{\nabla}F(\bx_i;\xi_i) - \EE[\hat{\nabla}F(\bx_i;\xi_i) \given \cF_{i-1}]\|^2]$ is uniformly bounded. Thus, we apply Lemma \ref{pre_lemma:sum_of_product_2} and have $\cE_k^1 = O(\beta_k)$. For the term $\cE_k^2$, we note that
\begin{align*}
&\nbr{\sum_{i=0}^k\prod_{j=i+1}^k(1-\beta_j)\beta_i(\EE[\hat{\nabla}F(\bx_i;\xi_i) \given \cF_{i-1}]- \nabla f_i)}\\&\le \sum_{i=0}^k\prod_{j=i+1}^k(1-\beta_j)\beta_i\|\EE[\hat{\nabla}F(\bx_i;\xi_i)\given \cF_{i-1}] - \nabla f_i\|\le \sum_{i=0}^k\prod_{j=i+1}^k(1-\beta_j)\beta_i\|\EE[\hat{\nabla}F(\bx_i;\xi_i) - \nabla f_i\given \cF_{i-1}]\|.
\end{align*}
Then, by Lemma \ref{lemma:approximation_rate} and Lemma \ref{pre_lemma:sum_of_product_2}, $\|\sum_{i=0}^k\prod_{j=i+1}^k(1-\beta_j)\beta_i(\EE[\hat{\nabla}F(\bx_i;\xi_i)\given \cF_{i-1}] - \nabla f_i)\| = O(b_k^2).$
Thus, we have $\cE_k^2 = O(b_k^4)$. Finally, for the drift term $\cE_k^3$, we apply Minkowski's inequality and~have
\begin{equation*}
\sqrt{\cE_k^3}\le \sum_{i=0}^k\prod_{j=i}^k(1-\beta_j)\sqrt{\EE[\|\nabla f_{i-1} - \nabla f_i\|^2]} \le \Upsilon_u\sum_{i=0}^k\prod_{j=i}^k(1-\beta_j) \sqrt{\EE[\|\bar{\alpha}_{i-1}{\bz}_{i-1,s}\|^2]},
\end{equation*}
where we can freely define $\bar{\alpha}_{-1}$ and ${\bz}_{-1,s}$ to make the inequality hold for $i = 0$. 
By Assumption \ref{assumption:stepsize_selection} and Lemma~\ref{pre_lemma:sketching_1} under Assumption~\ref{assumption:sketching}, we have $\EE[\|\bar{\alpha}_{i-1}{\bz}_{i-1,s}\|^2] \le \kappa_{3,\mu}^2\alpha_{i-1}^2\EE[\|{\bz}_{i-1,s}\|^2] \le 4\kappa_{3,\mu}^2\alpha_{i-1}^2\EE[\|\tilde{\bz}_{i-1}\|^2]$,
where again we can define $\alpha_{-1}$ and $\tilde{\bz}_{-1}$ to make the above inequality hold for $i = 0$. 
Combining the analysis in \eqref{eq:nabla_f_step_difference_bound} and \eqref{eq:bar_g_r_moment_bound}, we have that $\EE[\|\tilde{\bz}_{i-1}\|^2]$ is uniformly bounded. Therefore, $\sqrt{\cE_k^3} \le \cC_1\sum_{i=0}^{k}\prod_{j=i}^k(1-\beta_j)\beta_i \frac{\alpha_i}{\beta_i} = O({\alpha_k}/{\beta_k})$.

Combining the martingale $\cE_k^1$, bias $\cE_k^2$, and drift $\cE_k^3$ bounds, and using $\EE[\|\bar{\bg}_k - \nabla f_k\|^2] \le 3(\cE_k^1 + \cE_k^2 + \cE_k^3)$, yields the desired rate for the gradient estimator: $\EE[\|\bar{\bg}_k - \nabla f_k\|^2] = O(\beta_k + b_k^4 + \alpha_k^2/\beta_k^2)$.
A similar analysis gives $\EE[\|\bar{G}_k - G_k\|^2] = O(\beta_k + b_k^4 + \alpha_k^2/\beta_k^2)$. This completes the proof.

\subsection{Proof of Lemma \ref{lemma:global_convergence}}\label{proof_of_lemma:global_convergence}

By Assumption \ref{assumption:boundedness_1}, we know for some constants $\cC_1, \cC_2>0$ that
\begin{equation*}
\|\nabla_{\bx}^2\cL_k\|\le \cC_1
\quad\quad\quad 
\text{and}
\quad \quad\quad
\nbr{\begin{pmatrix}
I + \nu\nabla_{\bx}^2\cL_k &\sigma \tilde{G}_k^\top\\
\nu \tilde{G}_k & I
\end{pmatrix}} \le \cC_2.
\end{equation*} 
Note that $\cC_2$ depends on $(\nu, \sigma)$ that will be specified later. We have
\begin{align}\label{ns:equ:2}
(\nabla \cL_{\sigma, \nu}^k)^\top \tilde{\bz}_k & = \binom{\nabla_{\bx}\cL_{\sigma, \nu}^k}{\nabla_{\blambda}\cL_{\sigma,\nu}^k}^\top\binom{\tilde{\Delta}\bx_k}{\tilde{\Delta}\blambda_k} = \binom{\tilde{\Delta}\bx_k}{\tilde{\Delta}\blambda_k}^\top
\begin{pmatrix}
I + \nu\nabla_{\bx}^2\cL_k &\sigma G_k^\top\\
\nu G_k & I
\end{pmatrix}
\binom{\nabla_{\bx}\cL_k}{c_k} \nonumber\\
&\le \binom{\tilde{\Delta}\bx_k}{\tilde{\Delta}\blambda_k}^\top
\begin{pmatrix}
I + \nu\nabla_{\bx}^2\cL_k &\sigma \tilde{G}_k^\top\\
\nu \tilde{G}_k & I
\end{pmatrix}\binom{\nabla_{\bx}\cL_k}{c_k} + \cC_3\|\tilde{G}_k - G_k\|\|\tilde{\bz}_k\| \nonumber\\
&\le \binom{\tilde{\Delta}\bx_k}{\tilde{\Delta}\blambda_k}^\top
\begin{pmatrix}
I + \nu\nabla_{\bx}^2\cL_k &\sigma \tilde{G}_k^\top\\
\nu \tilde{G}_k & I
\end{pmatrix}
\binom{\bar{\nabla}_{\bx}\cL_k}{c_k} + \cC_2\|\nabla_{\bx}\cL_k - \bar{\nabla}_{\bx}\cL_k\|\|\tilde{\bz}_k\| + \cC_3\|\tilde{G}_k - G_k\|\|\tilde{\bz}_k\|\nonumber\\
&\stackrel{\eqref{equ:KKT}}{=}  -\binom{\tilde{\Delta}\bx_k}{\tilde{\Delta}\blambda_k}^\top
\begin{pmatrix}
I + \nu\nabla_{\bx}^2\cL_k &\sigma \tilde{G}_k^\top\\
\nu \tilde{G}_k & I
\end{pmatrix}
\begin{pmatrix}
\tilde{B}_k &\tilde{G}_k^\top\\
\tilde{G}_k &\zero
\end{pmatrix}
\binom{\tilde{\Delta}\bx_k}{\tilde{\Delta}\blambda_k}+ \cC_2\|\nabla_{\bx}\cL_k - \bar{\nabla}_{\bx}\cL_k\|\|\tilde{\bz}_k\| \nonumber\\
&\quad + \cC_3\|\tilde{G}_k - G_k\|\|\tilde{\bz}_k\| \nonumber\\
&= -\binom{\tilde{\Delta}\bx_k}{\tilde{\Delta}\blambda_k}^\top
\begin{pmatrix}
\tilde{B}_k + \nu\nabla^2_{\bx}\cL_k\tilde{B}_k + \sigma \tilde{G}_k^\top\tilde{G}_k &\tilde{G}_k^\top + \nu\nabla^2_{\bx}\cL_k\tilde{G}_k^\top\\
\tilde{G}_k + \nu\tilde{G}_k\tilde{B}_k &\nu\tilde{G}_k\tilde{G}_k^\top
\end{pmatrix}
\binom{\tilde{\Delta}\bx_k}{\tilde{\Delta}\blambda_k} \nonumber\\
&\quad + \cC_2\|\nabla_{\bx}\cL_k - \bar{\nabla}_{\bx}\cL_k\|\|\tilde{\bz}_k\| + \cC_3\|\tilde{G}_k - G_k\|\|\tilde{\bz}_k\|,
\end{align}
where $\cC_3$ is some constant such that $\max\{\nu,\sigma\}\nbr{(\nabla_{\bx}\cL_k, c_k)} \le \cC_3$, which also depends on $(\nu,\sigma)$ selected later. We next bound the leading quadratic form. We note that
\begin{align*}
-\binom{\tilde{\Delta}\bx_k}{\tilde{\Delta}\blambda_k}^\top &
\begin{pmatrix}
\tilde{B}_k + \nu\nabla^2_{\bx}\cL_k\tilde{B}_k + \sigma \tilde{G}_k^\top\tilde{G}_k &\tilde{G}_k^\top + \nu\nabla^2_{\bx}\cL_k\tilde{G}_k^\top\\
\tilde{G}_k + \nu\tilde{G}_k\tilde{B}_k &\nu\tilde{G}_k\tilde{G}_k^\top
\end{pmatrix}
\binom{\tilde{\Delta}\bx_k}{\tilde{\Delta}\blambda_k}\\
&\le -\tilde{\Delta}\bx_k^\top \tilde{B}_k\tilde{\Delta}\bx_k + \nu{\kappa_{2,\tilde{B}}}\cC_1\|\tilde{\Delta}\bx_k\|^2 - \sigma\|\tilde{G}_k\tilde{\Delta}\bx_k\|^2 - 2\tilde{\Delta}\blambda_k^\top \tilde{G}_k\tilde{\Delta}\bx_k\\
&\quad + \nu(\cC_1 + {\kappa_{2,\tilde{B}}})\|\tilde{\Delta}\bx_k\|\|\tilde{G}_k^\top\tilde{\Delta}\blambda_k\| - \nu\|\tilde{G}_k^\top\tilde{\Delta}\blambda_k\|^2\\
&\stackrel{\eqref{equ:KKT}}{=} -\tilde{\Delta}\bx_k^\top\tilde{B}_k\tilde{\Delta}\bx_k + \nu{\kappa_{2,\tilde{B}}}\cC_1\|\tilde{\Delta}\bx_k\|^2 - \sigma\nbr{c_k}^2 + 2c_k^\top\tilde{\Delta}\blambda_k \\
&\quad + \nu(\cC_1 + {\kappa_{2,\tilde{B}}})\|\tilde{\Delta}\bx_k\|\|\tilde{G}_k^\top\tilde{\Delta}\blambda_k\| - \nu\|\tilde{G}_k^\top\tilde{\Delta}\blambda_k\|^2\\
&\le -\tilde{\Delta}\bx_k^\top\tilde{B}_k\tilde{\Delta}\bx_k + \nu{\kappa_{2,\tilde{B}}}\cC_1\|\tilde{\Delta}\bx_k\|^2 - \sigma\|c_k\|^2 + 2c_k^\top\tilde{\Delta}\blambda_k + \frac{\nu(\cC_1 + {\kappa_{2,\tilde{B}}})^2}{2}\|\tilde{\Delta}\bx_k\|^2- \frac{\nu}{2}\|\tilde{G}_k^\top\tilde{\Delta}\blambda_k\|^2\\
&\le -\tilde{\Delta}\bx_k^\top\tilde{B}_k\tilde{\Delta}\bx_k + \nu(\cC_1 + {\kappa_{2,\tilde{B}}})^2\|\tilde{\Delta}\bx_k\|^2 - \sigma \|c_k\|^2 + \frac{8}{\nu\kappa_{1,\tilde{G}}}\|c_k\|^2 + \frac{\nu\kappa_{1,\tilde{G}}}{8}\|\tilde{\Delta}\blambda_k\|^2\\
&\quad - \frac{\nu\kappa_{1,\tilde{G}}}{4}\|\tilde{\Delta}\blambda_k\|^2 - \frac{\nu}{4}\|\tilde{G}_k^\top\tilde{\Delta}\blambda_k\|^2\\
&\stackrel{\eqref{equ:KKT}}{=} -\tilde{\Delta}\bx_k^\top\tilde{B}_k\tilde{\Delta}\bx_k + \nu(\cC_1 + {\kappa_{2,\tilde{B}}})^2\|\tilde{\Delta}\bx_k\|^2 - \rbr{\sigma-\frac{8}{\nu\kappa_{1,\tilde{G}}}}\nbr{c_k}^2 - \frac{\nu\kappa_{1,\tilde{G}}}{8}\|\tilde{\Delta}\blambda_k\|^2 \\
&\quad - \frac{\nu}{4}\|\tilde{B}_k\tilde{\Delta}\bx_k + \bar{\nabla}_{\bx}\cL_k\|^2.
\end{align*}
The last squared term contains the stationarity residual. We lower bound it by a multiple of $\|\bar{\nabla}_{\bx}\cL_k\|^2$, at the cost of an additional primal-step term. Specifically, by Assumption \ref{assumption:boundedness_1},
\begin{align*}
\|\tilde{B}_k\tilde{\Delta}\bx_k + \bar{\nabla}_{\bx}\cL_k\|^2 &= \frac{\|\bar{\nabla}_{\bx}\cL_k\|^2}{2} - \|\tilde{B}_k\tilde{\Delta}\bx_k\|^2 + \nbr{\frac{\bar{\nabla}_{\bx}\cL_k}{\sqrt{2}} + \sqrt{2}\tilde{B}_k\tilde{\Delta}\bx_k}^2\\
&\ge \frac{\|\bar{\nabla}_{\bx}\cL_k\|^2}{2} - \|\tilde{B}_k\tilde{\Delta}\bx_k\|^2 \ge \frac{\|\bar{\nabla}_{\bx}\cL_k\|^2}{2} - \kappa_{2,\tilde{B}}^2\|\tilde{\Delta}\bx_k\|^2.
\end{align*}
Combining the above two displays, we obtain 
\begin{align}\label{ns:equ:1}
-\binom{\tilde{\Delta}\bx_k}{\tilde{\Delta}\blambda_k}^\top &
\begin{pmatrix}
\tilde{B}_k + \nu\nabla^2_{\bx}\cL_k\tilde{B}_k + \sigma \tilde{G}_k^\top\tilde{G}_k &\tilde{G}_k^\top + \nu\nabla^2_{\bx}\cL_k\tilde{G}_k^\top\\
\tilde{G}_k + \nu\tilde{G}_k\tilde{B}_k &\nu\tilde{G}_k\tilde{G}_k^\top
\end{pmatrix}
\binom{\tilde{\Delta}\bx_k}{\tilde{\Delta}\blambda_k} \nonumber\\
&\le -\tilde{\Delta}\bx_k^\top \tilde{B}_k\tilde{\Delta}\bx_k + 2\nu(\cC_1 + {\kappa_{2,\tilde{B}}})^2\|\tilde{\Delta}\bx_k\|^2 - \rbr{\sigma - \frac{8}{\nu\kappa_{1,\tilde{G}}}}\|c_k\|^2 - \frac{\nu\kappa_{1,\tilde{G}}}{8}\|\tilde{\Delta}\blambda_k\|^2 \nonumber\\
&\quad - \frac{\nu}{8}\|\nabla_{\bx}\cL_k\|^2 + \frac{\nu}{8}\|\nabla_{\bx}\cL_k +\bar{\nabla}_{\bx}\cL_k\|\|\nabla_{\bx}\cL_k - \bar{\nabla}_{\bx}\cL_k\| \nonumber\\
&\le -\tilde{\Delta}\bx_k^\top \tilde{B}_k\tilde{\Delta}\bx_k + 2\nu(\cC_1 + {\kappa_{2,\tilde{B}}})^2\|\tilde{\Delta}\bx_k\|^2 - \rbr{\sigma - \frac{8}{\nu\kappa_{1,\tilde{G}}}}\|c_k\|^2 - \frac{\nu\kappa_{1,\tilde{G}}}{8}\|\tilde{\Delta}\blambda_k\|^2 \nonumber\\
&\quad - \frac{\nu}{8}\|\nabla_{\bx}\cL_k\|^2 + \nu\cC_4(1+ \|\bar \bg_k\|)\|\nabla_{\bx}\cL_k - \bar{\nabla}_{\bx}\cL_k\|,
\end{align}
where the existence of $\cC_4$ in the last inequality is due to Assumption \ref{assumption:boundedness_1}. To handle the first two~primal-step terms, we decompose $\tilde{\Delta}\bx_k$ into two subspaces as $\tilde{\Delta}\bx_k = \bu_k + \bv_k$, where $\bu_k\in \Span{(\tilde{G}_k^\top)}$ and $\tilde{G}_k\bv_k = \zero$.
Then, we obtain
\begin{align*}
-\tilde{\Delta}\bx_k^\top\tilde{B}_k\tilde{\Delta}\bx_k & + 2\nu(\cC_1 + {\kappa_{2,\tilde{B}}})^2\|\tilde{\Delta}\bx_k\|^2\\
&=-\bu_k^\top\tilde{B}_k\bu_k - 2\bu_k^\top\tilde{B}_k\bv_k - \bv_k^\top\tilde{B}_k\bv_k + 2\nu(\cC_1 + {\kappa_{2,\tilde{B}}})^2\|\tilde{\Delta}\bx_k\|^2\\
&\le -\kappa_{1,\tilde{B}}\|\bv_k\|^2 + {\kappa_{2,\tilde{B}}}\|\bu_k\|^2 + \frac{\kappa_{1,\tilde{B}}}{2}\|\bv_k\|^2 + \frac{2\kappa^2_{2,\tilde{B}}}{\kappa_{1,\tilde{B}}}\|\bu_k\|^2 + 2\nu(\cC_1 + {\kappa_{2,\tilde{B}}})^2\|\tilde{\Delta}\bx_k\|^2\\
&= -\frac{\kappa_{1,\tilde{B}}}{2}\|\bv_k\|^2 + \rbr{{\kappa_{2,\tilde{B}}} + \frac{2\kappa^2_{2,\tilde{B}}}{\kappa_{1,\tilde{B}}}}\|\bu_k\|^2 + 2\nu(\cC_1 + {\kappa_{2,\tilde{B}}})^2\|\tilde{\Delta}\bx_k\|^2\\
&=\rbr{{\kappa_{2,\tilde{B}}} + \frac{2\kappa^2_{2,\tilde{B}}}{\kappa_{1,\tilde{B}}} + \frac{\kappa_{1,\tilde{B}}}{2}}\|\bu_k\|^2 - \rbr{\frac{\kappa_{1,\tilde{B}}}{2} - 2\nu(\cC_1 + {\kappa_{2,\tilde{B}}})^2}\|\tilde{\Delta}\bx_k\|^2,
\end{align*}
where the inequality is due to Assumption \ref{assumption:boundedness_1} and the last equality uses the fact that $\|\tilde{\Delta}\bx_k\|^2 = \|\bu_k\|^2 + \|\bv_k\|^2$. Note that 
\begin{equation*}
\|c_k\|^2 = \|\tilde{G}_k\tilde{\Delta}\bx_k\|^2 = \|\tilde{G}_k\bu_k\|^2\ge \kappa_{1,\tilde{G}}\|\bu_k\|^2.
\end{equation*}
The above two displays lead to
\begin{multline*}
-\tilde{\Delta}\bx_k^\top \tilde{B}_k\tilde{\Delta}\bx_k + 2\nu(\cC_1 + {\kappa_{2,\tilde{B}}})^2\|\tilde{\Delta}\bx_k\|^2\\
\le \frac{1}{\kappa_{1,\tilde{G}}}\rbr{{\kappa_{2,\tilde{B}}} + \frac{2\kappa^2_{2,\tilde{B}}}{\kappa_{1,\tilde{B}}} + \frac{\kappa_{1,\tilde{B}}}{2}}\|c_k\|^2 - \rbr{\frac{\kappa_{1,\tilde{B}}}{2} - 2\nu(\cC_1 + {\kappa_{2,\tilde{B}}})^2}\|\tilde{\Delta}\bx_k\|^2.
\end{multline*}
Plugging the above display into \eqref{ns:equ:2} and \eqref{ns:equ:1}, we obtain
\begin{align*}
&(\nabla \cL_{\sigma, \nu}^k)^\top \tilde{\bz}_k\\
&\le \frac{1}{\kappa_{1,\tilde{G}}}\rbr{{\kappa_{2,\tilde{B}}} + \frac{2\kappa^2_{2,\tilde{B}}}{\kappa_{1,\tilde{B}}} + \frac{\kappa_{1,\tilde{B}}}{2}}\|c_k\|^2 - \rbr{\frac{\kappa_{1,\tilde{B}}}{2} - 2\nu(\cC_1 + {\kappa_{2,\tilde{B}}})^2}\|\tilde{\Delta}\bx_k\|^2\\
&\quad - \rbr{\sigma - \frac{8}{\nu\kappa_{1,\tilde{G}}}}\|c_k\|^2 - \frac{\nu\kappa_{1,\tilde{G}}}{8}\|\tilde{\Delta}\blambda_k\|^2- \frac{\nu}{8}\|\nabla_{\bx}\cL_k\|^2 + \nu\cC_4(1+ \|\bar \bg_k\|)\|\nabla_{\bx}\cL_k - \bar{\nabla}_{\bx}\cL_k\|\\
&\quad + \cC_2\|\nabla_{\bx}\cL_k - \bar{\nabla}_{\bx}\cL_k\|\|\tilde{\bz}_k\| + \cC_3\|\tilde{G}_k - G_k\|\|\tilde{\bz}_k\|\\
&\le - \frac{\nu\kappa_{1,\tilde{G}}}{8}\|\tilde{\bz}_k\|^2 - \frac{\nu}{8}\|\nabla\cL_k\|^2 - \rbr{\sigma - \frac{1}{\kappa_{1,\tilde{G}}}({\kappa_{2,\tilde{B}}} + \frac{2{\kappa^2_{2,\tilde{B}}}}{\kappa_{1,\tilde{B}}} + \frac{\kappa_{1,\tilde{B}}}{2} + \frac{8}{\nu}) - \frac{\nu}{8}}\|c_k\|^2\\
&\quad - \rbr{\frac{\kappa_{1,\tilde{B}}}{2} - 2\nu(\cC_1+{\kappa_{2,\tilde{B}}})^2 - \frac{\nu\kappa_{1,\tilde{G}}}{8}}\|\tilde{\Delta}\bx_k\|^2 + \cC_2\|\nabla_{\bx}\cL_k - \bar{\nabla}_{\bx}\cL_k\|\|\tilde{\bz}_k\| + \cC_3\|\tilde{G}_k - G_k\|\|\tilde{\bz}_k\| \\
&\quad + \nu\cC_4(1+ \|\bar \bg_k\|)\|\nabla_{\bx}\cL_k - \bar{\nabla}_{\bx}\cL_k\|.
\end{align*}
Then, it is clear that we can choose $(\nu, \sigma)$ such that 
\begin{equation*}
\frac{\kappa_{1,\tilde{B}}}{2} - 2\nu(\cC_1+{\kappa_{2,\tilde{B}}})^2 - \frac{\nu\kappa_{1,\tilde{G}}}{8} > 0, \qquad \sigma - \frac{1}{\kappa_{1,\tilde{G}}}\rbr{{\kappa_{2,\tilde{B}}} + \frac{2{\kappa^2_{2,\tilde{B}}}}{\kappa_{1,\tilde{B}}} + \frac{\kappa_{1,\tilde{B}}}{2} + \frac{8}{\nu}} - \frac{\nu}{8} > 0.
\end{equation*}
With these choices, we have
\begin{align}
(\nabla \cL_{\sigma, \nu}^k)^\top \tilde{\bz}_k
&\le -2\gamma_1(\|\tilde{\bz}_k\|^2 + \|\nabla\cL_k\|^2) \label{eq:directional_reduction_1}\\
&\quad + \cC_2\|\nabla_{\bx}\cL_k - \bar{\nabla}_{\bx}\cL_k\|\|\tilde{\bz}_k\| + \cC_3\|\tilde{G}_k - G_k\|\|\tilde{\bz}_k\| + \nu\cC_4(1+ \|\bar \bg_k\|)\|\nabla_{\bx}\cL_k - \bar{\nabla}_{\bx}\cL_k\|, 
\end{align}
where $\gamma_1 := \nu\min\{\kappa_{1,\tilde{G}}, 1\}/16$. It remains to rewrite the derivative estimation error in terms of the gradient and Jacobian tracking errors that were bounded earlier. We have
\begin{equation}
\|\nabla_{\bx}\cL_k - \bar{\nabla}_{\bx}\cL_k\| \stackrel{\eqref{eq:debiased_Lagrangian_Hessian}}{=} \|(\nabla f_k - \bar \bg_k) + (G_k - \tilde{G}_k)^\top\blambda_k\| \le \|\nabla f_k - \bar \bg_k\| + \Upsilon_{\blambda}\|G_k - \tilde{G}_k\|. \label{eq:bar_cL_and_cL_difference}
\end{equation}
Then 
\begin{align}
&\cC_2\|\nabla_{\bx}\cL_k - \bar{\nabla}_{\bx}\cL_k\|\|\tilde{\bz}_k\| + \cC_3\|\tilde{G}_k - G_k\|\|\tilde{\bz}_k\| + \nu\cC_4(1+ \|\bar \bg_k\|)\|\nabla_{\bx}\cL_k - \bar{\nabla}_{\bx}\cL_k\|\\
&\le (\cC_2\|\tilde{\bz}_k\| + \nu\cC_4(1+ \|\bar \bg_k\|))\|\nabla f_k - \bar \bg_k\| + (\Upsilon_{\blambda}(\cC_2\|\tilde{\bz}_k\| + \nu\cC_4(1+ \|\bar \bg_k\|))+ \cC_3\|\tilde{\bz}_k\|)\|G_k - \tilde{G}_k\|\\
&\le ((\Upsilon_{\blambda}+1)(\cC_2\|\tilde{\bz}_k\| + \nu\cC_4(1+ \|\bar \bg_k\|)) + \cC_3\|\tilde{\bz}_k\|)(\|\nabla f_k - \bar \bg_k\| + \|G_k - \tilde{G}_k\|)\\
&\le \cC_5(\|\tilde{\bz}_k\| + \|\bar \bg_k\| + 1)(\|\nabla f_k - \bar \bg_k\| + \|G_k - \tilde{G}_k\|)\\
&\le \cC_6(\|\tilde{\bz}_k\| + \|\nabla f_k - \bar \bg_k\| + 1)(\|\nabla f_k - \bar \bg_k\| + \|G_k - \tilde{G}_k\|)\\
&\le \gamma_1\|\tilde{\bz}_k\|^2 + \cC_7(\|\nabla f_k - \bar{\bg}_k\| + \|G_k - \tilde{G}_k\| + \|\nabla f_k - \bar{\bg}_k\|^2+ \|G_k - \tilde{G}_k\|^2)\label{eq:directional_reduction_remainder_bound}
\end{align}
for some large constants $\cC_5, \cC_6, \cC_7>0$. Finally, combining \eqref{eq:directional_reduction_1} and \eqref{eq:directional_reduction_remainder_bound} and letting $\Upsilon_1\leftarrow \cC_7$, we complete the proof.

\subsection{Proof of Lemma \ref{lemma2:global_convergence}}\label{proof_of_lemma2:global_convergence}

We apply a one-step Taylor expansion of the merit function along the sketched direction:
\begin{equation*}
\cL_{\sigma,\nu}^{k+1} \le \cL_{\sigma,\nu}^{k} + \bar{\alpha}_k(\nabla\cL_{\sigma,\nu}^k)^\top\bz_{k,s} + \frac{\Upsilon_L}{2}\bar{\alpha}_k^2\nbr{\bz_{k,s}}^2,
\end{equation*}
where $\Upsilon_L$ is the Lipschitz constant of $\nabla\cL_{\sigma, \nu}$, and its existence is ensured by \eqref{eq:cal_L_sigma_nu_definition} and Assumption~\ref{assumption:boundedness_1}. For the above expansion, we apply $(\nabla\cL_{\sigma,\nu}^{k})^\top \bz_{k,s} = (\nabla\cL_{\sigma,\nu}^{k})^\top\tilde{\bz}_{k} + (\nabla\cL_{\sigma,\nu}^{k})^\top(\bz_{k,s} - \tilde{\bz}_{k})$ and have
\begin{align}\label{eq:cL_sigma_nu_reduction_step1}
&\EE[\cL_{\sigma,\nu}^{k+1}\given \cF_{k-2/3}] \\
&\le \cL_{\sigma,\nu}^{k} + \EE[\bar{\alpha}_k\given \cF_{k-2/3}](\nabla\cL_{\sigma,\nu}^{k})^\top\tilde{\bz}_{k} +\|\nabla\cL_{\sigma,\nu}^{k}\|\EE[\bar{\alpha}_k\|\bz_{k,s} - \tilde{\bz}_{k}\|\given \cF_{k-2/3}] + \frac{\Upsilon_L}{2}\EE[\bar{\alpha}^2_k\|\bz_{k,s}\|^2\given \cF_{k-2/3}]. \nonumber
\end{align}
Note that by Assumption \ref{assumption:boundedness_1} (recall that $\cC_1, \cC_2, \ldots$ denote generic deterministic constants throughout the section), 
\begin{multline*}
\|\nabla \cL_{\sigma,\nu}^k\| \le \nbr{\begin{pmatrix}
I + \nu\nabla_{\bx}^2\cL_k &\sigma G_k^\top\\
\nu G_k & I
\end{pmatrix}}\nbr{\binom{\nabla_{\bx}\cL_k}{c_k}} \\ \le (1+\nu(1+\sqrt{m}\Upsilon_{\blambda})\Upsilon_u + (\sigma+\nu)\sqrt{\kappa_{2,G}})\|\nabla \cL_k\| \eqqcolon \cC_1 \|\nabla \cL_k\|.
\end{multline*}
Applying Lemma~\ref{pre_lemma:sketching_1} under Assumption~\ref{assumption:sketching}, we then obtain 
\begin{align*}
&\|\nabla\cL_{\sigma,\nu}^{k}\|\EE[\bar{\alpha}_k\|\bz_{k,s} - \tilde{\bz}_{k}\|\given \cF_{k-2/3}]\\
&\le \cC_1\kappa_{3,\mu}\alpha_k\|\nabla \cL_k\|\EE[\|\bz_{k,s} - \tilde{\bz}_k\|\given \cF_{k-2/3}]\le \cC_1\kappa_{3,\mu}\alpha_k\|\nabla \cL_k\|\rbr{\EE[\|\bz_{k,s} - \tilde{\bz}_k\|^2\given \cF_{k-2/3}]}^{1/2}\\
&\le \cC_1\kappa_{3,\mu}\alpha_k\rho^{s/2}\|\nabla \cL_k\|\|\tilde{\bz}_k\|\le \cC_1\kappa_{3,\mu}\alpha_k\rho^{s/2}(\|\nabla \cL_k\|^2 + \|\tilde{\bz}_k\|^2),
\end{align*}
where the first inequality is by \eqref{eq:bar_alpha_bound} and the second last inequality is by Lemma~\ref{pre_lemma:sketching_1}. We now combine the above display with \eqref{eq:cL_sigma_nu_reduction_step1} and obtain
\begin{align*}
&\EE[\cL_{\sigma,\nu}^{k+1}\given \cF_{k-2/3}]\\
&\stackrel{\mathclap{\eqref{eq:bar_alpha_bound}}}{\leq} \cL_{\sigma,\nu}^{k} + \EE[\bar{\alpha}_k\given \cF_{k-2/3}](\nabla\cL_{\sigma,\nu}^{k})^\top\tilde{\bz}_{k} +\cC_1\kappa_{3,\mu}\alpha_k\rho^{s/2}(\|\nabla \cL_k\|^2 + \|\tilde{\bz}_k\|^2)\\
&\quad + {\Upsilon_L}\kappa^2_{3,\mu}\alpha^2_k\EE[\|\bz_{k,s}\|^2\given \cF_{k-2/3}]\\
&\le \cL_{\sigma,\nu}^{k} - \kappa_{1,\mu}\gamma_1\alpha_k(\|\tilde{\bz}_k\|^2 + \|\nabla\cL_k\|^2)\\
&\quad + \kappa_{3,\mu}\alpha_k\Upsilon_1(\|\nabla f_k - \bar{\bg}_k\| + \|G_k - \tilde{G}_k\|+ \|\nabla f_k - \bar{\bg}_k\|^2 + \|G_k - \tilde{G}_k\|^2)\\
&\quad +\cC_1\kappa_{3,\mu}\alpha_k\rho^{s/2}(\|\nabla \cL_k\|^2 + \|\tilde{\bz}_k\|^2) + {\Upsilon_L}\kappa^2_{3,\mu}\alpha^2_k\EE[\|\bz_{k,s}\|^2\given \cF_{k-2/3}]\\
&\le \cL_{\sigma,\nu}^{k} -  (\kappa_{1,\mu}\gamma_1 - \cC_1\kappa_{3,\mu}\rho^{s/2} - 4\Upsilon_L\kappa_{3,\mu}^2\alpha_k)\alpha_k(\|\tilde{\bz}_k\|^2 + \|\nabla\cL_k\|^2)\\
&\quad + \kappa_{3,\mu}\alpha_k\Upsilon_1(\|\nabla f_k - \bar{\bg}_k\| + \|G_k - \tilde{G}_k\|+ \|\nabla f_k - \bar{\bg}_k\|^2 + \|G_k - \tilde{G}_k\|^2),
\end{align*}
where the second inequality follows from Lemma~\ref{lemma:global_convergence} and the conditions \eqref{eq:bar_alpha_bound}, \eqref{eq:stepsize_selection}, and Assumption~\ref{assumption:stepsize_selection} for bounding $\EE[\bar{\alpha}_k\given \cF_{k-2/3}]$, and the last inequality is due to Lemma \ref{pre_lemma:sketching_1}(a).
When $K'$ is large enough, we have $4\Upsilon_L\kappa_{3,\mu}^2\alpha_k \le \kappa_{1,\mu}\gamma_1/2$ for $k\ge K'$. Choosing the number of sketching steps $s$ large enough such that $\kappa_{1,\mu}\gamma_1/(2\kappa_{3,\mu}\cC_1)  > \rho^{s/2}$ with $\rho = 1-\gamma_S$, we can define
\begin{equation*}
\gamma_2 := \kappa_{1,\mu}\gamma_1/2 - \cC_1\kappa_{3,\mu}\rho^{s/2} \qquad\text{ and }\qquad \Upsilon_2 := \kappa_{3,\mu}\Upsilon_1,
\end{equation*}
and this completes the proof.

\subsection{Proof of Theorem \ref{thm:global_convergence}}\label{proof_of_thm:global_convergence}

In this proof, we let $k \ge K'$ for the natural number $K'$ introduced in Lemma~\ref{lemma2:global_convergence}. From Lemma~\ref{lemma2:global_convergence}, under the event $\{\tilde{G}_k = \bar{G}_k\} \in \cF_{k-2/3}$, we have
\begin{align}
&\EE[\cL_{\sigma,\nu}^{k+1}\1_{\delta^G_k= \zero}\given \cF_{k-2/3}] \\
&\le \cL_{\sigma,\nu}^k\1_{\delta^G_k= \zero} - \gamma_2\alpha_k(\|\tilde{\bz}_k\|^2 + \|\nabla\cL_k\|^2)\1_{\delta^G_k= \zero}\\
&\quad + \Upsilon_2\alpha_k(\|\nabla f_k - \bar{\bg}_k\| + \|G_k - \bar{G}_k\| + \|\nabla f_k - \bar{\bg}_k\|^2 + \|G_k - \bar{G}_k\|^2)\1_{\delta^G_k= \zero}\\
&\leq \cL_{\sigma,\nu}^k\1_{\delta^G_{k-1}= \zero} - \gamma_2\alpha_k(\|\tilde{\bz}_k\|^2 + \|\nabla\cL_k\|^2)\1_{\delta^G_k= \zero}\\
&\quad + \Upsilon_2\alpha_k(\|\nabla f_k - \bar{\bg}_k\| + \|G_k - \bar{G}_k\| + \|\nabla f_k - \bar{\bg}_k\|^2 + \|G_k - \bar{G}_k\|^2)\1_{\delta^G_k= \zero}\\
&\quad +  |\cL_{\sigma,\nu}^k(\1_{\delta^G_k= \zero} -\1_{\delta^G_{k-1}= \zero})|. \label{eq:RS_theorem_inequality}
\end{align}
Our goal is to prove that $\sum_{k=K'}^\infty\alpha_k(\|\tilde{\bz}_k\|^2 + \|\nabla\cL_k\|^2) < \infty$ using Robbins–Siegmund theorem. To this end, we first show that, almost surely,
\begin{equation}
\sum_{k=K'}^\infty  \Upsilon_2\alpha_k(\|\nabla f_k - \bar{\bg}_k\| + \|G_k - \bar{G}_k\| + \|\nabla f_k - \bar{\bg}_k\|^2 + \|G_k - \bar{G}_k\|^2)\1_{\delta^G_k= \zero} < \infty,\label{eq:Robbins–Siegmund_theorem_series_convergence_1}
\end{equation}
and 
\begin{equation}
\sum_{k=K'}^\infty |\cL_{\sigma,\nu}^k(\1_{\delta^G_k= \zero} -\1_{\delta^G_{k-1}= \zero})| < \infty. \label{eq:Robbins–Siegmund_theorem_series_convergence_2}
\end{equation}
For \eqref{eq:Robbins–Siegmund_theorem_series_convergence_1}, by Lemma~\ref{lemma:gradient_approx_rate} (noting that \eqref{eq:parameter_conditions:thm:global_convergence} implies \eqref{eq:parameter_conditions:lemma:gradient_approx_rate}), we have
\begin{equation}
\EE[\|\nabla f_k - \bar{\bg}_k\|^2 + \|G_k - \bar{G}_k\|^2] \le \cC_1({\beta_k} + b_k^4 + \alpha_k^2/\beta_k^2), \label{eq:RS_theorem_error_bound_1}
\end{equation}
and then
\begin{equation}
\EE[\|\nabla f_k - \bar{\bg}_k\| + \|G_k - \bar{G}_k\|] \le (\EE[(\|\nabla f_k - \bar{\bg}_k\| + \|G_k - \bar{G}_k\|)^2])^{1/2} \le \cC_2(\sqrt{\beta_k} + b_k^2 + \alpha_k/\beta_k).\label{eq:RS_theorem_error_bound_2}
\end{equation}
Then taking expectation to the left-hand side of \eqref{eq:Robbins–Siegmund_theorem_series_convergence_1}, enlarging it by removing indicators, and applying Tonelli's theorem, we obtain that
\begin{align*}
&\EE\sbr{\sum_{k=K'}^\infty  \Upsilon_2\alpha_k(\|\nabla f_k - \bar{\bg}_k\| + \|G_k - \bar{G}_k\| + \|\nabla f_k - \bar{\bg}_k\|^2 + \|G_k - \bar{G}_k\|^2)\1_{\delta^G_k= \zero}}\\
&\leq \Upsilon_2\sum_{k=K'}^\infty\alpha_k\EE[\|\nabla f_k - \bar{\bg}_k\| + \|G_k - \bar{G}_k\| + \|\nabla f_k - \bar{\bg}_k\|^2 + \|G_k - \bar{G}_k\|^2]\\
&\leq \cC_3\sum_{k=K'}^\infty(\alpha_k^2 + \alpha_k\sqrt{\beta_k} + \alpha_k b_k^2 + \alpha_k^2/\beta_k),
\end{align*}
where the last inequality combines \eqref{eq:RS_theorem_error_bound_1} and \eqref{eq:RS_theorem_error_bound_2}, and absorbs the constant $\Upsilon_2$ into the generic constant $\cC_3$. By our assumptions on $p_1, p_2, p_3$, we have $\sum_{k=K'}^\infty(\alpha_k^2 + \alpha_k\sqrt{\beta_k} + \alpha_k b_k^2 + \alpha_k^2/\beta_k) < \infty$,
which implies \eqref{eq:Robbins–Siegmund_theorem_series_convergence_1}. For \eqref{eq:Robbins–Siegmund_theorem_series_convergence_2}, under the condition that $\delta_k^G$ ensures $[\kappa_{1,G}, \kappa_{2,G}] \subseteq (\kappa_{1,\tilde{G}}, \kappa_{2,\tilde{G}})$, by Lemma~\ref{lemma:gradient_approx_convergence}, we have that $\delta_k^G = \zero$ for $k\ge K_G^\star$.  Then the left-hand side of \eqref{eq:Robbins–Siegmund_theorem_series_convergence_2} becomes 
\begin{equation*}
\sum_{k=K'}^{K_G^\star}|\cL_{\sigma,\nu}^k(\1_{\delta^G_k= \zero} -\1_{\delta^G_{k-1}= \zero})| \leq 2\sum_{k=K'}^{K_G^\star}|\cL_{\sigma,\nu}^k| < \infty,
\end{equation*}
where the last inequality comes from the finiteness of $K_G^\star$ and $\cL_{\sigma,\nu}^k$ and holds almost surely. This shows \eqref{eq:Robbins–Siegmund_theorem_series_convergence_2}. To utilize Robbins–Siegmund theorem, we need to change the sequence $\{\cL_{\sigma,\nu}^k\1_{\delta^G_{k-1}= \zero}\}$ in \eqref{eq:RS_theorem_inequality} to some non-negative sequence. To this end, we need to find a lower bound of such sequence~and subtract it from both sides of the inequality \eqref{eq:RS_theorem_inequality}.  By \eqref{eq:cal_L_sigma_nu_definition} and Assumption~\ref{assumption:boundedness_1}, we have $\cL_{\sigma,\nu}^k \ge \cL(\bx_k,\blambda_k) = f(\bx_k) + \blambda_k^\top c(\bx_k)\ge f_{\rm inf} - \Upsilon_{\blambda}\kappa_c$. Therefore, the lower bound of $\cL_{\sigma,\nu}^k\1_{\delta^G_{k-1}= \zero}$ is given by $\min\{f_{\rm inf} - \Upsilon_{\blambda}\kappa_c, 0\}$. Now, we are able to apply Robbins–Siegmund theorem to obtain that, almost surely, $\sum_{k=K'}^\infty \alpha_k(\|\tilde{\bz}_k\|^2 + \|\nabla\cL_k\|^2)\1_{\delta^G_k= \zero} < \infty$.
Since
\begin{align*}
\sum_{k=K'}^\infty \alpha_k(\|\tilde{\bz}_k\|^2 + \|\nabla\cL_k\|^2) = \sum_{k=K'}^\infty \alpha_k(\|\tilde{\bz}_k\|^2 + \|\nabla\cL_k\|^2)\1_{\delta^G_k= \zero} + \sum_{k=K'}^{K_G^\star - 1}\alpha_k(\|\tilde{\bz}_k\|^2 + \|\nabla\cL_k\|^2)\1_{\delta^G_k\neq \zero},
\end{align*}
we have $\sum_{k=K'}^\infty \alpha_k(\|\tilde{\bz}_k\|^2 + \|\nabla\cL_k\|^2) < \infty$ almost surely. To upgrade summability to convergence of the residual itself, we note that
\begin{align*}
&|\|\nabla\cL_{k+\tau}\| - \|\nabla \cL_{k}\||\le \nbr{\nabla \cL_{k+\tau} -\nabla \cL_{k}}\le \cC_4\nbr{(\bx_{k+\tau}, \blambda_{k+\tau}) - (\bx_k, \blambda_k)}= \cC_4\nbr{\sum_{i=k}^{k+\tau-1}\bar{\alpha}_i{\bz}_{i,s}}\\
&\le\cC_4\nbr{\sum_{i=k}^{k+\tau - 1}\bar{\alpha}_i(I-\tilde{C}_i)\tilde{\bz}_i}\stackrel{\eqref{equ:KKT}}{=} \cC_4\nbr{\sum_{i=k}^{k+\tau - 1}\bar{\alpha}_i(I-\tilde{C}_i)\tilde{W}_i^{-1}\bar{\nabla}\cL_i}\\
&\le \cC_4\nbr{\sum_{i=k}^{k+\tau - 1}\bar{\alpha}_i(I-\tilde{C}_i)\tilde{W}_i^{-1}{\nabla}\cL_i} + \cC_4\nbr{\sum_{i=k}^{k+\tau - 1}\bar{\alpha}_i(I-\tilde{C}_i)\tilde{W}_i^{-1}(\bar{\nabla}\cL_i - \nabla\cL_i)}\\
&\le 2\cC_4\kappa_{3,\mu}\Upsilon_W\sum_{i=k}^{k+\tau-1}\alpha_i\nbr{\nabla\cL_i} + \cC_4\nbr{\sum_{i=k}^{k+\tau-1}\bar{\alpha}_i (I-\tilde{C}_i)\tilde{W}_i^{-1}(\bar{\nabla}\cL_i - \nabla \cL_i)},
\end{align*}
where the fourth inequality is by Lemma \ref{pre_lemma:sketching_1}(a) and the last inequality is due to Lemma \ref{pre_lemma:sketching_1}(a),~\eqref{eq:bar_alpha_bound}, and the fact that $\|\tilde{W}_k^{-1}\| \le \Upsilon_W$. In order to use Lemma~\ref{pre_lemma:convergence_to_zero} to conclude that $\|\nabla\cL_k\|$ converges to $0$, we need to show that $\|\sum_{i=k}^\infty \bar{\alpha}_i(I-\tilde{C}_i)\tilde{W}_i^{-1}(\bar{\nabla}\cL_i - \nabla \cL_i)\| < \infty$ almost surely.
By \eqref{eq:bar_alpha_bound}, we have
\begin{equation}\label{eq:thm:global_convergence:summability}
\nbr{\sum_{i=k}^\infty \bar{\alpha}_i(I-\tilde{C}_i)\tilde{W}_i^{-1}(\bar{\nabla}\cL_i - \nabla \cL_i)} \le 2\kappa_{3,\mu}\Upsilon_W\sum_{i=k}^\infty\alpha_i(\|\bar{\bg}_i - \nabla f_i\| + \|\blambda_i\|\|\tilde{G}_i - {G}_i\|).
\end{equation}
By Lemma~\ref{lemma:gradient_approx_rate} and Jensen's inequality, we have 
\begin{equation}
\EE\sbr{\sum_{i=k}^\infty\alpha_i\|\bar{\bg}_i - \nabla f_i\|}\le \sum_{i=k}^\infty\alpha_i(\EE[\|\bar{\bg}_i - \nabla f_i\|^2])^{1/2} \le \sum_{i=k}^\infty\alpha_i(\beta_i + b_i^4 + \alpha_i^2/\beta_i^2)^{1/2}. \label{eq:thm:global_convergence:summability_in_expectation_1}
\end{equation}
By our conditions on $p_1, p_2, p_3$, we know the right-hand side of \eqref{eq:thm:global_convergence:summability_in_expectation_1} is finite. Therefore, $\sum_{i=k}^\infty\alpha_i\|\bar{\bg}_i - \nabla f_i\|$ is finite almost surely. For the remaining part in \eqref{eq:thm:global_convergence:summability}, we have
\begin{equation*}
\sum_{i=k}^\infty\alpha_i\|\blambda_i\|\|\tilde{G}_i - {G}_i\| \le \Upsilon_{\blambda}\sum_{i=k}^{K_G^
\star-1}\alpha_i\|\tilde{G}_i - {G}_i\| + \Upsilon_{\blambda}\sum_{i=K_G^
\star}^\infty\alpha_i\|\bar{G}_i - {G}_i\|,
\end{equation*}
and both terms are finite almost surely by the same argument as in \eqref{eq:thm:global_convergence:summability_in_expectation_1}. Thus, the conditions of Lemma~\ref{pre_lemma:convergence_to_zero} are satisfied, so $\|\nabla\cL_k\|\to 0$. Finally, we note that
\begin{align*}
\EE\sbr{\sum_{k=K'}^\infty\|(\bx_{k+1}, \blambda_{k+1}) - (\bx_k, \blambda_k)\|^2} = \sum_{k=K'}^\infty \EE[\bar{\alpha}_k^2\|\bz_{k,s}\|^2]\le \cC_5\sum_{k=K'}^\infty\alpha_k^2 < \infty,
\end{align*}
where the second inequality is by combining the analysis in \eqref{eq:nabla_f_step_difference_bound} and \eqref{eq:bar_g_r_moment_bound} to show $\EE[\|\bz_{k,s}\|^2]$ is~uniformly bounded. The above result suggests that $\nbr{(\bx_{k+1}, \blambda_{k+1}) - (\bx_k, \blambda_k)}$ converges to $0$ almost surely. We complete the proof.

\section{Proofs of Section \ref{sec:local_convergence}} 

\subsection{Proof of Lemma \ref{lemma:Hessian_approx_convergence}}\label{proof_of_lemma:Hessian_approx_convergence}

Recall that
$\hat{\nabla}_{\bx}^2\cL_i
=\hat{\nabla}^2F(\bx_i;\xi_i)
+\sum_{j=1}^m\lambda_i^j\hat{\nabla}^2c_i^j$
and
$\bar B_k=(1-\beta_k)\bar B_{k-1}
+\beta_k\hat{\nabla}_{\bx}^2\cL_k$.
Iterating this recursion gives
\begin{align}
\bar B_k-\nabla_{\bx}^2\cL^\star
&=\prod_{\ell=0}^k(1-\beta_\ell)
(\bar B_{-1}-\nabla_{\bx}^2\cL^\star) +\sum_{i=0}^k\prod_{\ell=i+1}^k(1-\beta_\ell)\beta_i
\rbr{\EE[\hat{\nabla}_{\bx}^2\cL_i\given\cF_{i-1}]
-\nabla_{\bx}^2\cL^\star}\notag\\
&\quad+\sum_{i=0}^k\prod_{\ell=i+1}^k(1-\beta_\ell)\beta_i
\rbr{\hat{\nabla}_{\bx}^2\cL_i
-\EE[\hat{\nabla}_{\bx}^2\cL_i\given\cF_{i-1}]}.
\label{eq:1_proof_of_lemma:Hessian_approx_convergence}
\end{align}
Since $p_2\in(1/2,1)$ and $0<\beta_i\le1$, we have
$\sum_i\beta_i=\infty$, $\sum_i\beta_i^2<\infty$, and
$\prod_{\ell=0}^k(1-\beta_\ell)
(\bar B_{-1}-\nabla_{\bx}^2\cL^\star)\to\zero$
almost surely.

We first note that, for any matrix sequence $E_i\to\zero$
almost surely,
\begin{equation}
\sum_{i=0}^k\prod_{\ell=i+1}^k(1-\beta_\ell)\beta_iE_i
\to\zero
\qquad\text{almost surely}.
\label{eq:auxilliary_result_for_proof_of_lemma:local_convergence}
\end{equation}
Indeed, on any path where $E_i\to\zero$, choose $K_\varepsilon$
so that $\|E_i\|\le\varepsilon$ for $i\ge K_\varepsilon$.
For $k\ge K_\varepsilon$,
\begin{equation*}
\nbr{\sum_{i=0}^k\prod_{\ell=i+1}^k(1-\beta_\ell)\beta_iE_i}
\le
\sum_{i=0}^{K_\varepsilon-1}
\prod_{\ell=i+1}^k(1-\beta_\ell)\beta_i\|E_i\|
+\varepsilon,
\end{equation*}
where
$\sum_{i=K_\varepsilon}^k
\prod_{\ell=i+1}^k(1-\beta_\ell)\beta_i
=1-\prod_{\ell=K_\varepsilon}^k(1-\beta_\ell)\le1$.
Each product in the finite sum tends to zero.
Letting $k\to\infty$ and then $\varepsilon\downarrow0$ proves
\eqref{eq:auxilliary_result_for_proof_of_lemma:local_convergence}.

Since $\blambda_i$ is $\cF_{i-1}$-measurable and
$\|\blambda_i\|\le\Upsilon_{\blambda}$,
Lemma~\ref{lemma:approximation_rate} and
Assumption~\ref{assumption:boundedness_1} give
\begin{align}
\nbr{\EE[\hat{\nabla}_{\bx}^2\cL_i\given\cF_{i-1}]
-\nabla_{\bx}^2\cL^\star}&\le\cC_1(b_i+\tilde b_i^2/b_i)
+\|\nabla^2f_i-\nabla^2f(\bx^\star)\|
+\sum_{j=1}^m|\lambda_i^j|
\|\nabla^2c_i^j-\nabla^2c^j(\bx^\star)\|\notag\\
&\qquad+
\sum_{j=1}^m|\lambda_i^j-(\lambda^\star)^j|
\|\nabla^2c^j(\bx^\star)\|
\le\cC_2\rbr{
b_i+\tilde b_i^2/b_i+\|\btheta_i-\btheta^\star\|}.
\label{eq:Lagrangian_Hessian_mean_error}
\end{align}
By $2p_4>p_3$ and Assumption~\ref{assumption:LICQ_SOSC},
the bound in \eqref{eq:Lagrangian_Hessian_mean_error}
tends to zero almost surely.
Thus \eqref{eq:auxilliary_result_for_proof_of_lemma:local_convergence}
yields
\begin{equation*}
\sum_{i=0}^k\prod_{\ell=i+1}^k(1-\beta_\ell)\beta_i
\rbr{\EE[\hat{\nabla}_{\bx}^2\cL_i\given\cF_{i-1}]
-\nabla_{\bx}^2\cL^\star}
\to\zero
\qquad\text{almost surely}.
\end{equation*}

Choose a deterministic $K'$ such that
$\sqrt d\kappa_{\bDelta_2}(b_{K'}+\tilde b_{K'})
\le\delta_{\rm loc}/2$.
For $i\ge K'$, $\|\bx_i-\bx^\star\|\le\delta_{\rm loc}/2$,
$|s_1|\le b_i$, and $0\le s_2\le\tilde b_i$, we have
$\|\bx_i+s_1\bDelta_i+s_2\tilde{\bDelta}_i-\bx^\star\|
\le\delta_{\rm loc}$.
The fundamental theorem of calculus gives
\begin{align*}
\hat{\nabla}^2F(\bx_i;\xi_i)
=\frac{1}{2b_i\tilde b_i}
\int_0^{\tilde b_i}\int_{-b_i}^{b_i}
\bDelta_i^\top
\nabla^2F(\bx_i+s_1\bDelta_i+s_2\tilde{\bDelta}_i;\xi_i)
\tilde{\bDelta}_i\,ds_1ds_2\cdot
\frac{
\bDelta_i^{-1}\tilde{\bDelta}_i^{-\top}
+\tilde{\bDelta}_i^{-1}\bDelta_i^{-\top}}{2}.
\end{align*}
The same identity holds with $F(\cdot;\xi_i)$ replaced by $c^j$.
Hence Assumptions~\ref{assumption:boundedness_1},
\ref{assumption:boundedness_3}, and
\ref{assumption:Hessian_bound} imply, on
$\{\|\bx_i-\bx^\star\|\le\delta_{\rm loc}/2\}$,
\begin{equation*}
\|\hat{\nabla}^2F(\bx_i;\xi_i)\|_F\le\cC_3L(\xi_i),
\qquad
\max_{1\le j\le m}\|\hat{\nabla}^2c_i^j\|_F\le\cC_3,
\qquad
\|\hat{\nabla}_{\bx}^2\cL_i\|_F^2
\le\cC_4(L^2(\xi_i)+1).
\end{equation*}
The indicator
$\1_{\|\bx_i-\bx^\star\|\le\delta_{\rm loc}/2}$
is $\cF_{i-1}$-measurable and
$\hat{\nabla}_{\bx}^2\cL_i$ is $\cF_i$-measurable.
Conditional centering and the conditional variance inequality
therefore give, for $i\ge K'$,
\begin{align}
&\EE\sbr{
\1_{\|\bx_i-\bx^\star\|\le\delta_{\rm loc}/2}
\rbr{\hat{\nabla}_{\bx}^2\cL_i
-\EE[\hat{\nabla}_{\bx}^2\cL_i\given\cF_{i-1}]}
\given\cF_{i-1}}
=\zero,\notag\\
&\EE\sbr{
\nbr{
\1_{\|\bx_i-\bx^\star\|\le\delta_{\rm loc}/2}
\rbr{\hat{\nabla}_{\bx}^2\cL_i
-\EE[\hat{\nabla}_{\bx}^2\cL_i\given\cF_{i-1}]}
}_F^2
\given\cF_{i-1}}\notag\\
&\qquad\le
\EE\sbr{
\1_{\|\bx_i-\bx^\star\|\le\delta_{\rm loc}/2}
\|\hat{\nabla}_{\bx}^2\cL_i\|_F^2
\given\cF_{i-1}}
\le\cC_4(\EE[L^2(\xi_i)]+1)
\le\cC_5.
\label{eq:localized_Lagrangian_Hessian_moments}
\end{align}
Here $\xi_i$ is independent of $\cF_{i-1}$.
Apply Lemma~\ref{pre_lemma:as_convergence} entrywise,
starting at $K'$ with zero initial value, innovations
$\1_{\|\bx_i-\bx^\star\|\le\delta_{\rm loc}/2}
(\hat{\nabla}_{\bx}^2\cL_i
-\EE[\hat{\nabla}_{\bx}^2\cL_i\given\cF_{i-1}])$,
weights $\beta_i$, and pre-update filtration $\cF_{i-1}$.
The projection is onto the whole space and
$\bm{\eta}_i=\bb_i=\zero$.
Conditions (a) and (d) are then immediate,
(b) follows from \eqref{eq:localized_Lagrangian_Hessian_moments},
and (c) follows from
$\sum_i\beta_i^2<\infty$ and $\sum_i\beta_i=\infty$.
Consequently,
\begin{equation}
\sum_{i=K'}^k\prod_{\ell=i+1}^k(1-\beta_\ell)\beta_i
\1_{\|\bx_i-\bx^\star\|\le\delta_{\rm loc}/2}
\rbr{\hat{\nabla}_{\bx}^2\cL_i
-\EE[\hat{\nabla}_{\bx}^2\cL_i\given\cF_{i-1}]}
\to\zero
\quad\text{almost surely}.
\label{eq:localized_Lagrangian_Hessian_average}
\end{equation}
On almost every sample path, $\bx_i\to\bx^\star$,
so there is a finite $J\ge K'$ such that
$\|\bx_i-\bx^\star\|\le\delta_{\rm loc}/2$ for $i\ge J$.
On this path, for $k\ge J$,
\begin{align*}
&\nbr{
\begin{aligned}
&\sum_{i=0}^k\prod_{\ell=i+1}^k(1-\beta_\ell)\beta_i
\rbr{\hat{\nabla}_{\bx}^2\cL_i
-\EE[\hat{\nabla}_{\bx}^2\cL_i\given\cF_{i-1}]}\\
&\qquad-
\sum_{i=K'}^k\prod_{\ell=i+1}^k(1-\beta_\ell)\beta_i
\1_{\|\bx_i-\bx^\star\|\le\delta_{\rm loc}/2}
\rbr{\hat{\nabla}_{\bx}^2\cL_i
-\EE[\hat{\nabla}_{\bx}^2\cL_i\given\cF_{i-1}]}
\end{aligned}
}\\
&\le
\prod_{\ell=J}^k(1-\beta_\ell)
\sum_{i=0}^{J-1}\prod_{\ell=i+1}^{J-1}(1-\beta_\ell)\beta_i
\nbr{\hat{\nabla}_{\bx}^2\cL_i
-\EE[\hat{\nabla}_{\bx}^2\cL_i\given\cF_{i-1}]}
\to0.
\end{align*}
Together with
\eqref{eq:localized_Lagrangian_Hessian_average} and
\eqref{eq:1_proof_of_lemma:Hessian_approx_convergence},
this proves
$\bar B_k\to\nabla_{\bx}^2\cL^\star$ almost surely.

For the second part, fix a sample path on which
$\bar B_k\to\nabla_{\bx}^2\cL^\star$,
$\tilde G_k\to G^\star$, and the conclusion of
Lemma~\ref{lemma:gradient_approx_convergence} holds.
When $m<d$, let
$\tilde Z_k,Z^\star\in\RR^{d\times(d-m)}$
be matrices whose columns are orthonormal bases of
$\ker(\tilde G_k)$ and $\ker(G^\star)$, respectively.
Since $\tilde G_k\to G^\star$ and $G^\star$ has full row rank,
by the Davis--Kahan $\sin(\Theta)$ theorem,
$\inf_{Q\in\cQ_{d-m}}\|\tilde Z_k-Z^\star Q\|
\le2\sqrt2\,
\|\tilde Z_k\tilde Z_k^\top-Z^\star(Z^\star)^\top\|\to0$,
where $\cQ_{d-m}$ denotes the set of
$(d-m)\times(d-m)$ orthogonal matrices.
Hence, for some $Q_k\in\cQ_{d-m}$,
$\|\tilde Z_k-Z^\star Q_k\|\to0$.
Therefore,
\begin{equation*}
\nbr{
\tilde Z_k^\top\bar B_k\tilde Z_k
-Q_k^\top(Z^\star)^\top
\nabla_{\bx}^2\cL^\star Z^\star Q_k
}\to0.
\end{equation*}
Writing $\lambda_{\min}$ for the smallest eigenvalue,
orthogonal-similarity invariance and Weyl's inequality give
$\lambda_{\min}(\tilde Z_k^\top\bar B_k\tilde Z_k)
\to
\lambda_{\min}((Z^\star)^\top
\nabla_{\bx}^2\cL^\star Z^\star)$.
By Assumption~\ref{assumption:LICQ_SOSC}, we then know
$\lambda_{\min}((Z^\star)^\top
\nabla_{\bx}^2\cL^\star Z^\star)\ge\kappa_{1,B}$.
Since $\kappa_{1,\tilde B}<\kappa_{1,B}$, we have
$\lambda_{\min}(\tilde Z_k^\top\bar B_k\tilde Z_k)
\ge\kappa_{1,\tilde B}$
for all sufficiently large $k$.
When $m=d$, $\ker(\tilde G_k)=\{\zero\}$,
so the required curvature condition holds trivially.
Moreover, since $\bar B_k\to\nabla_{\bx}^2\cL^\star$ and
$\nbr{\nabla_{\bx}^2\cL^\star}
\le\kappa_{2,B}<\kappa_{2,\tilde B}$,
we also have $\nbr{\bar B_k}\le\kappa_{2,\tilde B}$
for all sufficiently large $k$.
Thus, $\bar B_k$ already satisfies the required curvature
and boundedness conditions in the regularization step,
and hence $\delta_k^B=\zero$ eventually.
That is, there exists an almost surely finite random index
$K_B^\star$ such that $\tilde B_k=\bar B_k$ for $k\ge K_B^\star$.
This completes the proof.

\subsection{Proof of Lemma \ref{lemma:local_convergence}}\label{proof_of_lemma:local_convergence}

Let $\gamma_0:=\kappa_{1,\mu}(1-\rho^s)>0$. Fix $\epsilon>0$ sufficiently small such that $\epsilon\le \gamma_0/(16\kappa_{3,\mu})$ (cf. \eqref{eq:bar_alpha_bound} for the definition of $\kappa_{3,\mu}$), and let $N_0 \ge 1$ satisfy $\alpha_{N_0} \le \min\left\{\frac{\epsilon^4\gamma_0}{32\kappa_{3,\mu}^2},\frac{1}{2\gamma_0}\right\}$ and $\beta_{N_0}<1$. We need the following two lemmas.

\begin{lemma}\label{appendix_lemma:local_convergence_1}
Under the assumptions and parameter conditions of Lemma~\ref{lemma:local_convergence}, for $N\ge N_0$ and $k\ge N$, we have
\begin{align*}
\EE[\nbr{\btheta_{k+1}-\btheta^\star}^2\1_{\tau_N(\epsilon)>k+1}] &\le \Upsilon_{5}(N)\exp\cbr{-\frac{\gamma_0\iota_1k^{1-p_1}}{1-p_1}}\\
&\quad + \Upsilon_{6}\sum_{i=N}^k\prod_{j=i+1}^k(1-\gamma_0\alpha_j)\alpha_i\EE[\|\nabla f_i - \bar{\bg}_i\|^2 \1_{\tau_N(\epsilon)>i}]\\
&\quad + \Upsilon_{6}\sum_{i=N}^k\prod_{j=i+1}^k(1-\gamma_0\alpha_j)\alpha_i\EE[\|G_i-\tilde{G}_i\|^2  \1_{\tau_N(\epsilon)>i}]
\end{align*}
for some constant $\Upsilon_{5}$ depending on $N$ and absolute constants $\Upsilon_{6}, \gamma_0 > 0$. 
\end{lemma}

\begin{lemma}\label{appendix_lemma:local_convergence_2}
Under the assumptions and parameter conditions of Lemma~\ref{lemma:local_convergence}, for $k\ge N$, we have
\begin{align*}
&\EE[\|\bar{\bg}_{k+1} - \nabla f_{k+1}\|^2\1_{\tau_N(\epsilon)>{k+1}}] + \EE[\|\bar{G}_{k+1} - G_{k+1}\|^2\1_{\tau_N(\epsilon)>{k+1}}]\\
&\le \Upsilon_{7}(N)\exp\cbr{-\frac{2\iota_2k^{1-p_2}}{1-p_2}} + \Upsilon_{8}(\beta_k + b_k^4) + \Upsilon_{8}\rbr{\sum_{i=N}^k\prod_{j=i+1}^k(1-\beta_j)\alpha_i(\EE[\|\btheta_i-\btheta^\star\|^2\1_{\tau_N(\epsilon) > i}])^{1/2}}^2\\
&\quad + \Upsilon_{8}\rbr{\sum_{i=N}^k\prod_{j=i+1}^k(1-\beta_j)\alpha_i(\EE[\|\bar{\bg}_{i} - \nabla f_{i}\|^2\1_{\tau_N(\epsilon)>{i}}] + \EE[\|\bar{G}_i - G_i\|^2\1_{\tau_N(\epsilon)>{i}}])^{1/2}}^2
\end{align*}
for some constant $\Upsilon_{7}$ depending on $N$ and some absolute constant $\Upsilon_{8}$. 
\end{lemma}

We claim that for any integer $q\ge 0$, there exists a constant $\Upsilon_q$ depending on $N$ such that 
\begin{multline*}
\EE[\nbr{\btheta_k-\btheta^\star}^2\1_{\tau_N(\epsilon)>k}] + \EE[\|\bar{\bg}_{k} - \nabla f_{k}\|^2\1_{\tau_N(\epsilon)>{k}}] + \EE[\|\bar{G}_{k} - G_{k}\|^2\1_{\tau_N(\epsilon)>{k}}]\\
\le \Upsilon_q(N)(\beta_k + b_k^4 + (\alpha_k/\beta_k)^{2q}).
\end{multline*}
First, for $k\ge N$, we have $\EE[\nbr{\btheta_k-\btheta^\star}^2\1_{\tau_N(\epsilon) > k}] \le \epsilon^4$. By the analysis in the proof of Lemma \ref{lemma:gradient_approx_convergence}, we also have $\EE[\|\bar{\bg}_{k} - \nabla f_{k}\|^2\1_{\tau_N(\epsilon)>{k}}]$ and $\EE[\|\bar{G}_{k} - G_{k}\|^2\1_{\tau_N(\epsilon)>{k}}]$ are uniformly bounded. Thus, our claim holds for $q = 0$. We now assume, for the purpose of induction, that the claim holds for a fixed integer $q$. Then, we have
\begin{align*}
& \rbr{\sum_{i=N}^k\prod_{j=i+1}^k(1-\beta_j)\alpha_i(\EE[\nbr{\btheta_i-\btheta^\star}^2\1_{\tau_N(\epsilon)>i}])^{1/2}}^2\\
&\le \cC_1(N)\rbr{\sum_{i=N}^k\prod_{j=i+1}^k(1-\beta_j)\alpha_i(\beta_i+b_i^4 + (\alpha_i/\beta_i)^{2q})^{1/2}}^2\\
&= \cC_1(N)\rbr{\sum_{i=N}^k\prod_{j=i+1}^k(1-\beta_j)\beta_i(\alpha_i/\beta_i)(\beta_i+b_i^4 + (\alpha_i/\beta_i)^{2q})^{1/2}}^2\\
&\le \cC_2(N)((\alpha_k/\beta_k)^2(\beta_k + b_k^4) + (\alpha_k/\beta_k)^{2(q+1)}),
\end{align*}
where the last inequality comes from Lemma \ref{pre_lemma:sum_of_product_2}. Similarly, we have
\begin{multline*}
\rbr{\sum_{i=N}^k\prod_{j=i+1}^k(1-\beta_j)\alpha_i(\EE[\|\bar{\bg}_{i} - \nabla f_{i}\|^2\1_{\tau_N(\epsilon)>{i}}] + \EE[\|\bar{G}_{i} - G_{i}\|^2\1_{\tau_N(\epsilon)>{i}}])^{1/2}}^2 \\
\le \cC_3(N)((\alpha_k/\beta_k)^2(\beta_k + b_k^4) + (\alpha_k/\beta_k)^{2(q+1)}).
\end{multline*}
Since exponential decay is faster than any polynomial decay, Lemma \ref{appendix_lemma:local_convergence_2} gives, for some constant $\cC_4(N)$ depending on $N$, 
\begin{equation*}
\EE[\|\bar{\bg}_{k+1} - \nabla f_{k+1}\|^2\1_{\tau_N(\epsilon)>{k+1}}] + \EE[\|\bar{G}_{k+1} - G_{k+1}\|^2\1_{\tau_N(\epsilon)>{k+1}}] \le \cC_4(N)(\beta_k + b_k^4 + (\alpha_k/\beta_k)^{2(q+1)}),
\end{equation*}
so equivalently 
\begin{equation*}
\EE[\|\bar{\bg}_{k} - \nabla f_{k}\|^2\1_{\tau_N(\epsilon)>{k}}] + \EE[\|\bar{G}_{k} - G_{k}\|^2\1_{\tau_N(\epsilon)>{k}}] \le \cC_5(N)(\beta_k + b_k^4 + (\alpha_k/\beta_k)^{2(q+1)}),
\end{equation*}
for $k\ge N$. With this result, we can apply Lemmas~\ref{pre_lemma:sum_of_product_2}(a) and~\ref{appendix_lemma:local_convergence_1} to get 
\begin{equation*}
\EE[\nbr{\btheta_k-\btheta^\star}^2\1_{\tau_N(\epsilon)> k}] \le \cC_6(N)(\beta_k + b_k^4 + (\alpha_k/\beta_k)^{2(q+1)})
\end{equation*}
for some constant $\cC_6(N)$ depending on $N$. Then our claim holds for $q+1$ with $\;\Upsilon_{q+1}(N) = \cC_5(N) + \cC_6(N)$. Thus, we complete the proof of our claim through induction. We then pick a $q$ large enough so that $2q(p_1 - p_2) \ge p_2 + 4p_3$. Using our claim with this picked $q$, we conclude that 
\begin{equation*}
\EE[\nbr{\btheta_k-\btheta^\star}^2\1_{\tau_N(\epsilon)>k}] + \EE[\|\bar{\bg}_{k} - \nabla f_{k}\|^2\1_{\tau_N(\epsilon)>{k}}] + \EE[\|\bar{G}_{k} - G_{k}\|^2\1_{\tau_N(\epsilon)>{k}}] \le \Upsilon_q(N)(\beta_k + b_k^4).
\end{equation*}

\subsection{Proof of Lemma \ref{appendix_lemma:local_convergence_1}}

Recall that, by Assumption~\ref{assumption:stepsize_selection} and \eqref{eq:bar_alpha_bound},  we have $\kappa_{1,\mu}\alpha_k\le \bar{\alpha}_k \le \kappa_{3,\mu}\alpha_k$. Since $(\btheta_{k+1}-\btheta^\star) = (\btheta_k-\btheta^\star) + \bar{\alpha}_k \bz_{k,s}$, under the event $\{\tau_N(\epsilon) > k\}$, for $k\ge N$, we have
\begin{align}
&\|\btheta_{k+1}-\btheta^\star\|^2 = \|\btheta_{k}-\btheta^\star\|^2 + \bar{\alpha}_k^2\|\bz_{k,s}\|^2 + 2\inner{\btheta_k-\btheta^\star}{\bar{\alpha}_k\bz_{k,s}}\\
&\le \|\btheta_k-\btheta^\star\|^2 + \kappa_{3,\mu}^2\alpha_k^2\|\bz_{k,s}\|^2 + 2\inner{\btheta_k-\btheta^\star}{\bar{\alpha}_k\bz_{k,s}}\\
&= \|\btheta_k-\btheta^\star\|^2 + \kappa_{3,\mu}^2\alpha_k^2\|\bz_{k,s}\|^2
-2\bar\alpha_k\langle{\btheta_k-\btheta^\star},{(I-\tilde C_k)(\btheta_k-\btheta^\star)}\rangle\\
&\quad -2\bar\alpha_k\langle{\btheta_k-\btheta^\star},{(I-\tilde C_k)\tilde W_k^{-1}(\bar\nabla\cL_k-\nabla\cL_k)}\rangle\\
&\quad -2\bar\alpha_k\langle{\btheta_k-\btheta^\star},{(I-\tilde C_k)\tilde W_k^{-1}(\nabla\cL_k-\tilde W_k(\btheta_k-\btheta^\star))}\rangle,
\label{eq:1_appendix_lemma:local_convergence_1}
\end{align}
where the last equality follows from Lemma~\ref{pre_lemma:sketching_1}(a) and \eqref{equ:KKT}. Applying Lemma \ref{pre_lemma:sketching_1}, we have
\begin{align}
\nbr{\bz_{k,s}}^2 &= \|(I-\tilde{C}_k)\tilde{\bz}_k\|^2 \le 4\nbr{\tilde{\bz}_{k}}^2 \le 4\|\tilde{W}_k^{-1}\|^2\|\bar{\nabla}\cL_k\|^2\\
&\le 8\|\tilde{W}_k^{-1}\|^2\nbr{\nabla\cL_k}^2 + 8\|\tilde{W}_k^{-1}\|^2\|\nabla_{\bx}\cL_k - \bar{\nabla}_{\bx}\cL_k\|^2. \label{eq:bz_ks_squared_bound}
\end{align}
Since $\|\tilde C_k\|\le 1$, $\langle{\btheta_k-\btheta^\star},{(I-\tilde C_k)(\btheta_k-\btheta^\star)}\rangle\ge 0$. Moreover, $\{\tau_N(\epsilon)>k\}\in\cF_{k-2/3}$ and $\bar\alpha_k\ge\kappa_{1,\mu}\alpha_k$. Hence, by Lemma~\ref{pre_lemma:sketching_1}(c),
\begin{align}
\EE[\bar\alpha_k\langle{\btheta_k-\btheta^\star},{(I-\tilde C_k)(\btheta_k-\btheta^\star)}\rangle\1_{\{\tau_N(\epsilon)>k\}}] & \ge \kappa_{1,\mu}\alpha_k\EE\sbr{\inner{\btheta_k-\btheta^\star}{(I-C_k)(\btheta_k-\btheta^\star)}\1_{\{\tau_N(\epsilon)>k\}}}\notag\\
& \ge \gamma_0\alpha_k\EE\sbr{\nbr{\btheta_k-\btheta^\star}^2\1_{\{\tau_N(\epsilon)>k\}}}.
\label{eq:2_appendix_lemma:local_convergence_1}
\end{align}
Taking expectations in \eqref{eq:1_appendix_lemma:local_convergence_1} on both sides and using \eqref{eq:bz_ks_squared_bound}, \eqref{eq:2_appendix_lemma:local_convergence_1}, and the definition of $\tau_N(\epsilon)$, we~obtain
\begin{align*}
\EE[\nbr{\btheta_{k+1}-\btheta^\star}^2\1_{\tau_N(\epsilon)>{k+1}}] &\le \EE[\nbr{\btheta_k-\btheta^\star}^2\1_{\tau_N(\epsilon)>k}]
-2\gamma_0\alpha_k\EE[\nbr{\btheta_k-\btheta^\star}^2\1_{\tau_N(\epsilon)>k}]\\
&\quad +8\epsilon^{-4}\kappa_{3,\mu}^2\alpha_k^2\EE[\nbr{\btheta_k-\btheta^\star}^2\1_{\tau_N(\epsilon)>k}] +4\epsilon\kappa_{3,\mu}\alpha_k\EE[\nbr{\btheta_k-\btheta^\star}^2\1_{\tau_N(\epsilon)>k}]\\
&\quad +8\epsilon^{-2}\kappa_{3,\mu}^2\alpha_k^2\EE[\|\bar\nabla_{\bx}\cL_k-\nabla_{\bx}\cL_k\|^2\1_{\tau_N(\epsilon)>k}]\\
&\quad 
+4\epsilon^{-1}\kappa_{3,\mu}\alpha_k\EE[\nbr{\btheta_k-\btheta^\star}\|\bar\nabla_{\bx}\cL_k-\nabla_{\bx}\cL_k\|\1_{\tau_N(\epsilon)>k}].
\end{align*}
Applying Young's inequality leads to $4\epsilon^{-1}\kappa_{3,\mu}\nbr{\btheta_k-\btheta^\star}\|\bar\nabla_{\bx}\cL_k-\nabla_{\bx}\cL_k\|
\le \frac{\gamma_0}{2}\nbr{\btheta_k-\btheta^\star}^2
+\frac{8\kappa_{3,\mu}^2}{\gamma_0\epsilon^2}\|\bar\nabla_{\bx}\cL_k-\nabla_{\bx}\cL_k\|^2$, we have
\begin{align*}
\EE[\nbr{\btheta_{k+1}-\btheta^\star}^2\1_{\tau_N(\epsilon)>{k+1}}]
&\le \left(1-\left(\frac{3\gamma_0}{2}-4\kappa_{3,\mu}\epsilon-8\epsilon^{-4}\kappa_{3,\mu}^2\alpha_k\right)\alpha_k\right)
\EE[\nbr{\btheta_k-\btheta^\star}^2\1_{\tau_N(\epsilon)>k}]\\
&\quad +\left(\frac{8\kappa_{3,\mu}^2}{\gamma_0\epsilon^2}+8\epsilon^{-2}\kappa_{3,\mu}^2\alpha_k\right)\alpha_k
\EE[\|\bar\nabla_{\bx}\cL_k-\nabla_{\bx}\cL_k\|^2\1_{\tau_N(\epsilon)>k}].
\end{align*}
Therefore, by the choices of $\epsilon$ and $N_0$, for $N\ge N_0$,
\begin{equation*}
\EE[\nbr{\btheta_{k+1}-\btheta^\star}^2\1_{\tau_N(\epsilon)>{k+1}}] \le (1-\cC_1\alpha_k)\EE[\nbr{\btheta_k-\btheta^\star}^2\1_{\tau_N(\epsilon)>k}] + \cC_2\alpha_k\EE[\|\bar{\nabla}_{\bx}\cL_k - \nabla_{\bx}\cL_k\|^2\1_{\tau_N(\epsilon)>k}],
\end{equation*}
where $\cC_1 = \gamma_0$ and $\cC_2 = 8\kappa_{3,\mu}^2/(\gamma_0\epsilon^2)+8\epsilon^{-2}\kappa_{3,\mu}^2\alpha_{N_0}$. By \eqref{eq:bar_cL_and_cL_difference}, we have $\|\bar{\nabla}_{\bx}\cL_k - \nabla_{\bx}\cL_k\|^2 \le 2\|\nabla f_k - \bar{\bg}_k\|^2 + 2\Upsilon_{\blambda}^2\|G_k-\tilde{G}_k\|^2$. Then, we can obtain that
\begin{align*}
\EE[\nbr{\btheta_{k+1}-\btheta^\star}^2\1_{\tau_N(\epsilon)>{k+1}}] &\le (1-\cC_1\alpha_k)\EE[\nbr{\btheta_k-\btheta^\star}^2\1_{\tau_N(\epsilon)>k}]\\
&\quad + 2\cC_2\alpha_k\EE[\|\nabla f_k - \bar{\bg}_k\|^2 \1_{\tau_N(\epsilon)>k}] + 2\Upsilon_{\blambda}^2\cC_2\alpha_k\EE[\|G_k-\tilde{G}_k\|^2 \1_{\tau_N(\epsilon)>k}].
\end{align*}
It follows that 
\begin{align*}
&\EE[\nbr{\btheta_{k+1}-\btheta^\star}^2\1_{\tau_{N}(\epsilon)>k+1}]\\
&\le \prod_{j=N}^k(1-\cC_1\alpha_j)\EE[\|\btheta_N-\btheta^\star\|^2\1_{\tau_N(\epsilon) > N}] + 2\cC_2\sum_{i=N}^k\prod_{j=i+1}^k(1-\cC_1\alpha_j)\alpha_i\EE[\|\nabla f_i - \bar{\bg}_i\|^2 \1_{\tau_N(\epsilon)>i}]\\
&\quad + 2\Upsilon_{\blambda}^2\cC_2\sum_{i=N}^k\prod_{j=i+1}^k(1-\cC_1\alpha_j)\alpha_i\EE[\|G_i-\tilde{G}_i\|^2  \1_{\tau_N(\epsilon)>i}].
\end{align*}
Note that
\begin{align*}
&\prod_{j=N}^k(1-\cC_1\alpha_j)\EE[\nbr{\btheta_N-\btheta^\star}^2\1_{\tau_N(\epsilon)>N}] \le \epsilon^4\prod_{j=N}^k(1-\cC_1\alpha_j)\le \epsilon^4\exp\cbr{-\cC_1\sum_{j=N}^k\alpha_j}\\
&\le \epsilon^4\exp\cbr{-\cC_1\int_{N}^{k+1}\frac{\iota_1}{(j+1)^{p_1}}~dj}\le \epsilon^4\exp\cbr{\frac{\cC_1\iota_1(N+1)^{1-p_1}}{1-p_1}}\exp\cbr{-\frac{\cC_1\iota_1k^{1-p_1}}{1-p_1}}\\
&=: \cC_3(N)\exp\cbr{-\frac{\cC_1\iota_1k^{1-p_1}}{1-p_1}}.
\end{align*}
Therefore, by choosing $\Upsilon_{5}(N) = \cC_3(N)$, $\Upsilon_{6}= {2}\cC_2\max\{1,\Upsilon^2_{\blambda}\}$, and $\gamma_0 = \cC_1$, we complete the proof.

\subsection{Proof of Lemma \ref{appendix_lemma:local_convergence_2}}

We use $\bar{\bg}_k$ as an example, while the same analysis applies to $\bar{G}_k$. Let $\nabla f_{-1}:=\bar{\bg}_{-1}$. Under the event $\cbr{\tau_N\rbr{\epsilon} > k+1}$, we have
\begin{align*}
\bar{\bg}_{k+1} - \nabla f_{k+1} &\stackrel{\eqref{eq:gradient_approx}}{=} \sum_{i=0}^{k+1}\prod_{j=i+1}^{k+1}\rbr{1-\beta_j}\beta_i(\hat{\nabla}F\rbr{\bx_i;\xi_i} - \EE[\hat{\nabla}F\rbr{\bx_i;\xi_i}\given \cF_{i-1}])\\
&\quad + \sum_{i=0}^{k+1}\prod_{j=i+1}^{k+1}\rbr{1-\beta_j}\beta_i(\EE[\hat{\nabla}F\rbr{\bx_i;\xi_i}\given \cF_{i-1}] - \nabla f_i)\\
&\quad  + \sum_{i=N+1}^{k+1}\prod_{j=i}^{k+1}\rbr{1-\beta_j}\rbr{\nabla f_{i-1} - \nabla f_i} + \sum_{i=0}^{N}\prod_{j=i}^{k+1}\rbr{1-\beta_j}\rbr{\nabla f_{i-1} - \nabla f_i}\\
&\eqcolon \cT^1_{k+1} + \cT^2_{k+1} + \cT^3_{k+1} + \cT^4_{k+1}.
\end{align*}
Then, we have
\begin{equation}\label{eq:appendix_lemma:local_convergence_2:bar_g_bound}
\EE[\|\bar{\bg}_{k+1} - \nabla f_{k+1}\|^2\1_{\tau_N(\epsilon)>{k+1}}]\le 4\rbr{\sum_{j=1}^4\EE[\|\cT^j_{k+1}\|^2 \1_{\tau_N\rbr{\epsilon} > k+1}]}.
\end{equation}
For $0\le i\le k+1$, the event $\{\tau_N(\epsilon)>i-1\}$ is $\cF_{i-1}$-measurable. Thus,
\begin{align}
\EE[\|\cT^1_{k+1}\|^2 \1_{\tau_N(\epsilon)>k+1}] &\le
\EE\left\| \sum_{i=0}^{k+1}\prod_{j=i+1}^{k+1}(1-\beta_j)\beta_i (\hat{\nabla}F(\bx_i;\xi_i) -\EE[\hat{\nabla}F(\bx_i;\xi_i)\given\cF_{i-1}]) \1_{\tau_N(\epsilon)>i-1} \right\|^2\\
&= \sum_{i=0}^{k+1}\prod_{j=i+1}^{k+1}(1-\beta_j)^2\beta_i^2 \EE[ \|\hat{\nabla}F(\bx_i;\xi_i) -\EE[\hat{\nabla}F(\bx_i;\xi_i)\given\cF_{i-1}]\|^2 \1_{\tau_N(\epsilon)>i-1}]\\
&\le \cC_1\beta_k, \label{eq:appendix_lemma:local_convergence_2:bar_g_bound_first_term}
\end{align}
where the equality follows from martingale orthogonality and the last inequality follows from \eqref{eq:hat_nabla_F_second_moment_bound} and Lemma~\ref{pre_lemma:sum_of_product_2}. The bias bound in Lemma~\ref{lemma:approximation_rate} and Lemma~\ref{pre_lemma:sum_of_product_2} give
\begin{equation}\label{eq:appendix_lemma:local_convergence_2:bar_g_bound_second_term}
\EE[\|\cT^2_{k+1}\|^2 \1_{\tau_N\rbr{\epsilon} > k+1}] \le \cC_1 b_k^4. \end{equation}
For $\cT_{k+1}^3$, we also have
\begin{align}
\EE[\|\cT_{k+1}^3\|^2\1_{\tau_N(\epsilon)>k+1}] &\le \EE\sbr{\rbr{\sum_{i=N+1}^{k+1}\prod_{j=i}^{k+1}\rbr{1-\beta_j}\|\nabla f_{i-1} - \nabla f_i\|}^2\1_{\tau_N(\epsilon)>k+1}}\\
&\le \EE\sbr{\rbr{\sum_{i=N+1}^{k+1}\prod_{j=i}^{k+1}\rbr{1-\beta_j}\|\nabla f_{i-1} - \nabla f_i\|\1_{\tau_N(\epsilon)>i-1}}^2}\\
&\le \rbr{\sum_{i=N}^{k}\prod_{j=i+1}^{k}\rbr{1-\beta_j}(\EE[\|\nabla f_{i+1} - \nabla f_i\|^2\1_{\tau_N(\epsilon)>i}])^{1/2}}^2, \label{eq:appendix_lemma:local_convergence_2:term_3_bound_1}
\end{align}
where the last inequality follows from the Minkowski's inequality. Note that, by Lipschitz property of $\nabla f_i$, we have
\begin{equation}
\|\nabla f_{i+1} - \nabla f_i\| \le \Upsilon_u\bar{\alpha}_{i}\|\bz_{i,s}\|\le 2\Upsilon_u\kappa_{3,\mu}\alpha_{i}\|\tilde{\bz}_{i}\|. \label{eq:appendix_lemma:local_convergence_2:term_3_bound_2}
\end{equation}
Under the event $\cbr{\tau_{N}\rbr{\epsilon} > i}$, we have
\begin{align}
\|\tilde{\bz}_{i}\|^2 \le \|\tilde{W}_{i}^{-1}\|^2\|\bar{\nabla}\cL_{i}\|^2 &\le 2\epsilon^{-2}(\|\nabla \cL_{i}\|^2 + \|\bar{\nabla}_{\bx}\cL_{i} - \nabla_{\bx}\cL_{i}\|^2)\\
&\le 2\epsilon^{-4}\|\btheta_{i} - \btheta^\star\|^2 + 4\epsilon^{-2}\|\bar{\bg}_{i} - \nabla f_{i}\|^2 + 4\Upsilon_{\blambda}^2\epsilon^{-2}\|\bar{G}_{i} - G_{i}\|^2 \\
&\le \cC_2(\|\btheta_{i} - \btheta^\star\|^2 + \|\bar{\bg}_{i} - \nabla f_{i}\|^2 + \|\bar{G}_{i} - G_{i}\|^2),\label{eq:appendix_lemma:local_convergence_2:term_3_bound_3}
\end{align}
where we use \eqref{eq:debiased_Lagrangian_Hessian}, \eqref{equ:KKT}, and the stopping-time bounds, and $\cC_2:= \max\{2\epsilon^{-4}, 4\epsilon^{-2}, 4\Upsilon_{\blambda}^2\epsilon^{-2}\}$. Combining \eqref{eq:appendix_lemma:local_convergence_2:term_3_bound_1}, \eqref{eq:appendix_lemma:local_convergence_2:term_3_bound_2} and \eqref{eq:appendix_lemma:local_convergence_2:term_3_bound_3}, we have that 
\begin{align}
&\EE[\|\cT_{k+1}^3\|^2\1_{\tau_N(\epsilon)>k+1}] \\
&\le 4\Upsilon_u^2\kappa^2_{3,\mu}\rbr{\sum_{i=N}^{k}\prod_{j=i+1}^{k}\rbr{1-\beta_j}\alpha_{i}(\EE[\|\tilde{\bz}_{i}\|^2\1_{\tau_N(\epsilon)>i}])^{1/2}}^2\\
&\le 8\cC_2\Upsilon_u^2\kappa^2_{3,\mu}\rbr{\sum_{i=N}^{k}\prod_{j=i+1}^{k}\rbr{1-\beta_j}\alpha_{i}(\EE[\|\btheta_{i} - \btheta^\star\|^2\1_{\tau_N(\epsilon)>i}])^{1/2}}^2\\
&\quad + 8\cC_2\Upsilon_u^2\kappa^2_{3,\mu}\rbr{\sum_{i=N}^{k}\prod_{j=i+1}^{k}\rbr{1-\beta_j}\alpha_{i}(\EE[\|\bar{\bg}_{i} - \nabla f_{i}\|^2\1_{\tau_N(\epsilon)>i}] + \EE[\|\bar{G}_{i} - G_{i}\|^2\1_{\tau_N(\epsilon)>i}])^{1/2}}^2. \label{eq:appendix_lemma:local_convergence_2:bar_g_bound_third_term}
\end{align}
Finally, for $\cT_{k+1}^4$, we have 
\begin{equation*}
\EE[\|\cT^4_{k+1}\|^2 \1_{\tau_N\rbr{\epsilon} > k+1}] \le \rbr{\sum_{i=0}^{N}\prod_{j=i}^{k+1}|1-\beta_j|(\EE[\|\nabla f_{i-1} - \nabla f_i\|^2])^{1/2}}^2.
\end{equation*}
Since $\nabla f_{-1}=\bar{\bg}_{-1}$ is deterministic and $\|\nabla f_i\|\le\Upsilon_u$, $\EE[\|\nabla f_{i-1}-\nabla f_i\|^2]\le\cC_3<\infty$ for $0\le i\le N$. We further obtain
\begin{align}
\EE[\|\cT^4_{k+1}\|^2 \1_{\tau_N\rbr{\epsilon} > k+1}] & \le \cC_3\rbr{\sum_{i=0}^{N}\prod_{j=i}^{k+1}|1-\beta_j|}^2 = \cC_3\rbr{\sum_{i=0}^{N}\prod_{j=i}^{N-1}|1-\beta_j|}^2\prod_{j=N}^{k+1}(1-\beta_j)^2 \notag\\
&\eqcolon \cC_4(N)\prod_{j=N}^{k+1}(1-\beta_j)^2 \le \cC_4(N)\exp\cbr{-2\sum_{j=N}^{k+1}\beta_j} \nonumber\\
& \le \cC_4(N)\exp\cbr{-\int_{N}^{k+2}\frac{2\iota_2}{\rbr{j+1}^{p_2}}~dj}\nonumber\\
&\le \cC_4(N)\exp\cbr{\frac{2\iota_2\rbr{N+1}^{1-p_2}}{1-p_2}}\exp\cbr{-\frac{2\iota_2k^{1-p_2}}{1-p_2}}. \label{eq:appendix_lemma:local_convergence_2:bar_g_bound_fourth_term}
\end{align}
Therefore, putting \eqref{eq:appendix_lemma:local_convergence_2:bar_g_bound_first_term}, \eqref{eq:appendix_lemma:local_convergence_2:bar_g_bound_second_term}, \eqref{eq:appendix_lemma:local_convergence_2:bar_g_bound_third_term}, \eqref{eq:appendix_lemma:local_convergence_2:bar_g_bound_fourth_term} into \eqref{eq:appendix_lemma:local_convergence_2:bar_g_bound}, we complete the proof for the $\EE[\|\bar{\bg}_{k+1} - \nabla f_{k+1}\|^2\1_{\tau_N(\epsilon)>{k+1}}]$ part. For $\EE[\|\bar{G}_{k+1} - G_{k+1}\|^2\1_{\tau_N(\epsilon)>{k+1}}]$, the proof can be done similarly.

\subsection{Proof of Lemma \ref{lemma:W_convergence_rate}}\label{proof_of_lemma:W_convergence_rate}

Fix $\epsilon$ satisfying the condition in
Lemma~\ref{lemma:local_convergence}
and $\epsilon^2\le\delta_{\rm loc}/2$.
Choose a deterministic $N_0^\prime\ge\max\{N_0,K^\prime\}$,
where $N_0$ is from the proof of
Lemma~\ref{lemma:local_convergence} and $K^\prime$
is from \eqref{eq:localized_Lagrangian_Hessian_moments},
sufficiently large that
\begin{equation}
\sqrt d\kappa_{\bDelta_2}
(b_{N_0^\prime}+\tilde b_{N_0^\prime})
\le\delta_{\rm loc}/2,
\qquad b_{N_0^\prime}\le1.
\label{eq:proof_of_lemma:W_convergence_rate:N_0_prime_condition}
\end{equation}
For $N\ge N_0^\prime$ and $N\le i\le k$,
on $\{\tau_N(\epsilon)>k\}$ we have
$\|\bx_i-\bx^\star\|
\le\|\btheta_i-\btheta^\star\|
\le\epsilon^2\le\delta_{\rm loc}/2$.
Thus, for $s_1\in[-b_i,b_i]$ and $s_2\in[0,\tilde b_i]$,
\begin{equation*}
\|\bx_i+s_1\bDelta_i+s_2\tilde{\bDelta}_i-\bx^\star\|
\le\|\bx_i-\bx^\star\|
+\sqrt d\kappa_{\bDelta_2}(b_i+\tilde b_i)
\le\delta_{\rm loc}.
\end{equation*}
Moreover, $\tilde B_k=\bar B_k$ and $\tilde G_k=\bar G_k$
on this event, so
\begin{align}
\|\tilde W_k-W^\star\|^2\1_{\tau_N(\epsilon)>k}
&=\nbr{
\begin{pmatrix}
\bar B_k-\nabla_{\bx}^2\cL^\star
&(\bar G_k-G^\star)^\top\\
\bar G_k-G^\star&\zero
\end{pmatrix}
}^2\1_{\tau_N(\epsilon)>k}\notag\\
&\le
2\|\bar B_k-\nabla_{\bx}^2\cL^\star\|^2
\1_{\tau_N(\epsilon)>k}
+2\|\bar G_k-G^\star\|^2\1_{\tau_N(\epsilon)>k}.
\label{eq:W_error_block_bound}
\end{align}

Expanding the recursion in
\eqref{eq:debiased_Lagrangian_Hessian} gives
\begin{align}
\bar B_k-\nabla_{\bx}^2\cL^\star
&=\prod_{\ell=N+1}^k(1-\beta_\ell)
(\bar B_N-\nabla_{\bx}^2\cL^\star) +
\sum_{i=N+1}^k\prod_{\ell=i+1}^k(1-\beta_\ell)\beta_i
\rbr{\hat{\nabla}_{\bx}^2\cL_i
-\EE[\hat{\nabla}_{\bx}^2\cL_i\given\cF_{i-1}]}\notag\\
&\quad+
\sum_{i=N+1}^k\prod_{\ell=i+1}^k(1-\beta_\ell)\beta_i
\rbr{\EE[\hat{\nabla}_{\bx}^2\cL_i\given\cF_{i-1}]
-\nabla_{\bx}^2\cL^\star}.
\label{eq:proof_of_lemma:W_convergence_rate:K_decomposition}
\end{align}
We bound these three terms separately.
Since $\bar B_N=\tilde B_N$ on $\{\tau_N(\epsilon)>k\}$,
Assumptions~\ref{assumption:boundedness_1}
and \ref{assumption:LICQ_SOSC} give
\begin{equation*}
\|\bar B_N-\nabla_{\bx}^2\cL^\star\|^2
\1_{\tau_N(\epsilon)>k}
\le(\kappa_{2,\tilde B}+\kappa_{2,B})^2.
\end{equation*}
Therefore, using $0<\beta_i\le1$ and $p_2<1$,
\begin{align}
&\EE\sbr{
\nbr{
\prod_{\ell=N+1}^k(1-\beta_\ell)
(\bar B_N-\nabla_{\bx}^2\cL^\star)
}^2
\1_{\tau_N(\epsilon)>k}} \le
\cC_1\prod_{\ell=N+1}^k(1-\beta_\ell)^2
\le\cC_1\exp\rbr{-2\sum_{\ell=N+1}^k\beta_\ell}\notag\\
&\quad\le
\cC_1\exp\rbr{
-\frac{2\iota_2}{1-p_2}
\rbr{(k+2)^{1-p_2}-(N+2)^{1-p_2}}}
\le\cC_2(N)\beta_k.
\label{eq:proof_of_lemma:W_convergence_rate:8th_bound}
\end{align}

The derivations of \eqref{eq:Lagrangian_Hessian_mean_error}
and \eqref{eq:localized_Lagrangian_Hessian_moments}
apply under the current assumptions and do not require $p_2>1/2$.
For $N<i\le k$,
$\1_{\|\bx_i-\bx^\star\|\le\delta_{\rm loc}/2}=1$
on $\{\tau_N(\epsilon)>k\}$.
The conditional mean and second-moment bounds in
\eqref{eq:localized_Lagrangian_Hessian_moments},
together with the deterministic weights $\beta_i$,
allow Frobenius martingale orthogonality to give
\begin{align}
&\EE\sbr{
\nbr{
\sum_{i=N+1}^k
\prod_{\ell=i+1}^k(1-\beta_\ell)\beta_i
\rbr{\hat{\nabla}_{\bx}^2\cL_i
-\EE[\hat{\nabla}_{\bx}^2\cL_i\given\cF_{i-1}]}
}^2
\1_{\tau_N(\epsilon)>k}}\notag\\
&\quad\le
\EE\nbr{
\sum_{i=N+1}^k
\prod_{\ell=i+1}^k(1-\beta_\ell)\beta_i
\1_{\|\bx_i-\bx^\star\|\le\delta_{\rm loc}/2}
\rbr{\hat{\nabla}_{\bx}^2\cL_i
-\EE[\hat{\nabla}_{\bx}^2\cL_i\given\cF_{i-1}]}
}_F^2\notag\\
&\quad=
\sum_{i=N+1}^k
\prod_{\ell=i+1}^k(1-\beta_\ell)^2\beta_i^2
\EE\sbr{
\nbr{\hat{\nabla}_{\bx}^2\cL_i
-\EE[\hat{\nabla}_{\bx}^2\cL_i\given\cF_{i-1}]}
_F^2
\1_{\|\bx_i-\bx^\star\|\le\delta_{\rm loc}/2}}\notag\\
&\quad\le
\cC_3\sum_{i=N+1}^k
\prod_{\ell=i+1}^k(1-\beta_\ell)^2\beta_i^2
\le\cC_4\beta_k,
\label{eq:proof_of_lemma:W_convergence_rate:1st_bound}
\end{align}
where the last inequality follows from
Lemma~\ref{pre_lemma:sum_of_product_2}(a).

By \eqref{eq:Lagrangian_Hessian_mean_error},
Minkowski's inequality, and
$\{\tau_N(\epsilon)>k\}\subseteq\{\tau_N(\epsilon)>i\}$
for $i\le k$, we have
\begin{align*}
&\EE\sbr{
\nbr{
\sum_{i=N+1}^k
\prod_{\ell=i+1}^k(1-\beta_\ell)\beta_i
\rbr{\EE[\hat{\nabla}_{\bx}^2\cL_i\given\cF_{i-1}]
-\nabla_{\bx}^2\cL^\star}
}^2
\1_{\tau_N(\epsilon)>k}}^{1/2}\\
&\quad\le
\cC_5\sum_{i=N+1}^k
\prod_{\ell=i+1}^k(1-\beta_\ell)\beta_i
\rbr{
b_i+\tilde b_i^2/b_i
+\EE[\|\btheta_i-\btheta^\star\|^2
\1_{\tau_N(\epsilon)>i}]^{1/2}}\\
&\quad\le
\cC_6(N)\sum_{i=N+1}^k
\prod_{\ell=i+1}^k(1-\beta_\ell)\beta_i
\rbr{b_i+\tilde b_i^2/b_i+\sqrt{\beta_i}+b_i^2} \le
\cC_7(N)\rbr{b_k+\tilde b_k^2/b_k+\sqrt{\beta_k}}.
\end{align*}
The second inequality uses Lemma~\ref{lemma:local_convergence};
the last uses Lemma~\ref{pre_lemma:sum_of_product_2}(a),
$2p_4-p_3>0$, and $b_k\le1$.
The squared triangle inequality in
\eqref{eq:proof_of_lemma:W_convergence_rate:K_decomposition}
now yields
\begin{equation}
\EE[\|\bar B_k-\nabla_{\bx}^2\cL^\star\|^2
\1_{\tau_N(\epsilon)>k}]
\le\cC_8(N)\rbr{\beta_k+b_k^2+\tilde b_k^4/b_k^2}.
\label{eq:W_rate_Hessian_bound}
\end{equation}
Finally,
$\|G_k-G^\star\|\le\cC_9\|\btheta_k-\btheta^\star\|$
by Assumption~\ref{assumption:boundedness_1}.
Thus, Lemma~\ref{lemma:local_convergence} gives
\begin{align}
\EE[\|\bar G_k-G^\star\|^2\1_{\tau_N(\epsilon)>k}]
&\le
2\EE[\|\bar G_k-G_k\|^2\1_{\tau_N(\epsilon)>k}] +
2\cC_9^2\EE[\|\btheta_k-\btheta^\star\|^2
\1_{\tau_N(\epsilon)>k}]\notag\\
&\le\cC_{10}(N)(\beta_k+b_k^4).
\label{eq:W_rate_Jacobian_bound}
\end{align}
Combining \eqref{eq:W_error_block_bound},
\eqref{eq:W_rate_Hessian_bound}, and
\eqref{eq:W_rate_Jacobian_bound}
with $b_k^4\le b_k^2$ proves the lemma.

\subsection{Proof of Theorem \ref{theorem:local_convergence}}
\label{proof_of_thm:local_convergence}

By the two safeguard hypotheses and
Lemmas~\ref{lemma:gradient_approx_convergence} and
\ref{lemma:Hessian_approx_convergence},
$\tilde W_k\to W^\star$ almost surely; hence $C_k\to C^\star$
by Lemma~\ref{pre_lemma:sketching_2}.
The same lemma gives
$C^\star=(I-\EE[W^\star S(S^\top(W^\star)^2S)^\dagger
S^\top W^\star])^s$.
The matrix inside the expectation is an orthogonal projector,
so $C^\star$ is symmetric positive semidefinite.
Passing to the limit in $\|C_k\|\le\rho^s$, where
$\rho=1-\gamma_S<1$, from Lemma~\ref{pre_lemma:sketching_1}(c)
gives $\|C^\star\|\le\rho^s$.
Thus $I-C^\star\succeq(1-\rho^s)I\succ0$.

Define
\begin{equation*}
\begin{cases}
\cT_k:= -(W^\star)^{-1}
(\bar{\nabla}\cL_k-\nabla\cL_k)
\\
\cH_1^k:= (I-C^\star)^{-1}
(\bz_{k,s}-(I-C_k)\tilde{\bz}_k)
\\
\cH_2^k:= -(\tilde W_k^{-1}-(W^\star)^{-1})
(\bar{\nabla}\cL_k-\nabla\cL_k)
\\
\cH_3^k:= -(\tilde W_k^{-1}-(W^\star)^{-1})
\nabla\cL_k
\\
\cH_4^k:= -(W^\star)^{-1}
(\nabla\cL_k-W^\star(\btheta_k-\btheta^\star))
\\
\cH_5^k:= (I-C^\star)^{-1}(C^\star-C_k)
(\cT_k+\cH_2^k+\cH_3^k+\cH_4^k
-(\btheta_k-\btheta^\star))
\\
\cH_6^k:=
-\frac{\bar\alpha_k-\mu_k\alpha_k}{\bar\alpha_k}
(\btheta_k-\btheta^\star).
\end{cases}
\end{equation*}

By \eqref{equ:KKT} and these definitions,
\begin{align*}
\tilde{\bz}_k
&=-(\btheta_k-\btheta^\star)
+\cT_k+\cH_2^k+\cH_3^k+\cH_4^k,
\\
\bz_{k,s}
&=(I-C_k)\tilde{\bz}_k+(I-C^\star)\cH_1^k
=(I-C^\star)(\tilde{\bz}_k+\cH_1^k+\cH_5^k),
\end{align*}
where the last equality uses
$\cH_5^k=(I-C^\star)^{-1}(C^\star-C_k)\tilde{\bz}_k$.
Substituting into \eqref{eq:update} and using the definition
of $\cH_6^k$ gives
\begin{equation*}
\btheta_{k+1}-\btheta^\star
=(I-\mu_k\alpha_k(I-C^\star))(\btheta_k-\btheta^\star)
+\bar\alpha_k(I-C^\star)
\rbr{\cT_k+\sum_{\ell=1}^{6}\cH_\ell^k}.
\end{equation*}
Iterating this recursion and summing over $k$ gives
\begin{align}
&\frac{1}{\sqrt t}\sum_{k=1}^{t}(\btheta_k-\btheta^\star)
=
\frac{1}{\sqrt t}\sum_{k=0}^{t-1}
\prod_{i=0}^k(I-\mu_i\alpha_i(I-C^\star))
(\btheta_0-\btheta^\star)
\notag\\
&\quad+
\frac{1}{\sqrt t}\sum_{k=0}^{t-1}\sum_{i=0}^k
\prod_{j=i+1}^k(I-\mu_j\alpha_j(I-C^\star))
\bar\alpha_i(I-C^\star)\cT_i
\notag\\
&\quad+
\frac{1}{\sqrt t}\sum_{k=0}^{t-1}\sum_{i=0}^k
\prod_{j=i+1}^k(I-\mu_j\alpha_j(I-C^\star))
\bar\alpha_i(I-C^\star)
(\cH_1^i+\cH_2^i+\cH_3^i+\cH_4^i+\cH_5^i+\cH_6^i).
\label{eq:averaged_iterate_decomposition}
\end{align}

We use the following convolution lemma to control the remainder terms.

\begin{lemma}
\label{appendix_lemma_theorem:local_convergence_1}
Let $\{X_k\}_{k\ge0}$ be a sequence of random vectors.
Let $\tau_N$ be a stopping time indexed by $N$ with respect
to the filtration $\{\cF_k\}_{k\ge0}$ such that
$\PP(\tau_N<\infty)\to0$ as $N\to\infty$.
Suppose that, for every fixed $N$,
$\EE\sbr{\nbr{X_k}\1_{\{\tau_N>k\}}}
=O((k+1)^{-q})$ for some $q>1/2$.
Let $M$ be symmetric positive definite, let
$\alpha_k=\iota(k+1)^{-a}$ with $\iota>0$ and $a\in(0,1)$,
and let $\{\mu_k\}$ satisfy
Assumption~\ref{assumption:stepsize_selection}.
Then
\begin{equation*}
\max_{1\le n\le t}
\nbr{
\frac{1}{\sqrt t}\sum_{k=0}^{n-1}\sum_{i=0}^k
\prod_{j=i+1}^k(I-\mu_j\alpha_jM)\alpha_iMX_i
}
\to0
\qquad\text{in probability}.
\end{equation*}
In particular, taking $n=t$ gives
\begin{equation*}
\frac{1}{\sqrt t}\sum_{k=0}^{t-1}\sum_{i=0}^k
\prod_{j=i+1}^k(I-\mu_j\alpha_jM)\alpha_iMX_i
\to\zero
\qquad\text{in probability}.
\end{equation*}
\end{lemma}

We first verify that the stopping time defined in
Section~\ref{sec:local_convergence} satisfies the condition in
Lemma~\ref{appendix_lemma_theorem:local_convergence_1}.
Choose $\epsilon>0$ sufficiently small such that
$\epsilon^2\le\delta_{\rm loc}/2$, the condition on $\epsilon$
in Lemma~\ref{lemma:local_convergence} holds, and
$\epsilon\max\{\|(W^\star)^{-1}\|,\|W^\star\|,
\|\btheta^\star\|,\cC_0\}<1$.
Here, $\cC_0>0$ is a constant such that
$\|\nabla\cL(\bx,\blambda)-W^\star(\btheta-\btheta^\star)\|
\le\cC_0\|\btheta-\btheta^\star\|^2$
for all $\btheta=(\bx,\blambda)$ in a neighborhood of
$\btheta^\star$, which exists by the local Lipschitz continuity
of $\nabla^2\cL$ around $\btheta^\star$.

By Assumption~\ref{assumption:LICQ_SOSC},
$\btheta_k\to\btheta^\star$ almost surely.
Moreover, by the two safeguard hypotheses of
Theorem~\ref{theorem:local_convergence} and
Lemmas~\ref{lemma:gradient_approx_convergence} and
\ref{lemma:Hessian_approx_convergence},
$\tilde W_k\to W^\star$ almost surely, and
$\delta_k^G=\zero$, $\delta_k^B=\zero$ eventually almost surely.
Since $W^\star$ is nonsingular, we also have
$\tilde W_k^{-1}\to(W^\star)^{-1}$ almost surely.
Hence, almost surely, for all sufficiently large $k$,
$\|\btheta_k-\btheta^\star\|\le\epsilon^2$,
$\|\tilde W_k^{-1}\|\le1/\epsilon$, and
$\|\btheta_k\|\le1/\epsilon$.
Also, since $\epsilon\cC_0<1$,
$\|\nabla\cL_k-W^\star(\btheta_k-\btheta^\star)\|
\le\cC_0\|\btheta_k-\btheta^\star\|^2
\le\|\btheta_k-\btheta^\star\|^2/\epsilon$
for all sufficiently large $k$.
Finally,
\begin{align*}
\|\nabla\cL_k-\tilde W_k(\btheta_k-\btheta^\star)\|
&\le
\rbr{\cC_0\|\btheta_k-\btheta^\star\|
+\|W^\star-\tilde W_k\|}
\|\btheta_k-\btheta^\star\|
\le\epsilon^2\|\btheta_k-\btheta^\star\|,
\\
\|\nabla\cL_k\|
&\le
\rbr{\|W^\star\|+\cC_0\|\btheta_k-\btheta^\star\|}
\|\btheta_k-\btheta^\star\|
\le\|\btheta_k-\btheta^\star\|/\epsilon,
\end{align*}
where the last inequalities hold eventually almost surely
because $\tilde W_k\to W^\star$,
$\btheta_k\to\btheta^\star$, and $\epsilon\|W^\star\|<1$.
Therefore, almost surely, $\tau_N(\epsilon)=\infty$
for all sufficiently large $N$, so
$\1_{\{\tau_N(\epsilon)<\infty\}}\to0$ almost surely.
By dominated convergence,
$\PP(\tau_N(\epsilon)<\infty)\to0$ as $N\to\infty$.

For each fixed $N\ge\max\{N_0,N_0^\prime\}$,
where $N_0$ and $N_0^\prime$ are given by
Lemmas~\ref{lemma:local_convergence} and
\ref{lemma:W_convergence_rate} for this $\epsilon$,
Lemmas~\ref{lemma:W_convergence_rate},
\ref{lemma:local_convergence}, and
\ref{pre_lemma:sketching_2} give
\begin{align*}
\EE[
\nbr{\btheta_k-\btheta^\star}^2
\1_{\tau_N>k}
]^{1/2}
+
\EE[
\nbr{\bar\nabla_{\bx}\cL_k-\nabla_{\bx}\cL_k}^2
\1_{\tau_N>k}
]^{1/2}
&=
O(\beta_k^{1/2}+b_k^2),
\\
\EE[
\|\tilde W_k-W^\star\|^2\1_{\tau_N>k}
]^{1/2}
+
\EE[
\nbr{C_k-C^\star}^2\1_{\tau_N>k}
]^{1/2}
&=
O(\beta_k^{1/2}+b_k+\tilde b_k^2/b_k).
\end{align*}
Since the lower blocks of $\bar\nabla\cL_k$ and $\nabla\cL_k$
are both $c_k$, we have
$\|\bar\nabla\cL_k-\nabla\cL_k\|
=\|\bar\nabla_{\bx}\cL_k-\nabla_{\bx}\cL_k\|$.
These bounds also hold for $N<\max\{N_0,N_0^\prime\}$,
since $\tau_N(\epsilon)$ is nondecreasing in $N$.
On $\{\tau_N(\epsilon)>k\}$, the identity
$\tilde W_k^{-1}-(W^\star)^{-1}
=\tilde W_k^{-1}(W^\star-\tilde W_k)(W^\star)^{-1}$
and the stopping bounds control $\cH_2^k$ and $\cH_3^k$.
For $\cH_5^k$, use its expression in terms of
$\tilde{\bz}_k$ above and
\begin{equation*}
\|\tilde{\bz}_k\|
\le
\epsilon^{-1}\|\bar\nabla\cL_k-\nabla\cL_k\|
+\epsilon^{-2}\|\btheta_k-\btheta^\star\|.
\end{equation*}
Using the definitions of $\cH_2^k,\ldots,\cH_6^k$,
the stopping-time bounds, the stepsize condition in
\eqref{eq:stepsize_selection}, and Cauchy's inequality, we obtain
\begin{align*}
\sum_{\ell\in\{2,3,5\}}
\EE[\|\cH_\ell^k\|\1_{\tau_N>k}]
&=
O\rbr{
(\beta_k^{1/2}+b_k+\tilde b_k^2/b_k)
(\beta_k^{1/2}+b_k^2)
},
\\
\EE[\|\cH_4^k\|\1_{\tau_N>k}]
&=
O(\beta_k+b_k^4),
\\
\EE[\|\cH_6^k\|\1_{\tau_N>k}]
&=
O\rbr{\alpha_k^{p-1}(\beta_k^{1/2}+b_k^2)}.
\end{align*}
Under \eqref{equ:cond4.6}, we have $p_2>1/2$, $p_3>1/4$,
and $2p_4-p_3>(1-p_2)/2>0$, which imply
\begin{equation*}
\min\cbr{
p_2,\frac{p_2}{2}+p_3,3p_3,
\frac{p_2}{2}+2p_4-p_3,2p_4+p_3
}
>\frac12.
\end{equation*}
Also, $2p_3>1/2>p_2/2$ and
$p>1+(1-p_2)/(2p_1)$ yield
$p_1(p-1)+\min\{p_2/2,2p_3\}>1/2$.
Thus, for some $q>1/2$,
$\EE[\|\cH_\ell^k\|\1_{\tau_N>k}]=O((k+1)^{-q})$
for $\ell=2,\ldots,6$.
The bound in \eqref{eq:stepsize_selection}, together with
Assumption~\ref{assumption:stepsize_selection}(a),
implies that $\bar\alpha_i/\alpha_i$ is uniformly bounded.
It then follows from
Lemma~\ref{appendix_lemma_theorem:local_convergence_1}
that the terms involving $\cH_2^i,\ldots,\cH_6^i$
in \eqref{eq:averaged_iterate_decomposition}
converge to $\zero$ in probability.

For the initial-condition term, choose a deterministic $K$
such that $\kappa_{2,\mu}\alpha_K\|I-C^\star\|\le1$.
By Assumption~\ref{assumption:stepsize_selection}(a)
and $I-C^\star\succeq(1-\rho^s)I$, for $i\ge K$,
\begin{equation*}
\|I-\mu_i\alpha_i(I-C^\star)\|
\le
1-\kappa_{1,\mu}(1-\rho^s)\alpha_i
\le
\exp\{-\kappa_{1,\mu}(1-\rho^s)\alpha_i\}.
\end{equation*}
The finitely many earlier factors have deterministic bounds,
and $\sum_{i=K}^k\alpha_i$ grows at order $k^{1-p_1}$
with $p_1<1$.
Consequently, for a deterministic constant $\cC_1<\infty$,
$\sum_{k=0}^{\infty}
\|\prod_{i=0}^k(I-\mu_i\alpha_i(I-C^\star))\|
\le\cC_1$ almost surely.
Hence
\begin{equation*}
\max_{1\le n\le t}
\nbr{
\frac1{\sqrt t}\sum_{k=0}^{n-1}
\prod_{i=0}^k(I-\mu_i\alpha_i(I-C^\star))
(\btheta_0-\btheta^\star)
}
\le
\frac{\cC_1}{\sqrt t}\|\btheta_0-\btheta^\star\|
\to0
\qquad\text{almost surely}.
\end{equation*}

It remains to prove the following two lemmas.

\begin{lemma}
\label{appendix_lemma_theorem:local_convergence_2}
Under the assumptions and parameter conditions of
Theorem~\ref{theorem:local_convergence}, we have
\begin{equation*}
\frac{1}{\sqrt t}\sum_{k=0}^{t-1}\sum_{i=0}^k
\prod_{j=i+1}^k(I-\mu_j\alpha_j(I-C^\star))
\bar\alpha_i(I-C^\star)\cT_i
\overset{d}{\longrightarrow}
\cN(\zero,\bSigma^\star_{\rm ave}).
\end{equation*}
\end{lemma}

\begin{lemma}
\label{appendix_lemma_theorem:local_convergence_3}
Under the assumptions and parameter conditions of
Theorem~\ref{theorem:local_convergence}, we have
\begin{equation*}
\frac{1}{\sqrt t}\sum_{k=0}^{t-1}\sum_{i=0}^k
\prod_{j=i+1}^k(I-\mu_j\alpha_j(I-C^\star))
\bar\alpha_i(I-C^\star)\cH_1^i
\to\zero
\qquad\text{in probability}.
\end{equation*}
\end{lemma}

Combining the preceding estimates with
Lemmas~\ref{appendix_lemma_theorem:local_convergence_2} and
\ref{appendix_lemma_theorem:local_convergence_3}
yields the asserted Gaussian limit with summation indices
$k=1,\ldots,t$.
Finally, $\btheta_k\to\btheta^\star$ almost surely implies
$\frac{(\btheta_t-\btheta^\star)-(\btheta_0-\btheta^\star)}
{\sqrt t}\to\zero$ almost surely.
Thus,
\begin{equation*}
\sqrt t\cdot(\bar\btheta_t-\btheta^\star)
=
\frac{1}{\sqrt t}\sum_{k=0}^{t-1}
(\btheta_k-\btheta^\star)
\overset{d}{\longrightarrow}
\cN(\zero,\bSigma^\star_{\rm ave}),
\end{equation*}
which proves Theorem~\ref{theorem:local_convergence}.

\subsection{Proof of Lemma \ref{appendix_lemma_theorem:local_convergence_1}}
Let $K^\star$ be the almost surely finite stabilization index from
Assumption~\ref{assumption:stepsize_selection}. Fix $K$ and
$\ell\in\{1,\ldots,L\}$. On the event $E_{K,\ell}:=\{K^\star\leq K,\ \mu_{K^\star}=\zeta_\ell\}$, all factors with index greater than or equal to $K$ are the deterministic
matrices $I-\zeta_\ell\alpha_jM$. By
Lemma~\ref{pre_lemma:sum_of_product_1}, applied to
$\zeta_\ell M$, there is a deterministic constant $\cC_1(K)$ such that,
for all $t\geq1$ and $i<t$, 
\begin{equation}\label{eq:stabilized_kernel_bound}
\left\|\alpha_iM\sum_{k=i}^{t-1}
\prod_{j=i+1}^k(I-\mu_j\alpha_jM)\right\|
\leq \cC_1(K)
\qquad\text{on }E_{K,\ell}.
\end{equation}
For $i\geq K$ this follows directly from that lemma. For $i<K$, the
finitely many earlier factors and the same lemma give
\begin{align*}
&\left\|\alpha_iM\sum_{k=i}^{t-1}
\prod_{j=i+1}^k(I-\mu_j\alpha_jM)\right\|\\
&\leq \left\|\alpha_iM\sum_{k=i}^{K-1}
\prod_{j=i+1}^k(I-\mu_j\alpha_jM)\right\|+\left\|\alpha_iM
\prod_{j=i+1}^{K-1}(I-\mu_j\alpha_jM)
\sum_{k=K}^{t-1}\prod_{j=K}^k(I-\zeta_\ell\alpha_jM)\right\|\leq \cC_1(K).
\end{align*}

For any fixed $N$, on $E_{K,\ell}\cap\{\tau_N = \infty\}$, applying \eqref{eq:stabilized_kernel_bound} with $t$ replaced by each $1\leq n\leq t$ gives
\begin{align*}
\max_{1\leq n\leq t}\left\|
\frac{1}{\sqrt{t}}\sum_{i=0}^{n-1}
\left(\alpha_iM\sum_{k=i}^{n-1}
\prod_{j=i+1}^k(I-\mu_j\alpha_jM)\right)X_i
\right\|
&\leq
\frac{\cC_1(K)}{\sqrt t}\sum_{i=0}^{t-1}
\|X_i\|\1_{\{\tau_N > i\}}.
\end{align*}
Consequently, after exchanging the order of summation, Markov's inequality gives
\begin{align*}
&\PP\left(\max_{1\leq n\leq t}
\left\|
\frac{1}{\sqrt{t}}\sum_{k=0}^{n-1}\sum_{i=0}^k
\prod_{j=i+1}^k(I-\mu_j\alpha_jM)\alpha_iMX_i
\right\|>\varepsilon
\right)\\*
&\leq \PP(\tau_N<\infty)+\PP(K^\star>K)
+\sum_{\ell=1}^L
\frac{\cC_1(K)}{\varepsilon\sqrt t}\sum_{i=0}^{t-1}
\EE[\|X_i\|\1_{\{\tau_N > i\}}].
\end{align*}
For fixed $N$ and $K$, the final term on the right-hand side tends to zero because $q>1/2$.
Letting first $t\to \infty$, then $K\to\infty$, and finally
$N\to\infty$ proves the assertion.

\subsection{Proof of Lemma \ref{appendix_lemma_theorem:local_convergence_2}} \label{proof_appendix_lemma_theorem:local_convergence_2}
Let $\hat{\nabla}_{\bx}\cL(\bx_i;\xi_i) := \hat{\nabla} F(\bx_i;\xi_i) + (\hat{\nabla}c(\bx_i))^\top \blambda_i$.
For $k\geq1$, we have 
\begin{align*}
\bar{\nabla}_{\bx}\cL_k - \nabla_{\bx}\cL_k &= \bar{\bg}_k + \tilde{G}_k^\top\blambda_k - \nabla f_k - G_k^\top\blambda_k\\
&\stackrel{\eqref{eq:gradient_approx},\eqref{eq:debiased_Lagrangian_Hessian}}{=} (1-\beta_k)(\bar{\nabla}_{\bx}\cL_{k-1} - \nabla_{\bx} \cL_{k-1}) + \beta_k(\hat{\nabla}_{\bx}\cL(\bx_k; \xi_k) - \EE[\hat{\nabla}_{\bx}\cL(\bx_k; \xi_k) \given \cF_{k-1}])\\
&\quad + \beta_k (\EE[\hat{\nabla}_{\bx}\cL(\bx_k; \xi_k) \given \cF_{k-1}] - \nabla_{\bx}\cL_k) + (1-\beta_k)(\nabla f_{k-1} - \nabla f_k)\\
&\quad + (1-\beta_k)(G_{k-1}-G_k)^\top\blambda_k + (1-\beta_k)(\tilde{G}_{k-1} - G_{k-1})^\top(\blambda_k - \blambda_{k-1}) \\
&\quad + (1-\beta_k)(\bar{G}_{k-1} - \tilde{G}_{k-1})^\top\blambda_k + (\tilde{G}_k - \bar{G}_k)^\top\blambda_k.
\end{align*}
Fix the sufficiently small $\epsilon>0$ chosen in the proof of
Theorem~\ref{theorem:local_convergence}. For each deterministic $N\geq N_0$,
where $N_0$ is given by Lemma~\ref{lemma:local_convergence} for this $\epsilon$, $\PP(\tau_N(\epsilon)<\infty)\rightarrow0$ as $N\to\infty$. For $i\geq0$, let 
\begin{equation*}
\cD_i:=
\binom{
\hat{\nabla}_{\bx}\cL(\bx_i;\xi_i)
-\EE[\hat{\nabla}_{\bx}\cL(\bx_i;\xi_i)\given\cF_{i-1}]
}{\zero}.
\end{equation*}
For all $i\geq0$, \eqref{eq:hat_nabla_F_second_moment_bound},
Assumptions~\ref{assumption:boundedness_1} and
\ref{assumption:boundedness_3}, and \eqref{equ:Ups_u} give
\begin{align*}
\|\hat\nabla c(\bx_i)\|\leq
\left(\sum_{j=1}^m
\left\|
\frac{1}{2b_i}\int_{-b_i}^{b_i}
\bDelta_i^\top\nabla c^j(\bx_i+s\bDelta_i)\,ds\,
\bDelta_i^{-\top}
\right\|^2\right)^{1/2}\leq \sqrt m\,d\,\kappa_{\bDelta_1}^{-1}
\kappa_{\bDelta_2}\Upsilon_u,
\end{align*}
and 
\begin{align*}
\EE[\|\cD_i\|^2\given\cF_{i-1}] &\leq \EE[\|\hat\nabla_{\bx}\cL(\bx_i;\xi_i)\|^2\given\cF_{i-1}]\\
&\leq
2\EE[\|\hat\nabla F(\bx_i;\xi_i)\|^2\given\cF_{i-1}]
+2\Upsilon_{\blambda}^2
\EE[\|\hat\nabla c(\bx_i)\|^2\given\cF_{i-1}]
\leq\cC_1.
\end{align*}
Thus, $\{\cD_i\}_{i\geq0}$ is a martingale difference sequence satisfying
\begin{equation}
\EE[\cD_i\mid\cF_{i-1}]=\zero,
\qquad
\sup_{i\geq0}\EE\|\cD_i\|^2\leq\cC_1.
\label{eq:D_martingale_second_moment}
\end{equation}
For $k\geq0$, let
\begin{equation*}
\cR_k
:=\binom{\bar\nabla_{\bx}\cL_k-\nabla_{\bx}\cL_k}{\zero}
-\sum_{i=0}^k\prod_{j=i+1}^k(1-\beta_j)\beta_i\cD_i.
\end{equation*}
We emphasize that neither $\cD_k$ nor $\cR_k$ depends on $N$. Moreover, by the above recursion of $\bar\nabla_{\bx}\cL_k-\nabla_{\bx}\cL_k$, for any $N\geq 0$ and $k>N$, we have
\begin{align}
\cR_k
&=\prod_{j=N+1}^k(1-\beta_j)\cR_N +\sum_{i=N+1}^k\prod_{j=i+1}^k(1-\beta_j)\beta_i
\binom{
\EE[\hat\nabla_{\bx}\cL(\bx_i;\xi_i)\given\cF_{i-1}]
-\nabla_{\bx}\cL_i}{\zero}\notag\\
&\quad+\sum_{i=N+1}^k\prod_{j=i}^k(1-\beta_j)
\binom{\nabla f_{i-1}-\nabla f_i+(G_{i-1}-G_i)^\top\blambda_i}{\zero}+\sum_{i=N+1}^k\prod_{j=i}^k(1-\beta_j)
\binom{(\tilde G_{i-1}-G_{i-1})^\top(\blambda_i-\blambda_{i-1})}{\zero}\notag\\
&\quad+\sum_{i=N+1}^k\prod_{j=i}^k(1-\beta_j)
\binom{(\bar G_{i-1}-\tilde G_{i-1})^\top\blambda_i}{\zero} +\sum_{i=N+1}^k\prod_{j=i+1}^k(1-\beta_j)
\binom{(\tilde G_i-\bar G_i)^\top\blambda_i}{\zero}.
\label{eq:lagrangian_gradient_fixed_N_decomposition}
\end{align}
Since
\begin{equation*}
\cT_i=-(W^\star)^{-1}
\binom{\bar\nabla_{\bx}\cL_i-\nabla_{\bx}\cL_i}{\zero}
=-(W^\star)^{-1}\rbr{\sum_{m=0}^i\prod_{l=m+1}^i(1-\beta_l)\beta_m\cD_m+\cR_i}
\end{equation*}
for all $i\geq0$, the contribution of the martingale part is
\begin{align}
&-\frac{1}{\sqrt{t}}\sum_{k=0}^{t-1}\sum_{i=0}^k
\prod_{j=i+1}^k(I-\mu_j\alpha_j(I-C^\star))
\bar{\alpha}_i(I-C^\star)(W^\star)^{-1}
\sum_{m=0}^i\prod_{l=m+1}^i(1-\beta_l)\beta_m\cD_m\\
&= -\frac{1}{\sqrt{t}}\sum_{i=0}^{t-1}\sum_{k=i}^{t-1}
\prod_{j=i+1}^k(I-\mu_j\alpha_j(I-C^\star))
\bar{\alpha}_i(I-C^\star)(W^\star)^{-1}
\sum_{m=0}^i\prod_{l=m+1}^i(1-\beta_l)\beta_m\cD_m\\
&= -\frac{1}{\sqrt{t}}(W^\star)^{-1}
\sum_{i=0}^{t-1}\sum_{m=0}^i
\prod_{l=m+1}^i(1-\beta_l)\beta_m\cD_m
\label{eq:1_proof_appendix_lemma_theorem:local_convergence_2}\\
&\quad - \frac{1}{\sqrt{t}}\sum_{i=0}^{t-1}
\left(\sum_{k=i}^{t-1}\prod_{j=i+1}^k
(I-\mu_j\alpha_j(I-C^\star))\mu_i{\alpha}_i(I-C^\star)-I\right)
(W^\star)^{-1}\sum_{m=0}^i
\prod_{l=m+1}^i(1-\beta_l)\beta_m\cD_m
\label{eq:2_proof_appendix_lemma_theorem:local_convergence_2}\\
&\quad - \frac{1}{\sqrt{t}}\sum_{i=0}^{t-1}\sum_{k=i}^{t-1}
\prod_{j=i+1}^k(I-\mu_j\alpha_j(I-C^\star))
(\bar{\alpha}_i-\mu_i\alpha_i)(I-C^\star)
(W^\star)^{-1}\sum_{m=0}^i
\prod_{l=m+1}^i(1-\beta_l)\beta_m\cD_m.
\label{eq:3_proof_appendix_lemma_theorem:local_convergence_2}
\end{align}
For \eqref{eq:1_proof_appendix_lemma_theorem:local_convergence_2}, we use the following lemma.
\begin{lemma}\label{appendix_lemma_theorem:local_convergence_2_1}
Under the assumptions and parameter conditions of Theorem~\ref{theorem:local_convergence}, we have
\begin{equation*}
\frac{1}{\sqrt{t}}\sum_{i=0}^{t-1}\sum_{m=0}^i
\prod_{l=m+1}^i(1-\beta_l)\beta_m\cD_m
\overset{d}{\longrightarrow} \cN(\zero, \bar\bOmega^\star).
\end{equation*}
\end{lemma}
Fix a deterministic $K\geq 1$. Let $K^\star$ and
$\{\zeta_1,\ldots,\zeta_L\}$ be as in
Assumption~\ref{assumption:stepsize_selection}, and set $E_{K,\ell}:=\{K^\star\leq K,\ \mu_{K^\star}=\zeta_\ell\}$, so $\mu_j=\zeta_\ell$ for $j\geq K$ on $E_{K,\ell}$. For each $\ell\in\{1,\ldots,L\}$, define
\begin{equation*}
A_{i,\ell}^{t-1}:=
\begin{cases}
\displaystyle
\sum_{k=i}^{t-1}\prod_{j=i+1}^k
(I-\mu_j\alpha_j(I-C^\star))
\mu_i\alpha_i(I-C^\star)-I,
&0\leq i<\min\{K,t\},\\[2mm]
\displaystyle
\sum_{k=i}^{t-1}\prod_{j=i+1}^k
(I-\zeta_\ell\alpha_j(I-C^\star))
\zeta_\ell\alpha_i(I-C^\star)-I,
&K\leq i\leq t-1.
\end{cases}
\end{equation*}
On $E_{K,\ell}$, $A_{i,\ell}^{t-1}$ agrees with the matrix in
\eqref{eq:2_proof_appendix_lemma_theorem:local_convergence_2} for every
$0\leq i\leq t-1$.
Lemmas~\ref{pre_lemma:sum_of_product_1}
and~\ref{pre_lemma:sum_of_product_2} give
\begin{align}
&\sup_{\substack{t>K,\ K\leq i\leq t-1\\1\leq\ell\leq L}}
\|A_{i,\ell}^{t-1}\|\leq\cC_2,
\qquad
\frac1t\sum_{i=K}^{t-1}\|A_{i,\ell}^{t-1}\|^2\rightarrow0,
\qquad
\sup_{\substack{t\geq1,\ 0\leq i<\min\{K,t\}\\1\leq\ell\leq L}}
\|A_{i,\ell}^{t-1}\|\leq\cC_3.
\label{eq:stabilized_A_kernel}
\end{align}
Here, the first two bounds only involve the deterministic matrices $A_{i,\ell}^{t-1}$ with $i\geq K$, while the third bound holds on $E_{K,\ell}$ by the same argument as for \eqref{eq:stabilized_kernel_bound}.
Changing the order of summation and applying martingale orthogonality, Cauchy's inequality, \eqref{eq:D_martingale_second_moment}, and Lemma~\ref{pre_lemma:sum_of_product_1}, we obtain
\begin{align*}
\EE\left\|
\frac1{\sqrt t}\sum_{m=K}^{t-1}\sum_{i=m}^{t-1}
\prod_{l=m+1}^i(1-\beta_l)\beta_m
A_{i,\ell}^{t-1}(W^\star)^{-1}\cD_m
\right\|^2\leq\frac{\cC_4}{t}\sum_{i=K}^{t-1}
\|A_{i,\ell}^{t-1}\|^2\rightarrow0,
\end{align*}
where the convergence follows from \eqref{eq:stabilized_A_kernel}. On $E_{K,\ell}$, the same kernel bounds also give
\begin{align*}
\left\|
\frac1{\sqrt t}\sum_{m=0}^{K-1}\sum_{i=m}^{t-1}
\prod_{l=m+1}^i(1-\beta_l)\beta_m
A_{i,\ell}^{t-1}(W^\star)^{-1}\cD_m
\right\|\leq\frac{\cC_5}{\sqrt t}\sum_{m=0}^{K-1}\|\cD_m\|
\rightarrow0\quad\text{in probability}.
\end{align*}
For fixed $K$, the two preceding random sums converge to zero in
probability as $t\to\infty$. Moreover, $\PP((E_{K,1}\cup\cdots\cup E_{K,L})^c)=\PP(K^\star>K)\rightarrow0$ as $K\to\infty$. Therefore, letting first $t\to\infty$ and then
$K\to\infty$, the term in
\eqref{eq:2_proof_appendix_lemma_theorem:local_convergence_2} converges to zero
in probability.
The term in \eqref{eq:3_proof_appendix_lemma_theorem:local_convergence_2} is
\begin{equation*}
-\frac{1}{\sqrt{t}}\sum_{i=0}^{t-1}\sum_{k=i}^{t-1}
\prod_{j=i+1}^k(I-\mu_j\alpha_j(I-C^\star))
\mu_i\alpha_i(I-C^\star)
\frac{\bar{\alpha}_i - \mu_i\alpha_i}{\mu_i\alpha_i}
(W^\star)^{-1}\sum_{m=0}^i
\prod_{l=m+1}^i(1-\beta_l)\beta_m\cD_m.
\end{equation*}
For $m<n$, $\cD_m$ is $\cF_{n-1}$-measurable, and
\eqref{eq:D_martingale_second_moment} gives
\begin{align*}
\EE[
\rbr{(W^\star)^{-1}\cD_m}^\top
\rbr{(W^\star)^{-1}\cD_n}]
&=
\EE[
\EE[
\cD_m^\top(W^\star)^{-\top}(W^\star)^{-1}\cD_n
\given\cF_{n-1}]]
\\
&=
\EE[
\cD_m^\top(W^\star)^{-\top}(W^\star)^{-1}
\EE[\cD_n\mid\cF_{n-1}]]
=0.
\end{align*}
Hence, by \eqref{eq:stepsize_selection},
\begin{align*}
&\EE\left\|
\frac{\bar\alpha_i-\mu_i\alpha_i}{\alpha_i}(W^\star)^{-1}
\sum_{m=0}^i\prod_{l=m+1}^i(1-\beta_l)\beta_m\cD_m
\right\|^2
\\
&\leq
\cC_6\alpha_i^{2(p-1)}
\sum_{m=0}^i\prod_{l=m+1}^i(1-\beta_l)^2\beta_m^2
\EE\|\cD_m\|^2\leq
\cC_6\alpha_i^{2(p-1)}\beta_i,
\end{align*}
where the last inequality follows from
Lemma~\ref{pre_lemma:sum_of_product_2}. Therefore,
\begin{align*}
&\EE\left\|
\frac{\bar\alpha_i-\mu_i\alpha_i}{\alpha_i}(W^\star)^{-1}
\sum_{m=0}^i\prod_{l=m+1}^i(1-\beta_l)\beta_m\cD_m
\right\|
\\
&\leq
\left(
\EE\left\|
\frac{\bar\alpha_i-\mu_i\alpha_i}{\alpha_i}(W^\star)^{-1}
\sum_{m=0}^i\prod_{l=m+1}^i(1-\beta_l)\beta_m\cD_m
\right\|^2
\right)^{1/2} \leq\cC_7\alpha_i^{p-1}\beta_i^{1/2}
=O\left((i+1)^{-p_1(p-1)-p_2/2}\right).
\end{align*}
Since $p_1(p-1)+\frac{p_2}{2}>\frac12$, the term in
\eqref{eq:3_proof_appendix_lemma_theorem:local_convergence_2}
converges to zero in probability by
Lemma~\ref{appendix_lemma_theorem:local_convergence_1}.

It remains to control the remainder $\cR_k$. We fix any $N\geq N_0$ and use \eqref{eq:lagrangian_gradient_fixed_N_decomposition}. For $k>N$, on $\{\tau_N(\epsilon)>k\}$, $\tilde G_i=\bar G_i$ $(N\leq i\leq k)$ and $\|\blambda_i-\blambda_{i-1}\|
\leq\cC_8\alpha_{i-1}\|\tilde\bz_{i-1}\|$.
Lemmas~\ref{lemma:approximation_rate},
\ref{pre_lemma:sum_of_product_2}, and~\ref{lemma:local_convergence} give, for some $\cC_9>0$,
\begin{align}
\EE\left[\|\cR_k\|\1_{\{\tau_N(\epsilon)>k\}}\right]=O\left(
b_k^2
+\frac{\alpha_k}{\beta_k}(\beta_k^{1/2}+b_k^2)
+\frac{\alpha_k}{\beta_k}(\beta_k+b_k^4)
+e^{-\cC_9(k+1)^{1-p_2}}
\right).
\label{eq:nonmartingale_gradient_remainder_rate}
\end{align}
More precisely, we use the following localized first-moment bounds
for the summands with $N<i\leq k$ and the initialization term.
\begin{align*}
\EE\!\left[\left\|
\EE[\hat\nabla_{\bx}\cL(\bx_i;\xi_i)\given\cF_{i-1}]
-\nabla_{\bx}\cL_i
\right\|\1_{\{\tau_N(\epsilon)>k\}}\right]
&= O(b_i^2),\\
\EE\!\left[\left\|
\nabla f_{i-1}-\nabla f_i+(G_{i-1}-G_i)^\top\blambda_i
\right\|\1_{\{\tau_N(\epsilon)>k\}}\right]
&=
O\left(\alpha_{i-1}(\beta_{i-1}^{1/2}+b_{i-1}^2)\right),\\
\EE\!\left[\left\|
(\tilde G_{i-1}-G_{i-1})^\top
(\blambda_i-\blambda_{i-1})
\right\|\1_{\{\tau_N(\epsilon)>k\}}\right]
&=
O\left(\alpha_{i-1}(\beta_{i-1}+b_{i-1}^4)\right),\\
\EE\!\left[\left\|
\prod_{j=N+1}^k(1-\beta_j)\cR_N
\right\|\1_{\{\tau_N(\epsilon)>k\}}\right]
&=
O\left(e^{-\cC_9(k+1)^{1-p_2}}\right).
\end{align*}
The second bound follows from Lipschitz continuity and the localized second moment of the primal step in \eqref{eq:update}. For the third bound, we use $\tilde G_{i-1}=\bar G_{i-1}$ on $\{\tau_N(\epsilon)>k\}$ and apply Cauchy--Schwarz to the localized second moments of $\bar G_{i-1}-G_{i-1}$ and $\blambda_i-\blambda_{i-1}$. The last row uses $\EE\|\cR_N\|<\infty$, which follows from \eqref{eq:bar_g_r_moment_bound}, \eqref{eq:D_martingale_second_moment}, and Assumption~\ref{assumption:boundedness_1}. After summation against the weights in \eqref{eq:lagrangian_gradient_fixed_N_decomposition}, Lemma~\ref{pre_lemma:sum_of_product_2} yields the four rates in \eqref{eq:nonmartingale_gradient_remainder_rate}.
The two safeguard terms satisfy $\bar G_{i-1}-\tilde G_{i-1}=\zero$ and $\tilde G_i-\bar G_i=\zero$ for $N+1\leq i\leq k$.
on $\{\tau_N(\epsilon)>k\}$.
Moreover,
\begin{equation*}
\min\left\{
2p_3,\ p_1-\frac{p_2}{2},\ p_1-p_2+2p_3,\
p_1,\ p_1-p_2+4p_3
\right\}\stackrel{\eqref{equ:cond4.6}}{>}\frac12.
\end{equation*}
Hence, for every $N\geq N_0$, $\EE[\|\cR_k\|\1_{\{\tau_N(\epsilon)>k\}}]=O((k+1)^{-q})$ for some $q>1/2$. The same bound also holds for $N<N_0$, since $\tau_N(\epsilon)\leq\tau_{N_0}(\epsilon)$. Since $\bar\alpha_i/\alpha_i$ is uniformly bounded by \eqref{eq:bar_alpha_bound}, Lemma~\ref{appendix_lemma_theorem:local_convergence_1} then gives
\begin{align}
&\frac1{\sqrt t}\sum_{k=0}^{t-1}\sum_{i=0}^k
\prod_{j=i+1}^k(I-\mu_j\alpha_j(I-C^\star))
\bar\alpha_i(I-C^\star)(W^\star)^{-1}\cR_i
\rightarrow\zero\quad\text{in probability}.
\label{eq:nonmartingale_gradient_remainder_negligible}
\end{align}

Lemma~\ref{appendix_lemma_theorem:local_convergence_2_1} gives
\begin{equation*}
-\frac{1}{\sqrt{t}}(W^\star)^{-1}\sum_{i=0}^{t-1}
\sum_{m=0}^i
\prod_{l=m+1}^i(1-\beta_l)\beta_m\cD_m
\overset d\longrightarrow
\cN(\zero,(W^\star)^{-1}\bar\bOmega^\star(W^\star)^{-1})
=\cN(\zero,\bSigma^\star_{\rm ave}).
\end{equation*}
Combining this with the negligibility of the terms in \eqref{eq:2_proof_appendix_lemma_theorem:local_convergence_2} and \eqref{eq:3_proof_appendix_lemma_theorem:local_convergence_2}, the limit \eqref{eq:nonmartingale_gradient_remainder_negligible},
and Slutsky's theorem, we obtain
\begin{equation*}
\frac{1}{\sqrt{t}}\sum_{k=0}^{t-1}\sum_{i=0}^k
\prod_{j=i+1}^k(I-\mu_j\alpha_j(I-C^\star))\bar\alpha_i(I-C^\star)\cT_i
\overset d\longrightarrow
\cN(\zero,\bSigma^\star_{\rm ave}),
\end{equation*}
which completes the proof.

\subsection{Proof of Lemma \ref{appendix_lemma_theorem:local_convergence_2_1}}
\label{proof_appendix_lemma_theorem:local_convergence_2_1}

For $m\geq 0$, define
$\cF^{-}_{m-2/3}:=\sigma(\cF_{m-1},\bDelta_m,\tilde{\bDelta}_m)$.
Then
\begin{align}
\cD_m
&=\widehat{\nabla}\cL_m
-\EE[\widehat{\nabla}\cL_m\mid\cF_{m-1}]\\
&=
\begin{pmatrix}
\hat{\nabla}F(\bx_m;\xi_m)
-\EE[\hat{\nabla}F(\bx_m;\xi_m)\given\cF^{-}_{m-2/3}]\\
\zero
\end{pmatrix}
+\rbr{
\EE[\widehat{\nabla}\cL_m\mid\cF^{-}_{m-2/3}]
-\EE[\widehat{\nabla}\cL_m\mid\cF_{m-1}]
}\\
&=:
\begin{pmatrix}
A_m\\
\zero
\end{pmatrix}
+B_m,
\label{eq:3_proof_appendix_lemma_theorem:local_convergence_2_1}
\end{align}
where
\begin{equation*}
\widehat{\nabla}\cL_m
:=
\begin{pmatrix}
\hat{\nabla}_{\bx}\cL(\bx_m;\xi_m)\\
c_m
\end{pmatrix}.
\end{equation*}
Note that $c_m=c(\bx_m)$ is $\cF_{m-1}$-measurable
and thus cancels in the differences above.

Since $p_1-p_2<1/2$ and $r(p_1-p_2)>1$, we have $r>2$.
Writing the finite difference in \eqref{eq:random_approx}
as an integral of $\nabla F$ and using
Assumptions~\ref{assumption:boundedness_1}--\ref{assumption:boundedness_3}
and Jensen's inequality gives
$\sup_{m\ge0}\EE\|\hat\nabla F(\bx_m;\xi_m)\|^r<\infty$.
Conditional Jensen's inequality therefore gives
$\sup_{m\ge0}\EE\|A_m\|^r\le\cC_1$.
Let
$\cG_m:=\sigma(\cF_m,\bDelta_{m+1},\tilde{\bDelta}_{m+1})$
for $m\ge-1$, so that $\cG_{m-1}=\cF^-_{m-2/3}$.
The sequence $\{A_m\}$ is adapted to $\{\cG_m\}$ and
$\EE[A_m\given\cG_{m-1}]=\zero$ by its definition.
Thus $\{A_m\}$ is a square-integrable martingale difference sequence.
We apply the martingale central limit theorem
\citep{Hall2014Martingale} with this filtration.

For notational convenience, define
\begin{equation*}
\hat{\nabla}F(\bx,\bDelta,b;\xi)
:=
\frac{F(\bx+b\bDelta;\xi)-F(\bx-b\bDelta;\xi)}{2b}
\bDelta^{-1},
\end{equation*}
where only $\xi$ is random.
Also, define
$V(\bx,\bDelta,b):=\Cov(\hat{\nabla}F(\bx,\bDelta,b;\xi))$
and
$g(\bx,\bDelta):=
\Cov(\nabla F(\bx;\xi)\cdot\bDelta)
\bDelta^{-1}\bDelta^{-\top}$.
We use the following two lemmas.

\begin{lemma}
\label{appendix_lemma_theorem:local_convergence_2_1_1}
Under the assumptions and parameter conditions of
Theorem~\ref{theorem:local_convergence}, as $m\to\infty$,
\begin{equation*}
\nbr{V(\bx_m,\bDelta_m,b_m)-g(\bx_m,\bDelta_m)}
\to0
\qquad\text{almost surely}.
\end{equation*}
\end{lemma}

\begin{lemma}
\label{appendix_lemma_theorem:local_convergence_2_1_2}
Under the assumptions and parameter conditions of
Theorem~\ref{theorem:local_convergence}, as $m\to\infty$,
\begin{equation*}
\nbr{g(\bx_m,\bDelta_m)-g(\bx^\star,\bDelta_m)}
\to0
\qquad\text{almost surely}.
\end{equation*}
\end{lemma}

By these two lemmas and Ces\`aro averaging,
$\frac1t\sum_{m=0}^{t-1}
\|V(\bx_m,\bDelta_m,b_m)-g(\bx^\star,\bDelta_m)\|\to0$
almost surely.
The matrices $g(\bx^\star,\bDelta_m)$ are bounded and i.i.d.
by Assumptions~\ref{assumption:boundedness_2} and
\ref{assumption:boundedness_3}.
Hence the law of large numbers and \eqref{equ:Omega} give
\begin{align}
\frac1t\sum_{m=0}^{t-1}
\EE[A_mA_m^\top\given\cG_{m-1}]
&=
\frac1t\sum_{m=0}^{t-1}
V(\bx_m,\bDelta_m,b_m)\notag\\
&=
\frac1t\sum_{m=0}^{t-1}
g(\bx^\star,\bDelta_m)+o(1)
\to
\EE_{\bDelta\sim\cP_{\bDelta}}g(\bx^\star,\bDelta)
=
\bOmega^\star
\quad\text{almost surely}.
\label{eq:1_proof_appendix_lemma_theorem:local_convergence_2_1}
\end{align}
For every $\varepsilon>0$, the uniform $r$-moment bound also yields
\begin{equation*}
\EE\sbr{
\frac1t\sum_{m=0}^{t-1}
\EE[
\|A_m\|^2
\1_{\{\|A_m\|\ge\varepsilon\sqrt t\}}
\given\cG_{m-1}]
}
\le
\frac{\cC_1}{\varepsilon^{r-2}t^{(r-2)/2}}
\to0.
\end{equation*}
Thus the conditional Lindeberg condition holds by Markov's inequality,
and the martingale central limit theorem gives
$\frac1{\sqrt t}\sum_{m=0}^{t-1}A_m
\overset d\longrightarrow\cN(\zero,\bOmega^\star)$.

Applying Lemma~\ref{pre_lemma:sum_of_product_1} to $\beta_k$ shows that
$\sum_{i=m}^{t-1}\prod_{l=m+1}^i(1-\beta_l)\beta_m-1$
is uniformly bounded and its squared Ces\`aro mean tends to zero.
These coefficients are deterministic, so changing the order of summation
and using martingale orthogonality gives
\begin{align*}
\EE\nbr{
\frac1{\sqrt t}\sum_{i=0}^{t-1}\sum_{m=0}^i
\prod_{l=m+1}^i(1-\beta_l)\beta_m A_m
-\frac1{\sqrt t}\sum_{m=0}^{t-1}A_m
}^2 =
\frac1t\sum_{m=0}^{t-1}
\rbr{
\sum_{i=m}^{t-1}\prod_{l=m+1}^i(1-\beta_l)\beta_m-1
}^{2}
\EE\|A_m\|^2
\to0.
\end{align*}
Consequently,
\begin{equation*}
\frac1{\sqrt t}\sum_{i=0}^{t-1}\sum_{m=0}^i
\prod_{l=m+1}^i(1-\beta_l)\beta_m
\binom{A_m}{\zero}
\overset d\longrightarrow
\cN(\zero,\bar\bOmega^\star).
\end{equation*}

We now consider $B_m$ in
\eqref{eq:3_proof_appendix_lemma_theorem:local_convergence_2_1}.
The sample $\xi_m$ is independent of $\cF^-_{m-2/3}$, whereas
$(\bx_m,\blambda_m,\bDelta_m)$ is $\cF^-_{m-2/3}$-measurable.
By Taylor's theorem and
Assumptions~\ref{assumption:boundedness_1}--\ref{assumption:boundedness_3},
\begin{align}
\EE[
\hat\nabla_{\bx}\cL(\bx_m;\xi_m)
\given\cF^-_{m-2/3}]
&=
\frac{
\cL(\bx_m+b_m\bDelta_m,\blambda_m)
-\cL(\bx_m-b_m\bDelta_m,\blambda_m)
}{2b_m}
\bDelta_m^{-1}\notag\\
&=
\bDelta_m^{-1}\bDelta_m^\top\nabla_{\bx}\cL_m
+O(b_m^2),
\label{eq:4_proof_appendix_lemma_theorem:local_convergence_2_1}
\end{align}
where the constant in $O(b_m^2)$ is deterministic and independent of $m$
by the bounded third derivatives,
$\|\blambda_m\|\le\Upsilon_{\blambda}$,
and Assumption~\ref{assumption:boundedness_3}.
By Assumption~\ref{assumption:LICQ_SOSC} and the KKT conditions,
$\nabla_{\bx}\cL_m\to\nabla_{\bx}\cL^\star=\zero$
and $c_m\to c(\bx^\star)=\zero$ almost surely.
Consequently,
$\|\EE[\widehat\nabla\cL_m\given\cF^-_{m-2/3}]\|
\le
\cC_2(\|\nabla_{\bx}\cL_m\|+\|c_m\|+b_m^2)
\to0$
almost surely.
The bounded first derivatives, multipliers and directions,
together with $\|c_m\|\le\kappa_c$, also give
$\|\EE[\widehat\nabla\cL_m\given\cF^-_{m-2/3}]\|\le\cC_2$
for a deterministic constant $\cC_2$ and all $m$.
Dominated convergence therefore gives
$\EE\|\EE[\widehat\nabla\cL_m\given\cF^-_{m-2/3}]\|^2\to0$.

The vector $B_m$ is $\cF_m$-measurable, and the tower property gives
$\EE[B_m\given\cF_{m-1}]=\zero$.
Moreover, conditional Jensen's inequality gives $\EE\|B_m\|^2
\le
4\EE\|\EE[\widehat\nabla\cL_m\given\cF^-_{m-2/3}]\|^2
\to0$.
Thus $\{B_m\}$ is a square-integrable martingale difference sequence
with respect to $\{\cF_m\}$.
Changing the order of summation, martingale orthogonality and
Lemma~\ref{pre_lemma:sum_of_product_1} yield
\begin{align*}
\EE\nbr{
\frac1{\sqrt t}\sum_{i=0}^{t-1}\sum_{m=0}^i
\prod_{l=m+1}^i(1-\beta_l)\beta_m B_m
}^2 =
\frac1t\sum_{m=0}^{t-1}
\rbr{
\sum_{i=m}^{t-1}\prod_{l=m+1}^i(1-\beta_l)\beta_m
}^{2}
\EE\|B_m\|^2
\le
\frac{\cC_3}{t}\sum_{m=0}^{t-1}\EE\|B_m\|^2
\to0.
\end{align*}
Combining this with the limit for the $A_m$ terms and
\eqref{eq:3_proof_appendix_lemma_theorem:local_convergence_2_1},
we complete the proof by Slutsky's theorem.

\subsection{Proof of Lemma \ref{appendix_lemma_theorem:local_convergence_2_1_1}}\label{proof_appendix_lemma_theorem:local_convergence_2_1_1}

By Assumption~\ref{assumption:boundedness_1} and the convexity of $\cX$,
$\{\bx_m+s\bDelta_m:|s|\le b_m\}\subseteq\cX$.
Moreover,
$\sup_{|s|\le b_m}\|\bx_m+s\bDelta_m-\bx^\star\|
\le\|\bx_m-\bx^\star\|+b_m\sqrt d\,\kappa_{\bDelta_2}\to0$
almost surely.
Hence this segment is contained in $\cX_{\delta_{\rm loc}}$
for all sufficiently large $m$ almost surely.
For all such $m$, set
$(\bx,\bDelta,b)=(\bx_m,\bDelta_m,b_m)$,
with expectations and covariances taken only over $\xi$.
By Taylor's theorem,
\begin{align*}
B(\bx,\bDelta,b;\xi)
&:=\abr{
\frac{F(\bx+b\bDelta;\xi)-F(\bx-b\bDelta;\xi)}{2b}
-\nabla F(\bx;\xi)\cdot\bDelta}\\
&\le
\frac{b\|\bDelta\|^2}{4}
\rbr{
\|\nabla^2F(\bx^\prime;\xi)\|
+\|\nabla^2F(\bx^{\prime\prime};\xi)\|}
\le\frac12L(\xi)b\|\bDelta\|^2,
\end{align*}
where $\bx^\prime$ and $\bx^{\prime\prime}$ lie on the segments
joining $\bx$ to $\bx+b\bDelta$ and $\bx-b\bDelta$, respectively.
Assumptions~\ref{assumption:boundedness_2},
\ref{assumption:boundedness_3}, and
\ref{assumption:Hessian_bound}, together with \eqref{equ:Ups_u}, give $\EE[(\nabla F(\bx;\xi)\cdot\bDelta)^2]\le\cC_1$ and $\EE[B(\bx,\bDelta,b;\xi)^2]\le\cC_2b^2$. 
By the definitions of $V$ and $g$,
\begin{align*}
&V(\bx,\bDelta,b)-g(\bx,\bDelta)\\
&=
\Var\rbr{
\frac{F(\bx+b\bDelta;\xi)-F(\bx-b\bDelta;\xi)}{2b}
-\nabla F(\bx;\xi)\cdot\bDelta}
\bDelta^{-1}\bDelta^{-\top}\\
&\quad+
2\Cov\rbr{
\nabla F(\bx;\xi)\cdot\bDelta,\,
\frac{F(\bx+b\bDelta;\xi)-F(\bx-b\bDelta;\xi)}{2b}
-\nabla F(\bx;\xi)\cdot\bDelta}
\bDelta^{-1}\bDelta^{-\top}.
\end{align*}
Since $\|\bDelta^{-1}\bDelta^{-\top}\|\le d\kappa_{\bDelta_1}^{-2}$,
Cauchy's inequality gives
\begin{align*}
\|V(\bx,\bDelta,b)-g(\bx,\bDelta)\|\le
d\kappa_{\bDelta_1}^{-2}
\sbr{
\EE[B(\bx,\bDelta,b;\xi)^2]
+2\rbr{
\EE[(\nabla F(\bx;\xi)\cdot\bDelta)^2]
\EE[B(\bx,\bDelta,b;\xi)^2]
}^{1/2}}\le\cC_3(b^2+b).
\end{align*}
Consequently,
$\|V(\bx_m,\bDelta_m,b_m)-g(\bx_m,\bDelta_m)\|
\le\cC_3(b_m^2+b_m)\to0$ almost surely.

\subsection{Proof of Lemma \ref{appendix_lemma_theorem:local_convergence_2_1_2}}\label{proof_appendix_lemma_theorem:local_convergence_2_1_2}
Since $\bx_m\in\cX$, $\bx_m\to\bx^\star$, and $\cX$ is open and convex, $(1-s)\bx_m+s\bx^\star\in\cX$ for $0\le s<1$ and this segment is contained in $\{\bx:\|\bx-\bx^\star\|\le\delta_{\rm loc}\}$ for all sufficiently large $m$ almost surely. Therefore,
applying the mean-value bound on $[0,s]$ and then letting $s\to1$, using the almost-sure continuity of $\nabla F(\cdot;\xi)$, gives $\EE\nbr{\nabla F(\bx_m;\xi)-\nabla F(\bx^\star;\xi)}^2
\le \EE\sbr{L^2(\xi)}\|\bx_m-\bx^\star\|^2\to0$ almost surely. 
Moreover, Assumption~\ref{assumption:boundedness_2} and \eqref{equ:Ups_u} give the first bound below, and the preceding convergence gives the second:
\begin{equation*}
\sup_m\EE\nbr{\nabla F(\bx_m;\xi)}^2
\le 2\Upsilon_m^{2/r}+2\Upsilon_u^2,
\qquad
\EE\nbr{\nabla F(\bx^\star;\xi)}^2
\le 2\Upsilon_m^{2/r}+2\Upsilon_u^2.
\end{equation*}
Hence, by Cauchy's inequality,
\begin{align*}
&\abr{\EE\sbr{(\nabla F(\bx_m;\xi)\cdot\bDelta_m)^2}
-\EE\sbr{(\nabla F(\bx^\star;\xi)\cdot\bDelta_m)^2}}
\|\bDelta_m^{-1}\bDelta_m^{-\top}\|\\
&\le d^2\kappa_{\bDelta_1}^{-2}\kappa_{\bDelta_2}^2
\big(\EE\nbr{\nabla F(\bx_m;\xi)-\nabla F(\bx^\star;\xi)}^2\big)^{1/2}
\big(\EE\nbr{\nabla F(\bx_m;\xi)+\nabla F(\bx^\star;\xi)}^2\big)^{1/2}\le \cC_1\|\bx_m-\bx^\star\|.
\end{align*}
Similarly, $\abr{(\nabla f(\bx_m)\cdot\bDelta_m)^2
-(\nabla f(\bx^\star)\cdot\bDelta_m)^2}
\|\bDelta_m^{-1}\bDelta_m^{-\top}\|
\le \cC_2\|\bx_m-\bx^\star\|$.
Thus, 
\begin{equation*}
\nbr{g(\bx_m,\bDelta_m)-g(\bx^\star,\bDelta_m)}
\le (\cC_1+\cC_2)\|\bx_m-\bx^\star\|\to0
\qquad\text{almost surely}.
\end{equation*}

\subsection{Proof of Lemma \ref{appendix_lemma_theorem:local_convergence_3}} \label{proof_appendix_lemma_theorem:local_convergence_3}

By Lemma~\ref{pre_lemma:sketching_1}(a), we have 
\begin{align*}
\bz_{k,s} - (I - C_k)\tilde{\bz}_k = (I- \tilde{C}_k)\tilde{\bz}_k - (I-C_k)\tilde{\bz}_k = (C_k - \tilde{C}_k)\tilde{\bz}_k =: \cM_k,
\end{align*}
and $\cH_1^k =(I-C^\star)^{-1}\cM_k$. Since $\tilde\bz_k$ is
$\cF_{k-2/3}$-measurable, $\tilde C_k$ and $\cM_k$ are
$\cF_{k-1/3}$-measurable, and
$C_k=\EE[\tilde C_k\mid\cF_{k-2/3}]$, $\EE[\cM_k\mid\cF_{k-2/3}]
=(C_k-\EE[\tilde C_k\mid\cF_{k-2/3}])\tilde\bz_k
=\zero$. Fix the sufficiently small $\epsilon>0$ chosen in the proof of
Theorem~\ref{theorem:local_convergence} and a deterministic $N\geq N_0$,
where $N_0$ is given by Lemma~\ref{lemma:local_convergence}.
Moreover, $\|C_k-\tilde C_k\|\leq2$. The stopping-time bounds and
Lemma~\ref{lemma:local_convergence} give
\begin{align*}
\EE[\|\tilde\bz_k\|^2\1_{\{\tau_N(\epsilon)>k\}}]
\leq \cC_1
\EE[(\|\btheta_k-\btheta^\star\|^2+
\|\bar\nabla_{\bx}\cL_k-\nabla_{\bx}\cL_k\|^2)
\1_{\{\tau_N(\epsilon)>k\}}]=O(\beta_k+b_k^4),
\end{align*}
and hence
\begin{equation}\label{eq:sketch_martingale_second_moment}
\EE[\|\cM_k\|^2\1_{\{\tau_N(\epsilon)>k\}}]
\leq 4\EE[\|\tilde\bz_k\|^2\1_{\{\tau_N(\epsilon)>k\}}]
=O(\beta_k+b_k^4).
\end{equation}
Since $\1_{\{\tau_N(\epsilon)>k\}}$ is
$\cF_{k-2/3}$-measurable, $\EE[\cM_k\1_{\{\tau_N(\epsilon)>k\}}\mid\cF_{k-2/3}]
=\1_{\{\tau_N(\epsilon)>k\}}
\EE[\cM_k\mid\cF_{k-2/3}]=\zero$.
For $m<n$, $\cF_{m-1/3}\subseteq\cF_{n-2/3}$, and therefore
\begin{equation*}
\EE[(\cM_m\1_{\{\tau_N(\epsilon)>m\}})^\top
\cM_n\1_{\{\tau_N(\epsilon)>n\}}] =
\EE[(\cM_m\1_{\{\tau_N(\epsilon)>m\}})^\top
\1_{\{\tau_N(\epsilon)>n\}}
\EE[\cM_n\mid\cF_{n-2/3}]]
=0.
\end{equation*}

Write $\bar\alpha_i=\mu_i\alpha_i+
(\bar\alpha_i-\mu_i\alpha_i)$. Fix a deterministic $K>N$, and let $K^\star$ and $E_{K,\ell}$ be as in
the proof of Lemma~\ref{appendix_lemma_theorem:local_convergence_2}.
For $i\geq K$, the bounds obtained using the second case in the definition of
$A_{i,\ell}^{t-1}$, \eqref{eq:stabilized_A_kernel}, \eqref{eq:sketch_martingale_second_moment}, and the
preceding orthogonality give
\begin{equation*}
\EE\left\|
\frac1{\sqrt t}\sum_{i=K}^{t-1}
(I+A_{i,\ell}^{t-1})(I-C^\star)^{-1}\cM_i
\1_{\{\tau_N(\epsilon)>i\}}
\right\|^2
\leq
\frac{\cC_2(N)}{t}\sum_{i=K}^{t-1}
\|I+A_{i,\ell}^{t-1}\|^2(\beta_i+b_i^4)
\rightarrow0.
\end{equation*}
On $E_{K,\ell}$, \eqref{eq:stabilized_A_kernel} also gives
\begin{equation*}
\left\|
\frac1{\sqrt t}\sum_{i=0}^{K-1}
(I+A_{i,\ell}^{t-1})(I-C^\star)^{-1}\cM_i
\1_{\{\tau_N(\epsilon)>i\}}
\right\|
\leq
\frac{\cC_3(K)}{\sqrt t}\sum_{i=0}^{K-1}\|\cM_i\|
\rightarrow0
\quad\text{almost surely on }E_{K,\ell}.
\end{equation*}
On $E_{K,\ell}\cap\{\tau_N(\epsilon)=\infty\}$,
\begin{align*}
\frac1{\sqrt t}\sum_{k=0}^{t-1}\sum_{i=0}^k
\prod_{j=i+1}^k(I-\mu_j\alpha_j(I-C^\star))
\mu_i\alpha_i(I-C^\star)\cH_1^i=
\frac1{\sqrt t}\sum_{i=0}^{t-1}
(I+A_{i,\ell}^{t-1})(I-C^\star)^{-1}\cM_i
\1_{\{\tau_N(\epsilon)>i\}}.
\end{align*}
Since
$\{K^\star\leq K\}=\bigcup_{\ell=1}^L E_{K,\ell}$, the two
preceding displays imply, for every $\varepsilon>0$,
\begin{equation*}
\limsup_{t\to\infty}
\PP\left(
\left\|
\frac1{\sqrt t}\sum_{k=0}^{t-1}\sum_{i=0}^k
\prod_{j=i+1}^k(I-\mu_j\alpha_j(I-C^\star))
\mu_i\alpha_i(I-C^\star)\cH_1^i
\right\|>\varepsilon
\right) \leq
\PP(\tau_N(\epsilon)<\infty)+\PP(K^\star>K).
\end{equation*}

For the remaining part, \eqref{eq:stepsize_selection} and
\eqref{eq:sketch_martingale_second_moment} give
\begin{align*}
\EE\left[
\left\|
\frac{\bar\alpha_i-\mu_i\alpha_i}{\alpha_i}
(I-C^\star)^{-1}\cM_i
\right\|
\1_{\{\tau_N(\epsilon)>i\}}
\right]\leq
\cC_4\alpha_i^{p-1}
\EE[\|\cM_i\|^2\1_{\{\tau_N(\epsilon)>i\}}]^{1/2}=
O\left(\alpha_i^{p-1}(\beta_i^{1/2}+b_i^2)\right),
\end{align*}
where $p_1(p-1)+\min\left\{\frac{p_2}{2},2p_3\right\}>\frac12$ by \eqref{equ:cond4.6}. The same bound also holds for $N<N_0$, since $\tau_N(\epsilon)$ is nondecreasing in $N$. Hence, Lemma
\ref{appendix_lemma_theorem:local_convergence_1} gives
\begin{equation*}
\frac1{\sqrt t}\sum_{k=0}^{t-1}\sum_{i=0}^k
\prod_{j=i+1}^k(I-\mu_j\alpha_j(I-C^\star))
(\bar\alpha_i-\mu_i\alpha_i)(I-C^\star)\cH_1^i
\longrightarrow\zero
\quad\text{in probability}.
\end{equation*}
Letting first $t\to\infty$, then $K\to\infty$, and finally
$N\to\infty$ proves the lemma.

\section{Proofs of Section~\ref{sec:online_inference}}

\subsection{Proof of Theorem~\ref{theorem:online_FCLT}}\label{proof_of_theorem:online_FCLT}

Let $A:=I-C^\star$ and $N_t(u):=\lfloor ut\rfloor$. We use the recursive decomposition from the proof of
Theorem~\ref{theorem:local_convergence}:
\begin{equation*}
(\btheta_{k+1}-\btheta^\star)
=
(I-\mu_k\alpha_k A)(\btheta_k-\btheta^\star)
+
\bar\alpha_k A\cT_k
+
\bar\alpha_k A
(\cH_1^k+\cH_2^k+\cH_3^k+\cH_4^k+\cH_5^k+\cH_6^k).
\end{equation*}
Iterating this recursion and summing over $k$ gives
\begin{equation*}
\frac{1}{\sqrt t}
\sum_{k=0}^{N_t(u)-1}(\btheta_k-\btheta^\star)
=
\cR_{0,t}(u)
+
\cT_t(u)
+
\sum_{\ell=1}^{6}\cR_{\ell,t}(u)
-
\frac{1}{\sqrt t}(\btheta_{N_t(u)}-\btheta_0),
\end{equation*}
where
\begin{equation*}
\cR_{0,t}(u)
:=
\frac{1}{\sqrt t}
\sum_{k=0}^{N_t(u)-1}
\prod_{i=0}^{k}
(I-\mu_i\alpha_i A)(\btheta_0-\btheta^\star),
\end{equation*}
\begin{equation*}
\cT_t(u)
:=
\frac{1}{\sqrt t}
\sum_{k=0}^{N_t(u)-1}
\sum_{i=0}^{k}
\prod_{j=i+1}^{k}
(I-\mu_j\alpha_j A)
\bar\alpha_i A\cT_i,
\end{equation*}
and, for $1\le \ell\le 6$,
\begin{equation*}
\cR_{\ell,t}(u)
:=
\frac{1}{\sqrt t}
\sum_{k=0}^{N_t(u)-1}
\sum_{i=0}^{k}
\prod_{j=i+1}^{k}
(I-\mu_j\alpha_j A)
\bar\alpha_i A\cH_\ell^i.
\end{equation*}

We first show that all terms except $\cT_t(u)$ are negligible uniformly
over $u\in[0,1]$. The last term is negligible since $\sup_{u\in[0,1]}\|\btheta_{N_t(u)}-\btheta_0\|\le 2\sup_{k\geq0}\|\btheta_k-\btheta^\star\|<\infty$ almost surely by Assumption~\ref{assumption:LICQ_SOSC}. By the initial-condition estimate in
Appendix~\ref{proof_of_thm:local_convergence}, the first term satisfies 
\begin{equation}
\sup_{u\in[0,1]}\|\cR_{0,t}(u)\|
\to 0 \qquad \text{in probability.} \label{eq:7_proof_of_theorem:online_FCLT}
\end{equation}
Moreover, since $N_t(u)$ takes values in $\{0,1,\ldots,t\}$ for $u\in[0,1]$, we have for any sequence $\{X_i\}$,
\begin{equation*}
\sup_{u\in[0,1]}
\nbr{
\frac{1}{\sqrt t}
\sum_{k=0}^{N_t(u)-1}
\sum_{i=0}^{k}
\prod_{j=i+1}^{k}
(I-\mu_j\alpha_j A)
\bar\alpha_iAX_i
}
=
\max_{1\le n\le t}
\nbr{
\frac{1}{\sqrt t}
\sum_{k=0}^{n-1}
\sum_{i=0}^{k}
\prod_{j=i+1}^{k}
(I-\mu_j\alpha_j A)
\bar\alpha_iAX_i
}.
\end{equation*}
Therefore, after absorbing the uniformly bounded ratio
$\bar\alpha_i/\alpha_i$, we apply the estimates from the proof of
Theorem~\ref{theorem:local_convergence}, together with the maximal bound in
Lemma~\ref{appendix_lemma_theorem:local_convergence_1}, to
$\cH_2^i,\ldots,\cH_6^i$ and obtain
\begin{equation}
\sup_{u\in[0,1]}
\nbr{
\sum_{\ell=2}^{6}\cR_{\ell,t}(u)
}
\to 0 \qquad \text{in probability.} \label{eq:4_proof_of_theorem:online_FCLT}
\end{equation}

We next identify the weak limit of $\cT_t(u)$. We follow the same
decomposition as in the proof of Lemma~\ref{appendix_lemma_theorem:local_convergence_2}. Let
\begin{equation*}
\hat\nabla\cL_i
:=
\begin{pmatrix}
\hat\nabla F(\bx_i;\xi_i)
+
(\hat\nabla c(\bx_i))^\top\blambda_i \\
c_i
\end{pmatrix},
\end{equation*}
and, for this proof, define the martingale difference sequence $\cD_i
:=
-(W^\star)^{-1}\rbr{
\hat\nabla\cL_i
-
\EE[\hat\nabla\cL_i\mid \cF_{i-1}]}$.
The proof of Lemma~\ref{appendix_lemma_theorem:local_convergence_2} shows that the non-martingale terms in
$\bar\nabla\cL_i-\nabla\cL_i$, as well as the term arising from replacing $\bar\alpha_i$ by $\mu_i\alpha_i$, satisfy the moment condition of Lemma~\ref{appendix_lemma_theorem:local_convergence_1}. Hence, by the maximal bound in Lemma~\ref{appendix_lemma_theorem:local_convergence_1}, these terms are negligible uniformly
over $u\in[0,1]$, and $\cT_t(u)$ has the same weak limit as
\begin{equation}
\frac{1}{\sqrt t}
\sum_{k=0}^{N_t(u)-1}
\sum_{i=0}^{k}
\prod_{j=i+1}^{k}
(I-\mu_j\alpha_j A)
\mu_i\alpha_i A
\sum_{m=0}^{i}
\prod_{\ell=m+1}^{i}
(1-\beta_\ell)\beta_m\cD_m.
\label{eq:2_proof_of_theorem:online_FCLT}
\end{equation}
The fixed transformation $-(W^\star)^{-1}$ preserves the martingale
difference property and the uniform
$\varrho$-moment bound for some $\varrho\in(2,3]$, since
\eqref{equ:cond4.6} implies that the sampling moment exponent exceeds two.
Rearranging the sums in
\eqref{eq:2_proof_of_theorem:online_FCLT}, we obtain
\begin{align}
&\frac{1}{\sqrt t}
\sum_{k=0}^{N_t(u)-1}
\sum_{i=0}^{k}
\prod_{j=i+1}^{k}
(I-\mu_j\alpha_j A)
\mu_i\alpha_i A
\sum_{m=0}^{i}
\prod_{\ell=m+1}^{i}
(1-\beta_\ell)\beta_m\cD_m \notag\\
&=
\frac{1}{\sqrt t}
\sum_{m=0}^{N_t(u)-1}\cD_m
+
\frac{1}{\sqrt t}
\sum_{m=0}^{N_t(u)-1}
w_m^{N_t(u)-1}\cD_m,\label{eq:3_proof_of_theorem:online_FCLT}
\end{align}
where
\begin{equation*}
w_m^s
:=
\sum_{k=m}^{s}
\sum_{i=m}^{k}
\prod_{j=i+1}^{k}
(I-\mu_j\alpha_j A)
\mu_i\alpha_i A
\prod_{\ell=m+1}^{i}
(1-\beta_\ell)\beta_m
-I .
\end{equation*}
To identify the weak limit of the first term in \eqref{eq:3_proof_of_theorem:online_FCLT}, we split $\cD_k=\cD_k^{(1)}+\cD_k^{(2)}$ as in \eqref{eq:3_proof_appendix_lemma_theorem:local_convergence_2_1}, where
\begin{align*}
\cD_k^{(1)}&:=-(W^\star)^{-1}\rbr{\hat\nabla\cL_k-\EE[\hat\nabla\cL_k\mid\cF^{-}_{k-2/3}]},\\
\cD_k^{(2)}&:=-(W^\star)^{-1}\rbr{\EE[\hat\nabla\cL_k\mid\cF^{-}_{k-2/3}]-\EE[\hat\nabla\cL_k\mid\cF_{k-1}]},
\end{align*}
and $\cF^{-}_{k-2/3}=\sigma(\cF_{k-1},\bDelta_k,\tilde{\bDelta}_k)$. The proof of Lemma~\ref{appendix_lemma_theorem:local_convergence_2_1} shows that $\{\cD_k^{(1)}\}$ is a martingale difference sequence with respect to the filtration $\{\cG_k\}$ defined there, and the conditional covariance and Lindeberg estimates in that proof, after the fixed
transformation $-(W^\star)^{-1}$, give the limiting covariance
$(W^\star)^{-1}\bar\bOmega^\star(W^\star)^{-1}=\bSigma^\star_{\rm ave}$.
Hence, the martingale functional central limit theorem
\citep{Hall2014Martingale} yields
\begin{equation*}
\frac{1}{\sqrt t}
\sum_{k=0}^{\lfloor ut\rfloor-1}\cD_k^{(1)}
\overset{d}{\Longrightarrow}
(\bSigma^\star_{\rm ave})^{1/2}B_{d+m}(u)
\end{equation*}
in $D([0,1],\RR^{d+m})$. The same proof also shows that $\{\cD_k^{(2)}\}$ is a martingale difference sequence with respect to $\{\cF_k\}$ satisfying $\EE\|\cD_k^{(2)}\|^2\to0$. Thus, Doob's $L^2$ inequality gives
\begin{equation*}
\EE\sbr{\max_{1\le n\le t}\nbr{\frac{1}{\sqrt t}\sum_{k=0}^{n-1}\cD_k^{(2)}}^2}
\le \frac{4}{t}\sum_{k=0}^{t-1}\EE\|\cD_k^{(2)}\|^2\to0,
\end{equation*}
so the partial sums of $\cD_k^{(2)}$ are negligible uniformly over $u\in[0,1]$. Combining the above two displays, we obtain
\begin{equation}
\frac{1}{\sqrt t}
\sum_{k=0}^{\lfloor ut\rfloor-1}\cD_k
\overset{d}{\Longrightarrow}
(\bSigma^\star_{\rm ave})^{1/2}B_{d+m}(u) \label{eq:5_proof_of_theorem:online_FCLT}
\end{equation}
in $D([0,1],\RR^{d+m})$.

We next control the sketching remainder $\cR_{1,t}(u)$ uniformly over $u\in[0,1]$. We take the same sufficiently small $\epsilon$ as in the proof of
Lemma~\ref{appendix_lemma_theorem:local_convergence_3}, fix a deterministic $N$ large enough for the local estimates to apply, and write
$\tau_N=\tau_N(\epsilon)$. For the sketching error $\cM_i=A\cH_1^i$,
define $\bZ_i:=A^{-1}\cM_i\1_{\{\tau_N>i\}}$. The conditional mean
calculation in that proof shows that $\{\bZ_i\}$ is a martingale
difference sequence. By \eqref{eq:sketch_martingale_second_moment},
$\EE\|\bZ_i\|^2=O(\beta_i+b_i^4)$.
We write $n=N_t(u)-1$ and exchange the order of summation to obtain
\begin{equation*}
\cR_{1,t}(u)=\frac{1}{\sqrt t}\sum_{i=0}^{n}
\cK_{i,n}\frac{\bar\alpha_i}{\alpha_i}\cH_1^i,
\qquad
\cK_{i,n}:=\alpha_i A\sum_{k=i}^{n}
\prod_{j=i+1}^k(I-\mu_j\alpha_j A).
\end{equation*}
For each possible stabilized multiplier $\zeta_\ell$, fixing
$\mu_i=\zeta_\ell$ and replacing $\bar\alpha_i$ by $\mu_i\alpha_i$ gives
deterministic weights. In an orthonormal eigenbasis of $A$, let $X_i$
denote the coordinate of $\bZ_i$ associated with an eigenvalue
$\lambda>0$. With $a_i=\lambda\zeta_\ell\alpha_i$, we know the scalar
weight on $X_i$ is $h_{i,n}:=a_i\sum_{k=i}^{n}\prod_{j=i+1}^{k}(1-a_j)$. Since $a_i^{-1}$ is proportional to $(i+1)^{p_1}$ with $0<p_1<1$, the differences $d_i:=a_{i+1}^{-1}-a_i^{-1}$ are positive and decrease to zero. We choose a deterministic $K>N$ large enough that $a_i\le 1$ and $d_i\le 1/2$ for every $i\ge K$, every eigenvalue, and every $\zeta_\ell$. Then, for $K\le i\le n$, we have $h_{i,n}=a_i+\{1-a_i(1-d_i)\}h_{i+1,n}$ and $h_{n+1,n}=0$.
The right-hand side is a convex combination of $(1-d_i)^{-1}$ and
$h_{i+1,n}$. Since $(1-d_i)^{-1}$ is nonincreasing in $i$ and at most
$2$, backward induction gives $0\le h_{i+1,n}\le h_{i,n}\le 2$ for
$K\le i\le n$. Applying Abel summation, we obtain $|\sum_{i=K}^n h_{i,n} X_i|
\le 2\max_{K\le j\le n}|\sum_{i=K}^{j}X_i|$.
Taking the maximum over $K\le n\le t-1$ and applying Doob's $L^2$
inequality on the original probability space, we obtain
\begin{equation*}
\EE\!\left[\max_{K\le n\le t-1}
\left|\frac{1}{\sqrt t}\sum_{i=K}^n h_{i,n} X_i\right|^2\right]
\le \frac{16}{t}\sum_{i=K}^{t-1}\EE|X_i|^2
\le \frac{\cC_1(N)}{t}\sum_{i=K}^{t-1}(\beta_i+b_i^4)\to0.
\end{equation*}
The corresponding maximal $L^2$ bound for the weighted vector sums
follows by summing over the coordinates.

On $E_{K,\ell}\cap\{\tau_N=\infty\}$, where
$E_{K,\ell}=\{K^\star\le K,\ \mu_{K^\star}=\zeta_\ell\}$, the weights
for the $\mu_i\alpha_i$ part coincide with the deterministic weights for
$i\ge K$, and $\cH_1^i=\bZ_i$. The kernel bound
\eqref{eq:stabilized_kernel_bound} with $M=A$ holds uniformly in $n$, so
the finite prefix $i<K$ is negligible after division by $\sqrt t$,
uniformly over $u\in[0,1]$. Using the same bound on this event, we obtain
the following uniform upper bound on the contribution from
$\bar\alpha_i-\mu_i\alpha_i$
\begin{equation*}
\frac{\cC_2(K)}{\sqrt t}\sum_{i=0}^{t-1}
\frac{|\bar\alpha_i-\mu_i\alpha_i|}{\alpha_i}\|\bZ_i\|.
\end{equation*}
The proof of Lemma~\ref{appendix_lemma_theorem:local_convergence_3}
bounds the expectation of each summand by
$O(\alpha_i^{p-1}(\beta_i^{1/2}+b_i^2))$. Under \eqref{equ:cond4.6}, this
is $O((i+1)^{-q})$ for some $q>1/2$, so the displayed bound converges to
zero in probability. Since the events $E_{K,\ell}$ cover
$\{K^\star\le K\}$, we obtain, for every $\varepsilon>0$, $\limsup_{t\to\infty}\PP\left(
\sup_{u\in[0,1]}\|\cR_{1,t}(u)\|>\varepsilon\right)
\le \PP(K^\star>K)+\PP(\tau_N<\infty)$. Letting $K\to\infty$ and then $N\to\infty$, we obtain that
\begin{equation}
\sup_{u\in[0,1]}\|\cR_{1,t}(u)\|
\to 0 \qquad \text{in probability.} \label{eq:6_proof_of_theorem:online_FCLT}
\end{equation}
For the second term in \eqref{eq:3_proof_of_theorem:online_FCLT}, it is enough to show that
\begin{equation*}
\sup_{u\in[0,1]}
\nbr{
\frac{1}{\sqrt t}
\sum_{m=0}^{N_t(u)-1}
w_m^{N_t(u)-1}\cD_m
}
\to 0 \qquad\text{in probability.}
\end{equation*}

For each $\ell\in\{1,\ldots,L\}$, define the deterministic weights
\begin{equation*}
w_{m,\ell}^s
:=
\sum_{k=m}^{s}
\sum_{i=m}^{k}
\prod_{j=i+1}^{k}
(I-\zeta_\ell\alpha_j A)
\zeta_\ell\alpha_i A
\prod_{h=m+1}^{i}
(1-\beta_h)\beta_m-I .
\end{equation*}
We first prove the maximum bound for these weights and the original
martingale differences $\cD_m$ on the original probability space.
Take $t$ to be an integer. Then
\begin{align}
\sup_{u\in[0,1]}
\nbr{
\frac{1}{\sqrt t}
\sum_{m=0}^{N_t(u)-1}
w_{m,\ell}^{N_t(u)-1}\cD_m
} \le
\max_{0\le s\le t-1}
\nbr{
\frac{1}{\sqrt t}
\sum_{m=0}^{s}
w_{m,\ell}^{t-1}\cD_m
}
+
\max_{0\le s\le t-1}
\nbr{
\frac{1}{\sqrt t}
\sum_{m=0}^{s}
(w_{m,\ell}^s-w_{m,\ell}^{t-1})\cD_m
}.\label{eq:online_FCLT_weight_split}
\end{align}
For the first term on the right-hand side of \eqref{eq:online_FCLT_weight_split}, the weights
$w_{m,\ell}^{t-1}$ are deterministic for each fixed $t$ and $\ell$, and
$\EE[w_{m,\ell}^{t-1}\cD_m\mid\cF_{m-1}]=\zero$ for $m\ge0$.
Thus the partial sums are a martingale. Doob's maximal inequality gives,
for any $\varepsilon>0$,
\begin{equation*}
\PP\rbr{
\max_{0\le s\le t-1}
\nbr{
\frac{1}{\sqrt t}
\sum_{m=0}^{s}
w_{m,\ell}^{t-1}\cD_m
}
\ge \varepsilon
}
\le
\varepsilon^{-2}
\EE\nbr{
\frac{1}{\sqrt t}
\sum_{m=0}^{t-1}
w_{m,\ell}^{t-1}\cD_m
}^2
\to 0,
\end{equation*}
where martingale orthogonality bounds the expectation by
$\cC_2t^{-1}\sum_{m=0}^{t-1}\|w_{m,\ell}^{t-1}\|^2\to0$.
This last deterministic kernel estimate follows from
Lemmas~\ref{pre_lemma:sum_of_product_1} and
\ref{pre_lemma:sum_of_product_2}, exactly as for the stabilized
$\alpha$-kernel remainder and the $\beta$-kernel remainder in the proofs of
Lemmas~\ref{appendix_lemma_theorem:local_convergence_2} and
\ref{appendix_lemma_theorem:local_convergence_2_1}.

For the second term on the right-hand side of \eqref{eq:online_FCLT_weight_split}, splitting the
inner sum in the weight difference at $i=s$ and applying
Lemma~\ref{pre_lemma:sum_of_product_1} gives a constant $\cC_1>0$,
independent of $s,t,\ell$, such that
\begin{align*}
&\max_{0\le s\le t-1}
\nbr{
\frac{1}{\sqrt t}
\sum_{m=0}^{s}
(w_{m,\ell}^s-w_{m,\ell}^{t-1})\cD_m
}
\\
&\le
\cC_1
\max_{0\le s\le t-1}
\nbr{
\frac{1}{\sqrt t}
(\zeta_\ell\alpha_{s+1}A)^{-1}
\sum_{m=0}^{s}
\sum_{i=m}^{s}
\prod_{j=i+1}^{s+1}
(I-\zeta_\ell\alpha_jA)
\zeta_\ell\alpha_iA
\prod_{h=m+1}^{i}
(1-\beta_h)\beta_m\cD_m
}
\\
&\quad+
\cC_1^2
\max_{0\le s\le t-1}
\nbr{
\frac{1}{\sqrt t\,\beta_{s+1}}
\sum_{m=0}^{s}
\prod_{h=m+1}^{s}
(1-\beta_h)\beta_m\cD_m
}.
\end{align*}
Lemma~\ref{lemma:online_FCLT}, applied with $H=A$, the constant multiplier
sequence $(\zeta_\ell)_{j\ge0}$, and the martingale differences $\cD_m$,
shows that the first term converges to zero in probability.
The second term converges to zero in probability by
\citet[Lemma~4]{Li2023Online}, applied with the identity matrix $I_{d+m}$,
stepsizes $\beta_k$, and martingale differences $\cD_k$.
Its stepsize conditions hold since $\beta_k=\iota_2(k+1)^{-p_2}$ with
$p_2\in(1/2,1)$, and the required uniform $\varrho$-moment bound,
with $\varrho>2$, was established above.
Hence $\max_{0\le s\le t-1}
\nbr{
\frac{1}{\sqrt t}
\sum_{m=0}^{s}
(w_{m,\ell}^s-w_{m,\ell}^{t-1})\cD_m
}
\to 0$ in probability. Combining this with \eqref{eq:online_FCLT_weight_split}, we get
\begin{equation*}
\sup_{u\in[0,1]}
\nbr{
\frac{1}{\sqrt t}
\sum_{m=0}^{N_t(u)-1}
w_{m,\ell}^{N_t(u)-1}\cD_m
}
\to 0 \qquad \text{in probability.}
\end{equation*}
To compare with the original weights, let
$E_\ell:=\{\mu_{K^\star}=\zeta_\ell\}$.
On $E_\ell$, $w_m^s=w_{m,\ell}^s$ whenever $s\ge m\ge K^\star$.
The product bounds in Lemmas~\ref{pre_lemma:sum_of_product_1} and
\ref{pre_lemma:sum_of_product_2}, together with the uniform bounds on
$\mu_j$, give
$\sup_{v\ge m}(\|w_m^v\|+\|w_{m,\ell}^v\|)<\infty$
almost surely for all $m\ge0$ and $\ell\in\{1,\ldots,L\}$.
Since $K^\star<\infty$ almost surely, it follows on $E_\ell$ that
\begin{align*}
\max_{0\le s\le t-1}
\nbr{\frac1{\sqrt t}\sum_{m=0}^s
(w_m^s-w_{m,\ell}^s)\cD_m}\le
\frac1{\sqrt t}\sum_{m=0}^{K^\star-1}
\sup_{v\ge m}\|w_m^v-w_{m,\ell}^v\|\,\|\cD_m\|
\rightarrow0\qquad\text{almost surely.}
\end{align*}
This is a pathwise finite-prefix estimate and uses no stopping-time
property of $K^\star$.
Combining it with the frozen-weight convergence, for every $\varepsilon>0$
and every $\ell\in\{1,\ldots,L\}$ we obtain
\begin{equation*}
\PP\rbr{E_\ell\cap\cbr{
\sup_{u\in[0,1]}\nbr{
\frac1{\sqrt t}\sum_{m=0}^{N_t(u)-1}
w_m^{N_t(u)-1}\cD_m}>
\varepsilon}}
\rightarrow0 .
\end{equation*}
The events $E_1,\ldots,E_L$ form a finite partition up to a null set.
Summing the preceding probabilities therefore proves the required
convergence in probability for the original weights.
This convergence, together with \eqref{eq:7_proof_of_theorem:online_FCLT}, \eqref{eq:4_proof_of_theorem:online_FCLT}, \eqref{eq:5_proof_of_theorem:online_FCLT}, \eqref{eq:6_proof_of_theorem:online_FCLT}, and Slutsky's theorem, yields
\begin{equation*}
\frac{1}{\sqrt t}\sum_{k=0}^{\lfloor ut\rfloor-1}(\btheta_k-\btheta^\star)
\overset{d}{\Longrightarrow}
(\bSigma^\star_{\rm ave})^{1/2}B_{d+m}(u)
\end{equation*}
in $D([0,1],\RR^{d+m})$. This completes the proof.

\subsection{Proof of Lemma \ref{lemma:online_FCLT}}
\label{proof_of_lemma:online_FCLT}

We use the following scalar version of Lemma~\ref{lemma:online_FCLT} to
establish its vector form. The scalar result is proved in
Appendix~\ref{proof_of_lemma:proof_of_lemma:online_FCLT}.

\begin{lemma}\label{lemma:proof_of_lemma:online_FCLT}
Let $q\in(2,3]$, and let $\alpha_k=\iota_1/(k+1)^{p_1}$ and
$\beta_k=\iota_2/(k+1)^{p_2}$ satisfy $0<p_2<p_1<1$.
Fix $\vartheta\in[\vartheta_1,\vartheta_2]$, where
$0<\vartheta_1<\vartheta_2<\infty$.
Let $\{\eta_m\}_{m\ge0}$ be a real-valued martingale difference
sequence with respect to $\{\cF_m\}_{m\ge0}$ satisfying
$\sup_{m\ge0}\EE[|\eta_m|^{q}]<\infty$.
Then
\begin{equation*}
\max_{0\le s\le t-1}
\abr{
\frac{1}{\sqrt t}(\vartheta\alpha_{s+1})^{-1}
\sum_{m=0}^{s}\sum_{i=m}^{s}
\prod_{j=i+1}^{s+1}(1-\vartheta\alpha_j)
\vartheta\alpha_i
\prod_{\ell=m+1}^{i}(1-\beta_\ell)\beta_m\eta_m
}
\to0
\end{equation*}
in probability as $t\to\infty$.
\end{lemma}

For $0\le s\le t-1$, let $R_s$ denote the vector inside the norm in
Lemma~\ref{lemma:online_FCLT} with $a=s$.
By Assumption~\ref{assumption:stepsize_selection}, there exist an
almost surely finite $K^\star$ and a deterministic finite set
$\{\zeta_1,\ldots,\zeta_L\}\subseteq[\kappa_{1,\mu},\kappa_{2,\mu}]$
such that $\mu_k=\mu_{K^\star}\in\{\zeta_1,\ldots,\zeta_L\}$
for all $k\ge K^\star$.
For each $r\in\{1,\ldots,L\}$, define on the original probability space
\begin{equation*}
R_s^r:=\frac1{\sqrt t}(\zeta_r\alpha_{s+1}H)^{-1}
\sum_{m=0}^{s}\sum_{i=m}^{s}
\prod_{j=i+1}^{s+1}(I-\zeta_r\alpha_jH)\zeta_r\alpha_iH
\prod_{\ell=m+1}^{i}(1-\beta_\ell)\beta_m\bD_m.
\end{equation*}

Since $H$ is symmetric positive definite, write $H=U\Lambda U^\top$,
where $U\in\RR^{(d+m)\times(d+m)}$ is orthogonal and
$\Lambda=\operatorname{diag}(\lambda_1,\ldots,\lambda_{d+m})$
with $\lambda_a>0$ for $a=1,\ldots,d+m$.
Since $U$ is deterministic, $\{U^\top\bD_m\}$ is adapted to
$\{\mF_m\}$ and has uniformly bounded $q$-moments for
$q=\min\{\varrho,3\}>2$, with
$\EE[U^\top\bD_m\given\mF_{m-1}]
=U^\top\EE[\bD_m\given\mF_{m-1}]=\zero$.
The finite set $\{\zeta_r\lambda_a:1\le r\le L,\ 1\le a\le d+m\}$
is contained in an interval $[\vartheta_1,\vartheta_2]$ with
$0<\vartheta_1<\vartheta_2<\infty$.
Since $U^\top(I-\zeta_r\alpha_jH)U=I-\zeta_r\alpha_j\Lambda$,
Lemma~\ref{lemma:proof_of_lemma:online_FCLT}, applied with
$\vartheta=\zeta_r\lambda_a$ and $\eta_m=(U^\top\bD_m)_a$,
gives $\max_{0\le s\le t-1}|(U^\top R_s^r)_a|\to0$ in probability
for $a=1,\ldots,d+m$ and $r=1,\ldots,L$.
By $\|R_s^r\|^2=\sum_{a=1}^{d+m}|(U^\top R_s^r)_a|^2$,
$\max_{0\le s\le t-1}\|R_s^r\|\to0$ in probability
for every $r=1,\ldots,L$.

For all $j$ above a deterministic index, almost surely,
$\|I-\mu_j\alpha_jH\|\le
1-\kappa_{1,\mu}\lambda_{\min}(H)\alpha_j$.
Absorbing the finitely many earlier factors gives deterministic
constants $\cC_1,\cC_2>0$ such that, for $0\le i\le s$,
\begin{equation}
\nbr{(\mu_{s+1}\alpha_{s+1}H)^{-1}
\prod_{j=i+1}^{s+1}(I-\mu_j\alpha_jH)\mu_i\alpha_iH}
\le\frac{\cC_1\alpha_i}{\alpha_{s+1}}
\exp\rbr{-\cC_2\sum_{j=i+1}^{s+1}\alpha_j}.
\label{eq:finite_m_matrix_kernel_bound}
\end{equation}
Also, $0<\beta_j\le1$ and
$\beta_j/\alpha_j\ge\iota_2/\iota_1$ imply
$\abr{\prod_{\ell=m+1}^{i}(1-\beta_\ell)}\beta_m
\le\beta_m\exp\rbr{
-\frac{\iota_2}{\iota_1}\sum_{\ell=m+1}^{i}\alpha_\ell}$.
Set $\cC_3=\min\{\cC_2,\iota_2/\iota_1\}>0$.
Combining \eqref{eq:finite_m_matrix_kernel_bound} with the preceding
bound on the $\beta$-product and applying integral comparison gives,
for each fixed $m\ge0$,
\begin{align*}
&\nbr{(\mu_{s+1}\alpha_{s+1}H)^{-1}
\sum_{i=m}^{s}\prod_{j=i+1}^{s+1}(I-\mu_j\alpha_jH)
\mu_i\alpha_iH\prod_{\ell=m+1}^{i}(1-\beta_\ell)\beta_m}\\
&\qquad\le\cC_4(m)(s+2)
\exp\rbr{-\cC_5(s+2)^{1-p_1}}
\le\cC_6(m),\qquad s\ge m,
\end{align*}
where $\cC_5=\cC_3\iota_1/(1-p_1)>0$ and
$\cC_4(m),\cC_6(m)>0$ are deterministic constants independent of $s$ and $t$.
The last inequality follows from
$x\exp\rbr{-\cC_5x^{1-p_1}}
\le[e\cC_5(1-p_1)]^{-1/(1-p_1)}$ for every $x>0$.
The same bounds hold for $\mu_j\equiv\zeta_r$, uniformly over
$r=1,\ldots,L$.

Let $E_r:=\{\mu_{K^\star}=\zeta_r\}$.
On $E_r$, the coefficients of $\bD_m$ in $R_s$ and $R_s^r$
coincide whenever $s\ge m\ge K^\star$.
The preceding bounds therefore give, almost surely on $E_r$,
\begin{equation}
\max_{0\le s\le t-1}\|R_s-R_s^r\|
\le\frac2{\sqrt t}\sum_{m=0}^{K^\star-1}
\cC_6(m)\|\bD_m\|\to0,
\label{eq1:proof_of_lemma:online_FCLT}
\end{equation}
because $K^\star<\infty$ and every $\bD_m$ is finite almost surely.
This comparison is pathwise and does not require $K^\star$ to be a
stopping time.
For every $\varepsilon>0$, it follows that
\begin{align*}
\PP\rbr{E_r\cap
\cbr{\max_{0\le s\le t-1}\|R_s\|>\varepsilon}}
&\le
\PP\rbr{\max_{0\le s\le t-1}\|R_s^r\|>\frac\varepsilon2}\\*
&\quad+
\PP\rbr{E_r\cap
\cbr{\max_{0\le s\le t-1}\|R_s-R_s^r\|>
\frac\varepsilon2}}
\to0.
\end{align*}
The events $E_1,\ldots,E_L$ form a finite partition up to a null set.
Summing the preceding probabilities proves
$\max_{0\le s\le t-1}\|R_s\|\to0$ in probability.
This completes the proof.

\subsection{Proof of Lemma \ref{lemma:proof_of_lemma:online_FCLT}}\label{proof_of_lemma:proof_of_lemma:online_FCLT}

For $a\le b$, define $A_a^b
:=
\prod_{\ell=a}^{b}(1-\vartheta\alpha_\ell)$ and $B_a^b
:=
\prod_{\ell=a}^{b}(1-\beta_\ell)$.
with the convention that an empty product equals one.
Since $p_2<p_1<1$, there exist a fixed constant $\gamma_\beta\in(0,1)$ and a deterministic
integer $K_1$ such that, uniformly over
$\vartheta\in[\vartheta_1,\vartheta_2]$, for all $k\ge K_1$,
\begin{equation}
0<\vartheta\alpha_k<1,
\qquad 0<\beta_k<1,
\qquad
\frac{1-\beta_k}{1-\vartheta\alpha_k}
\le 1-\gamma_\beta\beta_k,\qquad \alpha_{k+2}\ge
\alpha_{k+1}(1-\vartheta_1\alpha_{k+2}).\label{eq:K1_uniform_tail_conditions}
\end{equation}
For simplicity, let
\begin{equation*}
y_{k+1}
:=
\sum_{i=0}^{k}
\sum_{j=i}^{k}
A_{j+1}^{k+1}
\vartheta\alpha_j
B_{i+1}^{j}\beta_i\eta_i .
\end{equation*}
It suffices to prove that
\begin{equation*}
\max_{0\le k\le t-1}
\frac{|y_{k+1}|}{\sqrt t\,\alpha_{k+1}}
\to0 \qquad\text{in probability.}
\end{equation*}

For each fixed $i<K_1$, the standard exponential bounds for the $A$- and
$B$-products, together with $p_2<p_1<1$, give
\begin{equation*}
\sup_{k\ge i}
\frac{1}{\alpha_{k+1}}
\abr{
\sum_{j=i}^{k}
A_{j+1}^{k+1}\vartheta\alpha_jB_{i+1}^{j}\beta_i
}
<\infty.
\end{equation*}
Consequently, the contribution of $i=0,\ldots,K_1-1$, divided by
$\sqrt t\,\alpha_{k+1}$ and maximized over $0\le k<t$, converges to zero
almost surely. It therefore remains to analyze the tail. Define
\begin{equation*}
\tilde y_{k+1}
:=
\sum_{i=K_1}^{k}
\sum_{j=i}^{k}
A_{j+1}^{k+1}
\vartheta\alpha_j
B_{i+1}^{j}\beta_i\eta_i,
\end{equation*}
where the sum is zero when $k<K_1$.

We first study the rate of $\EE[|\tilde y_{k+1}|^2]$. By the martingale
difference property and the uniform $q$-moment bound,
\begin{equation*}
\EE[|\tilde y_{k+1}|^2]
\le
\cC_1
\sum_{i=K_1}^{k}
\rbr{
\sum_{j=i}^{k}
A_{j+1}^{k+1}
\vartheta\alpha_j
B_{i+1}^{j}\beta_i
}^2 .
\end{equation*}
Moreover,
\begin{equation*}
\sum_{i=K_1}^{k}
\rbr{
\sum_{j=i}^{k}
A_{j+1}^{k+1}
\vartheta\alpha_j
B_{i+1}^{j}\beta_i
}^2\le
\cC_2
\sum_{j=K_1}^{k}
\sum_{j'=K_1}^{j}
A_{j+1}^{k+1}\alpha_j
A_{j'+1}^{k+1}\alpha_{j'}
B_{j'+1}^{j}
\sum_{i=K_1}^{j'}
(B_{i+1}^{j'})^2\beta_i^2 .
\end{equation*}
Since Lemma~\ref{pre_lemma:sum_of_product_2}(a) gives
\begin{equation}\label{eq:appendix_D:beta_second_moment}
\sum_{i=K_1}^{j'}
(B_{i+1}^{j'})^2\beta_i^2
\le
\cC_3\beta_{j'},
\end{equation}
we get
\begin{equation*}
\sum_{i=K_1}^{k}
\rbr{
\sum_{j=i}^{k}
A_{j+1}^{k+1}
\vartheta\alpha_j
B_{i+1}^{j}\beta_i
}^2\le
\cC_4
\sum_{j=K_1}^{k}
A_{j+1}^{k+1}\alpha_j
\sum_{j'=K_1}^{j}
A_{j'+1}^{k+1}\alpha_{j'}
B_{j'+1}^{j}\beta_{j'} .
\end{equation*}
By Lemma~\ref{pre_lemma:sum_of_product_2}(b), the last display is bounded by $\cC_4
\sum_{j=K_1}^{k}
A_{j+1}^{k+1}\alpha_j^2
\le
\cC_4\alpha_{k+1}$.
The last inequality uses Lemma~\ref{pre_lemma:sum_of_product_2}(a). Therefore,
\begin{equation}
\EE[|\tilde y_{k+1}|^2]
\le
\cC_5\alpha_{k+1}.
\label{eq1:proof_of_lemma:proof_of_lemma:online_FCLT}
\end{equation}

Now fix an integer $M\ge1$ and divide the interval $[K_1,t]$ into $M$
subintervals. Let
\begin{equation*}
h_{\mathtt m}:=K_1+\left\lfloor \frac{\mathtt m(t-K_1)}{M}\right\rfloor,
\qquad
\mathtt m=0,1,\ldots,M.
\end{equation*}
For $h_{\mathtt m}\le k<h_{\mathtt m+1}$, write $h:=h_{\mathtt m}$ and $h^{\prime}:=h_{\mathtt m+1}$. Then
\begin{align*}
\tilde y_{k+1}
&=
A_{h+1}^{k+1}\tilde y_h
+
\sum_{j=h}^{k}
A_{j+1}^{k+1}B_h^j\vartheta\alpha_j
\sum_{i=K_1}^{h-1}
B_{i+1}^{h-1}\beta_i\eta_i
\\
&\quad+
\frac{1}{A_{k+2}^{h^{\prime}}}
\sum_{i=h}^{k}
\sum_{j=i}^{h^{\prime}-1}
A_{j+1}^{h^{\prime}}\vartheta\alpha_j
B_{i+1}^{j}\beta_i\eta_i -
\sum_{i=h}^{k}
\sum_{j=k+1}^{h^{\prime}-1}
(A_{k+2}^{j})^{-1}
\vartheta\alpha_j
B_{i+1}^{j}\beta_i\eta_i .
\end{align*}
Thus, for every $\varepsilon>0$,
\begin{align*}
&\PP\rbr{
\sup_{h\le k<h^{\prime}}
\frac{|\tilde y_{k+1}|}{\alpha_{k+1}}
\ge
4\varepsilon\sqrt t
}
\\
&\le
\PP\rbr{
\sup_{h\le k<h^{\prime}}
\frac{A_{h+1}^{k+1}|\tilde y_h|}{\alpha_{k+1}}
\ge
\varepsilon\sqrt t
}
+
\PP\rbr{
\sup_{h\le k<h^{\prime}}
\frac{1}{\alpha_{k+1}}
\abr{
\sum_{j=h}^{k}
A_{j+1}^{k+1}B_h^j\vartheta\alpha_j
\sum_{i=K_1}^{h-1}
B_{i+1}^{h-1}\beta_i\eta_i
}
\ge
\varepsilon\sqrt t
}
\\
&\quad+
\PP\rbr{
\sup_{h\le k<h^{\prime}}
\frac{1}{\alpha_{k+1}}
\abr{
\frac{1}{A_{k+2}^{h^{\prime}}}
\sum_{i=h}^{k}
\sum_{j=i}^{h^{\prime}-1}
A_{j+1}^{h^{\prime}}\vartheta\alpha_j
B_{i+1}^{j}\beta_i\eta_i
}
\ge
\varepsilon\sqrt t
}
\\
&\quad+
\PP\rbr{
\sup_{h\le k<h^{\prime}}
\frac{1}{\alpha_{k+1}}
\abr{
\sum_{i=h}^{k}
\sum_{j=k+1}^{h^{\prime}-1}
(A_{k+2}^{j})^{-1}
\vartheta\alpha_j
B_{i+1}^{j}\beta_i\eta_i
}
\ge
\varepsilon\sqrt t
}
\\
&=:\cP_1^{\mathtt m}+\cP_2^{\mathtt m}+\cP_3^{\mathtt m}+\cP_4^{\mathtt m} .
\end{align*}

We first control $\cP_1^{\mathtt m}$. By Markov's inequality and \eqref{eq1:proof_of_lemma:proof_of_lemma:online_FCLT},
\begin{equation*}
\cP_1^{\mathtt m}
\le
\varepsilon^{-2}t^{-1}
\rbr{
\sup_{h\le k<h^{\prime}}
\frac{\alpha_hA_{h+1}^{k+1}}{\alpha_{k+1}}
}^2
\frac{\EE[|\tilde y_h|^2]}{\alpha_h^2}\le
\cC_5\varepsilon^{-2}
\frac{1}{t\alpha_h}
\rbr{
\sup_{h\le k<h^{\prime}}
\frac{\alpha_hA_{h+1}^{k+1}}{\alpha_{k+1}}
}^2 .
\end{equation*}
For $\mathtt m\ge1$, we have $h^{\prime}\le 2h$ for all sufficiently large $t$, so $\alpha_{h}/\alpha_{h^\prime} \le 2^{p_1}$. Thus, $\sup_{h\le k<h^{\prime}}\alpha_hA_{h+1}^{k+1}/\alpha_{k+1}\le 2^{p_1}$. Therefore, $\limsup_{t\to\infty}\cP_1^{\mathtt m}=0$.
For $\mathtt m=0$, we have $h=K_1$ and $\tilde y_h=\tilde y_{K_1}=0$, so again
$\limsup_{t\to\infty}\cP_1^{\mathtt m}=0$.

We next control $\cP_2^{\mathtt m}$. Let $z_h
:=
\sum_{i=K_1}^{h-1}
B_{i+1}^{h-1}\beta_i\eta_i$. Then, by the martingale difference property and \eqref{eq:appendix_D:beta_second_moment},
\begin{equation}\label{eq:appendix_D:z_second_moment}
\EE[|z_h|^2]
\le
\cC_6
\sum_{i=K_1}^{h-1}
(B_{i+1}^{h-1}\beta_i)^2
\le
\cC_6\beta_h.
\end{equation}
Moreover, for $\mathtt m\ge1$,
\begin{equation*}
\sup_{h\le k<h^{\prime}}
\frac{1}{\alpha_{k+1}}
\sum_{j=h}^{k}
A_{j+1}^{k+1}B_h^j\vartheta\alpha_j
\le
\cC_7\frac{\alpha_h}{\alpha_{h^{\prime}}}\frac{1}{\beta_h}
\sup_{h\le k<h^{\prime}}
\sum_{j=h}^{k}B_h^j\beta_h \le
\cC_7\beta_h^{-1}.
\end{equation*}
The last inequality uses Lemma~\ref{pre_lemma:sum_of_product_1}. Therefore, by Markov's inequality and \eqref{eq:appendix_D:z_second_moment},
\begin{equation*}
\cP_2^{\mathtt m}
\le
\cC_8\varepsilon^{-2}t^{-1}\beta_h^{-2}\EE[|z_h|^2]
\le
\cC_8\varepsilon^{-2}t^{-1}\beta_h^{-1}.
\end{equation*}
For $\mathtt m\ge1$, the last term converges to zero because $t\beta_h\to\infty$.
For $\mathtt m=0$, we have $z_h=0$. Hence $\limsup_{t\to\infty}\cP_2^{\mathtt m}=0$.

The following two lemmas control $\cP_3^{\mathtt m}$ and
$\cP_4^{\mathtt m}$.
\begin{lemma}
\label{lemma1:proof_of_lemma:proof_of_lemma:online_FCLT}
Under the assumptions of Lemma~\ref{lemma:proof_of_lemma:online_FCLT}, with the notation above, for $0\le \mathtt m\le M-1$, let $h=h_{\mathtt m}$ and $h^{\prime}=h_{\mathtt m+1}$. Then, for every
$\varepsilon>0$,
\begin{equation*}
\limsup_{t\to\infty}
\PP\rbr{
\sup_{h\le k<h^{\prime}}
\frac{1}{\alpha_{k+1}}
\abr{
\frac{1}{A_{k+2}^{h^{\prime}}}
\sum_{i=h}^{k}
\sum_{j=i}^{h^{\prime}-1}
A_{j+1}^{h^{\prime}}\vartheta\alpha_j
B_{i+1}^{j}\beta_i\eta_i
}
\ge
\varepsilon\sqrt t
}
\lesssim
\varepsilon^{-q}M^{-q/2},
\end{equation*}
where the constant hidden in $\lesssim$ is independent of $M$, $\mathtt m$, and $\varepsilon$.
\end{lemma}
\begin{lemma}
\label{lemma2:proof_of_lemma:proof_of_lemma:online_FCLT}
Under the assumptions of Lemma~\ref{lemma:proof_of_lemma:online_FCLT}, with the notation above, for $0\le \mathtt m\le M-1$, let $h=h_{\mathtt m}$ and $h^{\prime}=h_{\mathtt m+1}$. Then, for every
$\varepsilon>0$,
\begin{equation*}
\limsup_{t\to\infty}
\PP\rbr{
\sup_{h\le k<h^{\prime}}
\frac{1}{\alpha_{k+1}}
\abr{
\sum_{i=h}^{k}
\sum_{j=k+1}^{h^{\prime}-1}
(A_{k+2}^{j})^{-1}
\vartheta\alpha_j
B_{i+1}^{j}\beta_i\eta_i
}
\ge
\varepsilon\sqrt t
}
\lesssim
\varepsilon^{-q}M^{-q/2},
\end{equation*}
where the constant hidden in $\lesssim$ is independent of $M$, $\mathtt m$, and $\varepsilon$.
\end{lemma}
Combining the estimates for $\cP_1^{\mathtt m},\ldots,\cP_4^{\mathtt m}$
gives, for each $\mathtt m=0,\ldots,M-1$,
\begin{equation*}
\limsup_{t\to\infty}
\PP\rbr{
\sup_{h_{\mathtt m}\le k<h_{\mathtt m+1}}
\frac{|\tilde y_{k+1}|}{\alpha_{k+1}}
\ge
4\varepsilon\sqrt t
}
\le
\cC_9\varepsilon^{-q}M^{-q/2}.
\end{equation*}
Summing over $\mathtt m=0,\ldots,M-1$, we obtain
\begin{equation*}
\limsup_{t\to\infty}
\PP\rbr{
\max_{0\le k<t}
\frac{|\tilde y_{k+1}|}{\alpha_{k+1}}
\ge
4\varepsilon\sqrt t
}
\le
\cC_9\varepsilon^{-q}M^{1-q/2}.
\end{equation*}
Since $q>2$, we can let $M\to\infty$ and conclude that
\begin{equation*}
\max_{0\le k<t}
\frac{|\tilde y_{k+1}|}{\sqrt t\,\alpha_{k+1}}
\to0
\end{equation*}
in probability. Together with the finite-prefix estimate above, this proves the
desired claim for $y_{k+1}$ and completes the proof.

\subsection{Proof of Lemma~\ref{lemma1:proof_of_lemma:proof_of_lemma:online_FCLT}}\label{proof_of_lemma1:proof_of_lemma:proof_of_lemma:online_FCLT}

For $h\le k<h'$, define $b_k :=(\alpha_{k+1}A_{k+2}^{h'})^{-q}$ and $Y_k
:=
\abr{
\sum_{i=h}^{k}
\sum_{j=i}^{h'-1}
A_{j+1}^{h'}
\vartheta\alpha_j
B_{i+1}^{j}\beta_i\eta_i
}^q$.
It is enough to study the monotonicity of $a_k := (\alpha_{k+1}A_{k+2}^{h'})^{-1}$.
For $h\le k\le h'-2$, using $A_{k+2}^{h'}
=
(1-\vartheta\alpha_{k+2})A_{k+3}^{h'}$, 
we have
\begin{align*}
a_k-a_{k+1}
=
(\alpha_{k+1}A_{k+2}^{h'})^{-1}
-
(\alpha_{k+2}A_{k+3}^{h'})^{-1}
=
\rbr{
\frac{\alpha_{k+2}}{\alpha_{k+1}(1-\vartheta\alpha_{k+2})}
-1
}
a_{k+1}
=
\frac{
\alpha_{k+2}-\alpha_{k+1}(1-\vartheta\alpha_{k+2})
}{
\alpha_{k+1}(1-\vartheta\alpha_{k+2})
}
a_{k+1}.
\end{align*}
Since $h=h_{\mathtt m}\ge K_1$, the condition in \eqref{eq:K1_uniform_tail_conditions}
gives $a_k\ge a_{k+1}$ for $h\le k\le h'-2$, uniformly over
$\vartheta\in[\vartheta_1,\vartheta_2]$. Thus, $b_k$ is non-increasing on each block.

By conditional Jensen's inequality, $\{Y_k\}_{h\le k<h'}$ is a submartingale. Chow's inequality gives
\begin{equation}
(\varepsilon\sqrt t)^q
\PP(
\sup_{h\le k<h'}
b_kY_k
\ge
(\varepsilon\sqrt t)^q
)\le
\sum_{k=h}^{h'-2}
(b_k-b_{k+1})\EE[Y_k]
+
b_{h'-1}\EE[Y_{h'-1}].
\label{eq1:proof_of_lemma1:proof_of_lemma:proof_of_lemma:online_FCLT}
\end{equation}
Burkholder's and H\"older's inequalities, together with the uniform $q$-moment bound of $\eta_i$, give
\begin{align}
\EE[Y_k]
&\le
\cC_1(k-h+1)^{q/2-1}
\sum_{i=h}^{k}
\EE\abr{
\sum_{j=i}^{h'-1}
A_{j+1}^{h'}
\vartheta\alpha_j
B_{i+1}^{j}\beta_i\eta_i
}^q \notag\\
&\le
\cC_1(k-h+1)^{q/2-1}
\sum_{i=h}^{k}
\rbr{
\sum_{j=i}^{h'-1}
A_{j+1}^{h'}
\vartheta\alpha_j
B_{i+1}^{j}\beta_i
}^q \le
\cC_1(h'-h)^{q/2-1}
\sum_{i=h}^{k}
X_i^{h'},\label{eq2:proof_of_lemma1:proof_of_lemma:proof_of_lemma:online_FCLT}
\end{align}
where $X_i^{h'}
:=
\rbr{
\sum_{j=i}^{h'-1}
A_{j+1}^{h'}
\vartheta\alpha_j
B_{i+1}^{j}\beta_i
}^q$.
Substituting \eqref{eq2:proof_of_lemma1:proof_of_lemma:proof_of_lemma:online_FCLT}
into \eqref{eq1:proof_of_lemma1:proof_of_lemma:proof_of_lemma:online_FCLT}
yields
\begin{equation*}
(\varepsilon\sqrt t)^q
\PP(
\sup_{h\le k<h'}
b_kY_k
\ge
(\varepsilon\sqrt t)^q
)
\le
\cC_1(h'-h)^{q/2-1}
\sbr{
\sum_{k=h}^{h'-2}
(b_k-b_{k+1})
\sum_{i=h}^{k}
X_i^{h'}
+
b_{h'-1}
\sum_{i=h}^{h'-1}
X_i^{h'}
}.
\end{equation*}
Changing the order of summation gives
\begin{equation*}
\sum_{k=h}^{h'-2}
(b_k-b_{k+1})
\sum_{i=h}^{k}
X_i^{h'}
+
b_{h'-1}
\sum_{i=h}^{h'-1}
X_i^{h'}=
\sum_{i=h}^{h'-1}
b_iX_i^{h'}=
\sum_{i=h}^{h'-1}
\rbr{
\frac{1}{\alpha_{i+1}A_{i+2}^{h'}}
\sum_{j=i}^{h'-1}
A_{j+1}^{h'}
\vartheta\alpha_j
B_{i+1}^{j}\beta_i
}^q.
\end{equation*}
The following deterministic estimate gives the required bound.

\begin{lemma}
\label{lemma:proof_of_lemma1:proof_of_lemma:proof_of_lemma:online_FCLT}
Under the conditions on $\alpha_k, \beta_k$, and $\vartheta$ in Lemma~\ref{lemma:proof_of_lemma:online_FCLT}, with the notation above, for $q\in(2,3]$ and $K_1\le h<h'$, we have
\begin{equation*}
\sum_{i=h}^{h'-1}
\rbr{
\frac{1}{\alpha_{i+1}A_{i+2}^{h'}}
\sum_{j=i}^{h'-1}
A_{j+1}^{h'}
\vartheta\alpha_j
B_{i+1}^{j}\beta_i
}^q
\lesssim
h'-h,
\end{equation*}
where the constant hidden in $\lesssim$ is independent of $h$ and $h'$.
\end{lemma}

Applying
Lemma~\ref{lemma:proof_of_lemma1:proof_of_lemma:proof_of_lemma:online_FCLT}
gives the upper bound $\cC_2(h'-h)$ for the sum above. Therefore,
\begin{equation*}
\PP(
\sup_{h\le k<h'}
b_kY_k
\ge
(\varepsilon\sqrt t)^q
)
\le
\cC_3\varepsilon^{-q}t^{-q/2}(h'-h)^{q/2}.
\end{equation*}
Since $h'-h\le \cC_4 t/M$, we obtain
\begin{equation*}
\limsup_{t\to\infty}
\PP\rbr{
\sup_{h\le k<h'}
\frac{1}{\alpha_{k+1}}
\abr{
\frac{1}{A_{k+2}^{h'}}
\sum_{i=h}^{k}
\sum_{j=i}^{h'-1}
A_{j+1}^{h'}
\vartheta\alpha_j
B_{i+1}^{j}\beta_i\eta_i
}
\ge
\varepsilon\sqrt t
}
\lesssim
\varepsilon^{-q}M^{-q/2}.
\end{equation*}

This completes the proof.

\subsection{Proof of Lemma~\ref{lemma:proof_of_lemma1:proof_of_lemma:proof_of_lemma:online_FCLT}} \label{proof_of_lemma:proof_of_lemma1:proof_of_lemma:proof_of_lemma:online_FCLT}

For $h\le i<h'$, define
\begin{equation*}
Q_i^{h'}:=
\frac{1}{\alpha_{i+1}A_{i+2}^{h'}}
\sum_{j=i}^{h'-1}
A_{j+1}^{h'}\vartheta\alpha_jB_{i+1}^{j}\beta_i.
\end{equation*}
By \eqref{eq:K1_uniform_tail_conditions}, all factors are positive, and for
$i\le j<h'$, including $j=i$ under the empty-product convention,
\begin{equation*}
\frac{A_{j+1}^{h'}}{A_{i+2}^{h'}}B_{i+1}^{j}
= (1-\vartheta\alpha_{i+1})
\prod_{\ell=i+1}^{j}
\frac{1-\beta_\ell}{1-\vartheta\alpha_\ell}
\le \prod_{\ell=i+1}^{j}(1-\gamma_\beta\beta_\ell).
\end{equation*}
Since $\alpha_j/\alpha_{i+1}\le\alpha_i/\alpha_{i+1}\le 2^{p_1}$, $\vartheta\le\vartheta_2$, and $0<p_2<1$, Lemma~\ref{pre_lemma:sum_of_product_1}, applied with $a_k=\beta_k$ and the scalar matrix $M=\gamma_\beta$, gives
\begin{equation*}
0\le Q_i^{h'}
\le 2^{p_1}\vartheta_2\beta_i
\sum_{j=i}^{h'-1}
\prod_{\ell=i+1}^{j}(1-\gamma_\beta\beta_\ell)
\le \cC_1,
\end{equation*}
where $\cC_1$ is independent of $h$, $h'$, $i$, and $\vartheta\in[\vartheta_1,\vartheta_2]$. Therefore, $\sum_{i=h}^{h'-1}(Q_i^{h'})^q
\le \cC_1^q(h'-h)\lesssim h'-h$ for every $q\in(2,3]$. This completes the proof.

\subsection{Proof of Lemma~\ref{lemma2:proof_of_lemma:proof_of_lemma:online_FCLT}}\label{proof_of_lemma2:proof_of_lemma:proof_of_lemma:online_FCLT}

For $h\le k<h'$, write
\begin{equation*}
\sum_{i=h}^{k}
\sum_{j=k+1}^{h'-1}
(A_{k+2}^{j})^{-1}
\vartheta\alpha_jB_{i+1}^{j}\beta_i\eta_i=
\sum_{j=k+1}^{h'-1}
(A_{k+2}^{j}B_{j+1}^{h'-1})^{-1}
\vartheta\alpha_j
\sum_{i=h}^{k}B_{i+1}^{h'-1}\beta_i\eta_i .
\end{equation*}
Let
\begin{equation*}
b_k
:=
\rbr{
\frac{1}{\alpha_{k+1}}
\sum_{j=k+1}^{h'-1}
(A_{k+2}^{j}B_{j+1}^{h'-1})^{-1}
\vartheta\alpha_j
}^q
\qquad \text{and}\qquad
Y_k
:=
\abr{
\sum_{i=h}^{k}B_{i+1}^{h'-1}\beta_i\eta_i
}^q .
\end{equation*}
Then $\{Y_k\}_{h\le k<h'}$ is a submartingale. It is enough to study the
monotonicity of
\begin{equation*}
a_k
:=
\frac{1}{\alpha_{k+1}}
\sum_{j=k+1}^{h'-1}
(A_{k+2}^{j}B_{j+1}^{h'-1})^{-1}
\vartheta\alpha_j .
\end{equation*}
Since $A_{k+2}^{j}
=
(1-\vartheta\alpha_{k+2})A_{k+3}^{j}$ for $j\ge k+2$.
we have
\begin{equation*}
a_k
=
\frac{\alpha_{k+2}}{\alpha_{k+1}(1-\vartheta\alpha_{k+2})}
a_{k+1}
+
\vartheta(B_{k+2}^{h'-1})^{-1}.
\end{equation*}
Hence
\begin{align*}
a_k-a_{k+1}
=
\rbr{
\frac{\alpha_{k+2}}{\alpha_{k+1}(1-\vartheta\alpha_{k+2})}
-1
}
a_{k+1}
+
\vartheta(B_{k+2}^{h'-1})^{-1}
\ge
\frac{
\alpha_{k+2}-\alpha_{k+1}(1-\vartheta\alpha_{k+2})
}{
\alpha_{k+1}(1-\vartheta\alpha_{k+2})
}
a_{k+1}.
\end{align*}
By the condition in \eqref{eq:K1_uniform_tail_conditions}, the last lower bound is nonnegative for all $k\ge K_1$,
uniformly over $\vartheta\in[\vartheta_1,\vartheta_2]$. Thus, $b_k$ is
non-increasing for all $k\ge K_1$. Since $h=h_{\mathtt m}\ge K_1$ for every
$\mathtt m=0,\ldots,M-1$, this monotonicity holds throughout each block.

By Chow's inequality,
\begin{equation}
(\varepsilon\sqrt t)^{q}
\PP(
\sup_{h\le k<h'}b_kY_k
\ge
(\varepsilon\sqrt t)^{q}
)\le
\sum_{k=h}^{h'-2}
(b_k-b_{k+1})\EE[Y_k]
+
b_{h'-1}\EE[Y_{h'-1}] .
\label{eq1:proof_of_lemma2:proof_of_lemma:proof_of_lemma:online_FCLT}
\end{equation}
Burkholder's and H\"older's inequalities, together with the uniform $q$-moment bound on $\eta_i$, give
\begin{equation}
\EE[Y_k]
\le
\cC_1(k-h+1)^{q/2-1}
\sum_{i=h}^{k}
(B_{i+1}^{h'-1}\beta_i)^{q}\le
\cC_1(h'-h)^{q/2-1}
\sum_{i=h}^{k}
(B_{i+1}^{h'-1}\beta_i)^{q}.
\label{eq2:proof_of_lemma2:proof_of_lemma:proof_of_lemma:online_FCLT}
\end{equation}
Substituting \eqref{eq2:proof_of_lemma2:proof_of_lemma:proof_of_lemma:online_FCLT}
into \eqref{eq1:proof_of_lemma2:proof_of_lemma:proof_of_lemma:online_FCLT} and
changing the order of summation yields
\begin{align}
(\varepsilon\sqrt t)^{q}
\PP(
\sup_{h\le k<h'}b_kY_k
\ge
(\varepsilon\sqrt t)^{q}
) 
&\le
\cC_1(h'-h)^{q/2-1}
\sbr{
\sum_{k=h}^{h'-2}
(b_k-b_{k+1})
\sum_{i=h}^{k}
(B_{i+1}^{h'-1}\beta_i)^{q}
+
b_{h'-1}
\sum_{i=h}^{h'-1}
(B_{i+1}^{h'-1}\beta_i)^{q}
} \notag\\
&=
\cC_1(h'-h)^{q/2-1}
\sum_{i=h}^{h'-1}
b_i(B_{i+1}^{h'-1}\beta_i)^{q}.\label{eq3:proof_of_lemma2:proof_of_lemma:proof_of_lemma:online_FCLT}
\end{align}
It remains to bound the deterministic sum in
\eqref{eq3:proof_of_lemma2:proof_of_lemma:proof_of_lemma:online_FCLT}. We have
\begin{align*}
\sum_{i=h}^{h'-1}
b_i(B_{i+1}^{h'-1}\beta_i)^{q}
=
\sum_{i=h}^{h'-1}
\rbr{
\frac{1}{\alpha_{i+1}}
\sum_{j=i+1}^{h'-1}
(A_{i+2}^{j}B_{j+1}^{h'-1})^{-1}
\vartheta\alpha_j
B_{i+1}^{h'-1}\beta_i
}^{q}
=
\sum_{i=h}^{h'-1}
\rbr{
\sum_{j=i+1}^{h'-1}
(A_{i+2}^{j})^{-1}
\frac{\vartheta\alpha_j}{\alpha_{i+1}}
B_{i+1}^{j}\beta_i
}^{q}.
\end{align*}
Since $p_2<p_1$, the $\beta$-product dominates the inverse $\alpha$-product:
there exists a constant $\cC_2>0$ such that, uniformly in
$\vartheta\in[\vartheta_1,\vartheta_2]$, $(A_{i+2}^{j})^{-1}B_{i+1}^{j}
\le
\cC_2\prod_{\ell=i+1}^{j}(1-\gamma_\beta\beta_\ell)$,
where $\gamma_\beta$ is defined in \eqref{eq:K1_uniform_tail_conditions}.
Moreover, $\alpha_j\le \cC_2\alpha_{i+1}$ for $j\ge i+1$. Therefore, we obtain
\begin{equation*}
\sum_{j=i+1}^{h'-1}
(A_{i+2}^{j})^{-1}
\frac{\vartheta\alpha_j}{\alpha_{i+1}}
B_{i+1}^{j}\beta_i
\le
\cC_3\sum_{j=i+1}^{h'-1}
\prod_{\ell=i+1}^{j}(1-\gamma_\beta\beta_\ell)\beta_i
\le \cC_3.
\end{equation*}
The last inequality follows from Lemma~\ref{pre_lemma:sum_of_product_1}. Thus,
\begin{equation}
\sum_{i=h}^{h'-1}
b_i(B_{i+1}^{h'-1}\beta_i)^{q}
\le
\cC_4(h'-h).
\label{eq:proof_P4_deterministic_bound}
\end{equation}
Combining \eqref{eq3:proof_of_lemma2:proof_of_lemma:proof_of_lemma:online_FCLT} and
\eqref{eq:proof_P4_deterministic_bound}, we obtain
\begin{equation*}
\PP(
\sup_{h\le k<h'}b_kY_k
\ge
(\varepsilon\sqrt t)^{q}
)
\le
\cC_5\varepsilon^{-q}t^{-q/2}(h'-h)^{q/2}.
\end{equation*}
Since $h'-h\le \cC_6t/M$, it follows that
\begin{equation*}
\limsup_{t\to\infty}
\PP\rbr{
\sup_{h\le k<h'}
\frac{1}{\alpha_{k+1}}
\abr{
\sum_{i=h}^{k}
\sum_{j=k+1}^{h'-1}
(A_{k+2}^{j})^{-1}
\vartheta\alpha_jB_{i+1}^{j}\beta_i\eta_i
}
\ge
\varepsilon\sqrt t
}
\lesssim
\varepsilon^{-q}M^{-q/2}.
\end{equation*}

This completes the proof.

\subsection{Proof of Theorem~\ref{theorem:random_scaling}}
\label{proof_of_theorem:random_scaling}

Let
\begin{equation*}
C_t(u)
:=
\frac{1}{\sqrt t}
\sum_{k=0}^{\lfloor ut\rfloor-1}
\bw^\top(\btheta_k-\btheta^\star),
\qquad u\in[0,1].
\end{equation*}
Then $C_t(\cdot)\in D([0,1],\RR)$. By
Theorem~\ref{theorem:online_FCLT},
$C_t(\cdot)\overset{d}{\Longrightarrow}
\bw^\top(\bSigma^\star_{\rm ave})^{1/2}B_{d+m}(\cdot)$
in $D([0,1],\RR)$ equipped with the Skorokhod topology.
Recall that Theorem~\ref{theorem:random_scaling} assumes
$\bOmega^\star\succ0$.
This condition holds, in particular, whenever
$\gamma_F:=
\sigma_{\min}(\Cov_{\xi\sim\cP}(\nabla F(\bx^\star;\xi)))>0$.
Indeed, by \eqref{equ:Omega} and
Assumption~\ref{assumption:boundedness_3},
for any $\bv\in\RR^d$,
\begin{align*}
\bv^\top\bOmega^\star\bv
&=
\EE_{\bDelta\sim\cP_{\bDelta}}
\sbr{
(\bv^\top\bDelta^{-1})^2
\,\bDelta^\top
\Cov_{\xi\sim\cP}(\nabla F(\bx^\star;\xi))
\bDelta
}
\ge
\gamma_F\,d\,\kappa_{\bDelta_1}^2
\,\EE_{\bDelta\sim\cP_{\bDelta}}
\sbr{(\bv^\top\bDelta^{-1})^2}
\\
&=
\gamma_F\,d\,\kappa_{\bDelta_1}^2
\sum_{j=1}^d(\bv^j)^2
\,\EE\sbr{(\bDelta^j)^{-2}}
\ge
\gamma_F\,d\,
\frac{\kappa_{\bDelta_1}^2}{\kappa_{\bDelta_2}^2}
\nbr{\bv}^2,
\end{align*}
where the first inequality uses
$\nbr{\bDelta}^2\ge d\kappa_{\bDelta_1}^2$,
and the second equality uses that the entries of $\bDelta$
are mutually independent and symmetric about zero.
The following lemma ensures that the limiting variance is nonzero.

\begin{lemma}
\label{lemma:proof_of_theorem:random_scaling}
Under the assumptions of
Theorem~\ref{theorem:local_convergence},
suppose $\bOmega^\star\succ0$.
Then, for any $\bw\in\RR^{d+m}$ satisfying
$\bw\notin
\Span((\nabla c(\bx^\star))^\top)\times\{\zero_m\}$,
we have
$\sigma_{\bw}^2:=
\bw^\top\bSigma^\star_{\rm ave}\bw>0$.
\end{lemma}

By Lemma~\ref{lemma:proof_of_theorem:random_scaling},
$\sigma_{\bw}>0$, and
$\bw^\top(\bSigma^\star_{\rm ave})^{1/2}B_{d+m}(\cdot)
\overset{d}{=}\sigma_{\bw}B_1(\cdot)$.
Define
\begin{equation*}
X_t:=
\frac1t\sum_{i=1}^t
\rbr{
C_t\rbr{\frac it}-\frac it C_t(1)
}^2,
\qquad
Y_t:=
\int_0^1(C_t(u)-uC_t(1))^2\,du.
\end{equation*}
Since
$C_t(i/t)-(i/t)C_t(1)
=(i/\sqrt t)\bw^\top(\bar\btheta_i-\bar\btheta_t)$,
the definition of $V_t$ in \eqref{eq:random_scaling_limit}
gives $X_t=\bw^\top V_t\bw$.
Also,
$C_t(1)=
\sqrt t\,\bw^\top(\bar\btheta_t-\btheta^\star)$.

The process $C_t$ is constant on $[i/t,(i+1)/t)$
and $C_t(0)=0$, so
\begin{equation*}
\int_0^1 C_t^2(u)\,du
=
\frac1t\sum_{i=0}^{t-1}C_t^2\rbr{\frac it},
\qquad
\int_0^1uC_t(u)\,du
=
\frac{1}{2t^2}
\sum_{i=1}^{t-1}(2i+1)C_t\rbr{\frac it}.
\end{equation*}
Expanding $X_t-Y_t$ and using
$\sum_{i=1}^t i^2=t(t+1)(2t+1)/6$, we obtain
\begin{align}
X_t-Y_t
&=
\frac{C_t^2(1)}t
-2C_t(1)
\rbr{
\frac{C_t(1)}t
-\frac{1}{2t^2}
\sum_{i=1}^{t-1}C_t\rbr{\frac it}
}
+\frac{3t+1}{6t^2}C_t^2(1)
\notag\\
&=
\frac{C_t(1)}t\int_0^1C_t(u)\,du
-\frac{3t-1}{6t^2}C_t^2(1).
\label{eq:1_proof_of_theorem:random_scaling}
\end{align}
By Theorem~\ref{theorem:online_FCLT}
and the continuous mapping theorem,
$\sup_{u\in[0,1]}|C_t(u)|=O_{\PP}(1)$.
Since $(3t-1)/(6t^2)\le1/(2t)$,
\eqref{eq:1_proof_of_theorem:random_scaling} implies
\begin{equation}
\label{eq:appendix_D:scaling_difference_vanishes}
|X_t-Y_t|
\le
\frac{3}{2t}\sup_{u\in[0,1]}|C_t(u)|^2
\to0
\qquad\text{in probability}.
\end{equation}

The map
$f\mapsto
\rbr{f(1),\int_0^1(f(u)-uf(1))^2\,du}$
is continuous at continuous paths for the Skorokhod topology,
because Skorokhod convergence to a continuous path implies
uniform convergence.
The continuous mapping theorem therefore gives the joint limit
of $(C_t(1),Y_t)$.
Combining it with
\eqref{eq:appendix_D:scaling_difference_vanishes}
by Slutsky's theorem yields
\begin{equation*}
(C_t(1),X_t)
\overset{d}{\longrightarrow}
\rbr{
\sigma_{\bw}B_1(1),
\sigma_{\bw}^2
\int_0^1(B_1(u)-uB_1(1))^2\,du
}.
\end{equation*}
The integral is positive almost surely:
otherwise continuity would imply
$B_1(u)-uB_1(1)=0$ for every $u$,
in particular at $u=1/2$.
But $B_1(1/2)-B_1(1)/2$ is Gaussian with variance $1/4$,
so this event has probability zero.
Since $\sigma_{\bw}>0$, the limiting second component
is positive almost surely,
and the portmanteau theorem gives $\PP(X_t=0)\to0$.
Define both ratios below to be zero on $\{X_t=0\}$.
The continuous mapping theorem,
applied to $(x,y)\mapsto x/\sqrt y$ on $y>0$,
now gives
\begin{equation*}
\frac{
\sqrt t\,\bw^\top(\bar\btheta_t-\btheta^\star)
}{
\sqrt{\bw^\top V_t\bw}
}
=
\frac{C_t(1)}{\sqrt{X_t}}
\overset{d}{\longrightarrow}
\frac{
B_1(1)
}{
\sqrt{\int_0^1(B_1(u)-uB_1(1))^2\,du}
}.
\end{equation*}
This completes the proof.

\subsection{Proof of Lemma~\ref{lemma:proof_of_theorem:random_scaling}}\label{proof_of_lemma:proof_of_theorem:random_scaling}

By the definition of $\bSigma^\star_{\rm ave}$, we have $\bSigma^\star_{\rm ave}
=
(W^\star)^{-1}
\begin{pmatrix}
\bOmega^\star & \zero\\
\zero & \zero
\end{pmatrix}
(W^\star)^{-\top}$. 
Let 
$(\bu^\top,\bv^\top)^\top
:=
(W^\star)^{-\top}\bw$, where $\bu\in \RR^d$, $\bv\in \RR^m$. Then $\bw^\top\bSigma^\star_{\rm ave}\bw
=
\bu^\top\bOmega^\star \bu$.
Since $\bOmega^\star\succ0$, if
$\bw^\top\bSigma^\star_{\rm ave}\bw=\zero$, then $\bu=\zero$. Hence
\begin{equation*}
\bw
=
(W^\star)^{\top}
\begin{pmatrix}
\zero\\
\bv
\end{pmatrix}.
\end{equation*}
Using
\begin{equation*}
W^\star
=
\begin{pmatrix}
\nabla_{\bx}^2\cL^\star & (\nabla c(\bx^\star))^\top\\
\nabla c(\bx^\star) & \zero
\end{pmatrix},
\end{equation*}
we obtain
\begin{equation*}
\bw
=
\begin{pmatrix}
(\nabla c(\bx^\star))^\top v\\
\zero
\end{pmatrix}
\in
\Span((\nabla c(\bx^\star))^\top)\times\{\zero_m\},
\end{equation*}
which contradicts the assumption on $\bw$. Therefore, $\bw^\top\bSigma^\star_{\rm ave}\bw>0$. This completes the proof.

\end{document}